\documentclass[acmsmall]{acmart}

\AtBeginDocument{%
  \providecommand\BibTeX{{%
    \normalfont B\kern-0.5em{\scshape i\kern-0.25em b}\kern-0.8em\TeX}}}

\usepackage{filecontents}
\usepackage{multirow}
\usepackage{caption}
\usepackage{fancyhdr,graphicx,amsmath,amssymb}
\usepackage[ruled,vlined]{algorithm2e}
\include{pythonlisting}

\usepackage[]{graphicx}
\graphicspath{{../pdf/}{../jpeg/}}
\DeclareGraphicsExtensions{.pdf,.jpeg,.png}
\usepackage{array}

\usepackage{fixltx2e}
\usepackage{stfloats}
\usepackage{url}
\usepackage{times}
\usepackage{color}
\usepackage{verbatim}
\usepackage{hyperref}
\usepackage{subfiles}
\usepackage{caption}
\usepackage{subcaption}
\usepackage{lipsum}
\usepackage{xr}

\usepackage{export}

\usepackage{nameref}
\usepackage{zref-xr}
\usepackage{adjustbox}
\usepackage{amsmath, amsthm, amssymb}
\usepackage[ansinew]{inputenc}
\usepackage{mathtools}
\usepackage{esvect}
\usepackage{layouts}
\usepackage{xspace}	
\usepackage{siunitx}
\usepackage{soul}
\usepackage{blindtext}
\usepackage{tabularx}
\usepackage{etoolbox}
\usepackage[font=small,skip=0pt]{caption}
\usepackage{setspace}
\makeatletter
\def\@copyrightpermission{}
\makeatother
\begin{document}
\settopmatter{printacmref=false} 
\renewcommand\footnotetextcopyrightpermission[1]{} \authorsaddresses{}

\title{The Role of Compute in Autonomous Aerial Vehicles}

\author{Behzad Boroujerdian}
\email{behzadboro@gmail.com}
\affiliation{%
  \institution{University of Texas at Austin}
}

\author{Hasan Genc}
\email{hngenc@berkeley.edu}
\affiliation{%
  \institution{University of Texas at Austin and University of California Berkeley}
}
\author{Srivatsan Krishnan}
\email{srivatsan@seas.harvard.edu}
\affiliation{%
  \institution{Harvard University}
}
\author{Bardienus Pieter Duisterhof}
\email{B.P.Duisterhof@student.tudelft.nl}
\affiliation{%
  \institution{Harvard University and Delft University of Technology}
}

\author{Brian Plancher}
\email{brian_plancher@g.harvard.edu}
\affiliation{%
  \institution{Harvard University}
}
\author{Kayvan Mansoorshahi}
\email{kayvan.mansoor@gmail.com}
\affiliation{%
  \institution{University of Texas at Austin}
}
\author{Marcelino Almeida}
\email{marcelino.malmeidan@utexas.edu}
\affiliation{%
  \institution{University of Texas at Austin}
}

\author{Aleksandra Faust}
\email{sandrafaust@google.com}
\affiliation{%
  \institution{Google Brain}
}

\author{Vijay Janapa Reddi}
\email{vj@eecs.harvard.edu}
\affiliation{%
\institution{Harvard University and The University of Texas at Austin}
}

\renewcommand{\shortauthors}{Boroujerdian, et al.}

\hyphenation{op-tical net-works semi-conduc-tor}
\definecolor{fxnote}{rgb}{0.8000,0.0000,0.0000}
\renewcommand{\paragraph}[1]{\textbf{#1}}
\newcommand{\bench}[1]{\textit{#1}}
\newcommand{\solo}[0]{3DR~Solo\xspace}
\newcommand{\red}[1]{\textcolor{red}{#1}}
\newcommand{\blue}[1]{\textcolor{blue}{#1}}
\newcommand{\blueDebug}[1]{}
\newcommand{\Sec}[1]{Section~\ref{#1}}
\newcommand{\Fig}[1]{Figure~\ref{#1}}
\newcommand{\Tbl}[1]{Table~\ref{#1}}

\pagestyle{plain} 
\maketitle

Autonomous and mobile cyber-physical machines are becoming an inevitable part of our future. In particular, unmanned aerial vehicles have seen a resurgence in activity. With multiple use cases, such as surveillance, search and rescue, package delivery, and more, these unmanned aerial systems are on the cusp of demonstrating their full potential. Despite such promises, these systems face many challenges, one of the most prominent of which is their low endurance caused by their limited onboard energy. 
    Since the success of a mission depends on whether the drone can finish it within such duration and before it runs out of battery,
    improving both the time and energy associated with the mission are of high importance.  
    Such improvements have traditionally arrived at through the use of better algorithms. But our premise is that more powerful and efficient onboard compute can also address the problem. In this paper, we investigate how the compute subsystem, in a cyber-physical mobile machine, such as a Micro Aerial Vehicle (MAV), can impact mission time and energy.
    Specifically, we pose the question as ``\textit{what is the role of computing for cyber-physical mobile robots?}'' 
We show that compute and motion are tightly intertwined, and as such \textit{a close examination of cyber and physical processes and their impact on one another is necessary.} We show different ``impact paths'' through which compute impacts mission metrics and examine them using a combination
of analytical models, simulation, micro and end-to-end benchmarking. 
To enable similar studies, we open sourced \textit{MAVBench}, our tool-set, which consists of \textbf{(1)} a closed-loop real-time feedback simulator and \textbf{(2)} an end-to-end benchmark suite comprised of
state-of-the-art kernels. By combining MAVBench, analytical modeling, and an understanding of various compute impacts, we show up to 2X and 1.8X improvements for mission time and mission energy for two optimization case studies. 
Our investigations, as well as our optimizations, show that \textit{cyber-physical co-design, a methodology with which both the cyber and physical processes/quantities of the robot are developed with consideration of one another, similar to hardware-software co-design, is necessary for arriving at the design of the optimal robot.}


\section{Introduction}
\begin{figure}[t!]
\centering
    \begin{subfigure}{.45\columnwidth}
    \centering
    \includegraphics[trim=0 0 0 0, clip, width=1.0\columnwidth]{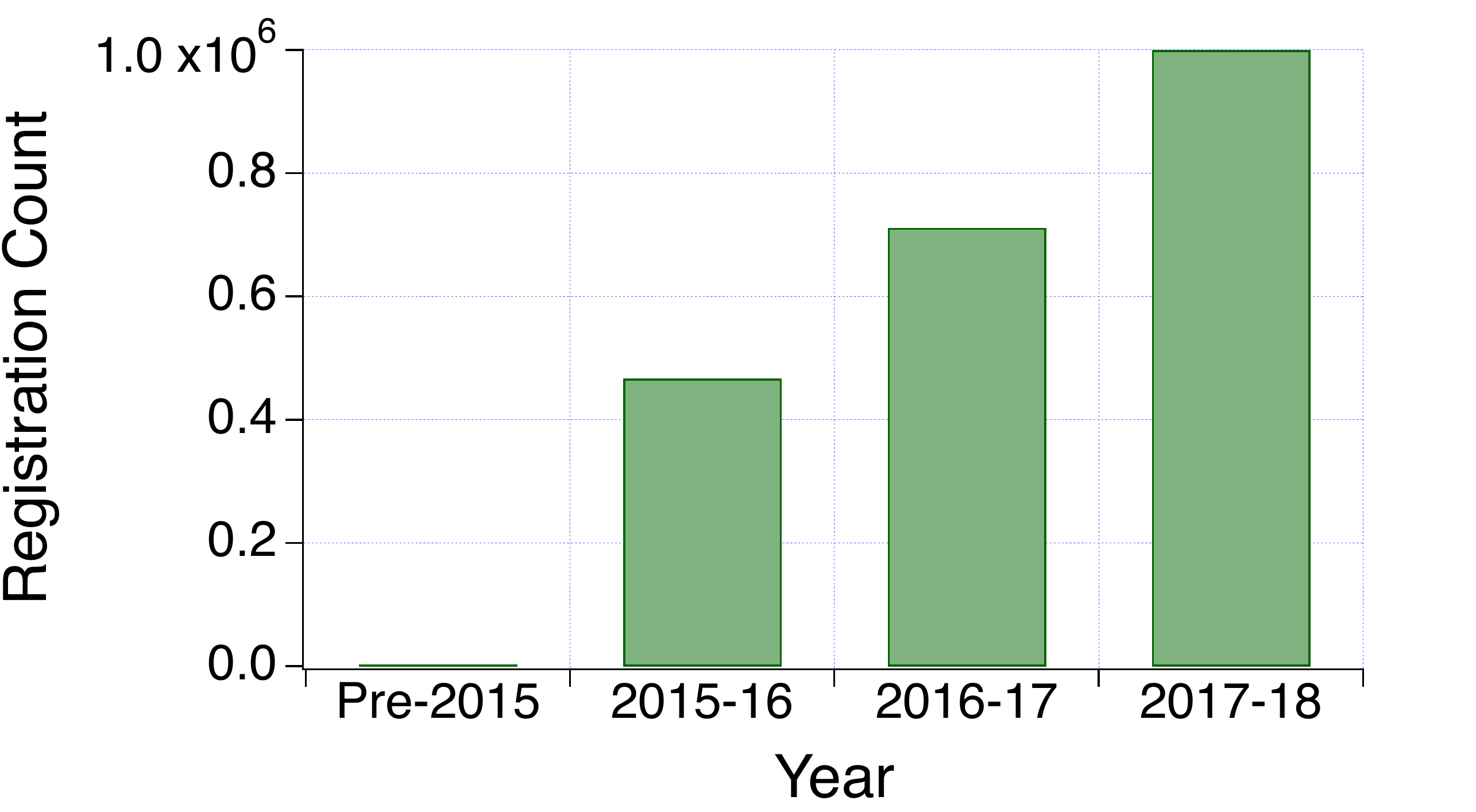}
    \caption{Registered UAVs in the FAA database.}
    \label{fig:faa-registrations}
    \end{subfigure}
    \hspace{5pt}
    \begin{subfigure}{.52\columnwidth}
   \centering
    \includegraphics[trim=0 0 0 0, clip, width=1.0\columnwidth]{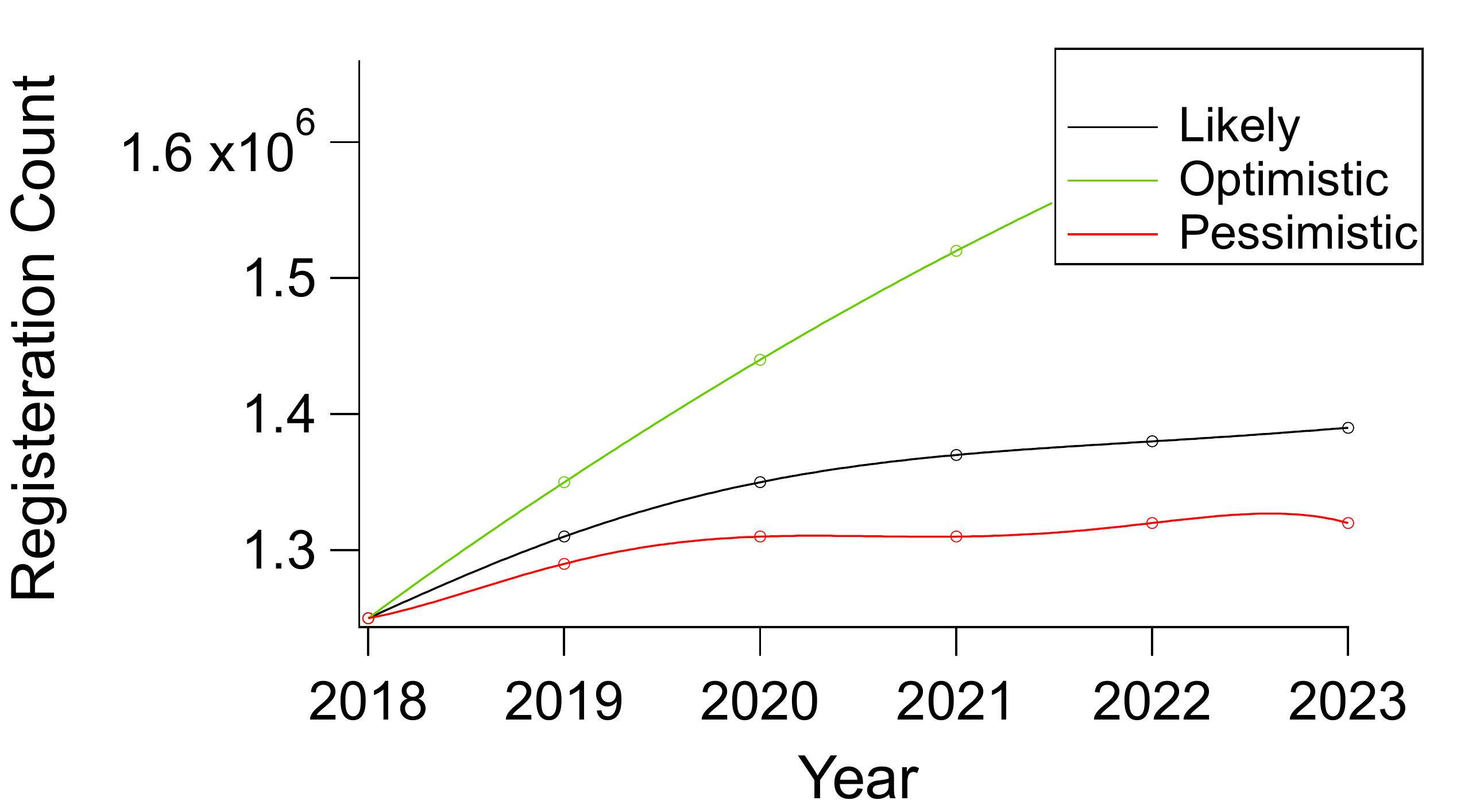}
    \caption{Predicted number of UAVs by the FAA.}
    \label{fig:faa-prediction}
    \end{subfigure}
       \caption{Currently and predicted number of registered UAVs according to FAA~\cite{FAA:online}. A visible growth indicates the significance of these vehicles which demands system designers attention.}
      \label{fig:current_predicated_UAVs}
\end{figure}

Unmanned aerial vehicles (UAVs), or drones, are rapidly increasing in number. Between 2015, when the U.S. Federal Aviation Administration (FAA) first required every owner to register their drone, and
2018, the number of drones has grown by over 100\%. The FAA indicates that there are over a million drones in the FAA drone registry database (Figure~\ref{fig:faa-registrations}). Due in large part to an increasing set of use cases, including sports photography~\cite{drone-sports1}, surveillance~\cite{drone-surveillance}, disaster management, search and rescue~\cite{disaster-drone,nepal-earthquake-drone}, transportation and package delivery~\cite{Amazondel:online,CSD-Amazon-Drone-patents,drone-package-delivery-google}, 
FAA predicts that this number will only increase over the next 5 years as indicated by the projections shown in Figure~\ref{fig:faa-prediction}.



The growth and significance of this emerging new domain 
calls for \emph{cyber-physical co-design} involving computer and system architects. Traditionally, the robotics domain has mostly been left to experts in mechanical engineering and controls. However, as we show in this paper, drones are challenged by limited battery capacity and therefore low endurance (how long the drone can last in the air). For example, most off-the-shelf drones have an endurance of less than 20 minutes ~\cite{CSD-Amazon-Drone-patents}. This need for greater endurance demands the attention of hardware and system architects. 

\newcommand{\csig}{cyber-physical interaction graph\xspace}

In this paper, we investigate how the compute subsystem in a cyber-physical mobile machine, such as a Micro UAV (MAV), can impact the mission time and energy and consequently the MAV's endurance. 
We illustrate that \textit{fundamentals
of compute and motion are tightly intertwined}. Hence, an efficient compute subsystem can directly impact mission time and energy. 
We use a directed acyclic graph, which we call the ``\csig", to capture the different ways (paths in the graph or ``impact paths'') through which compute can affect mission time and energy. 
By analyzing the impact paths, one can observe the effect that each subsystem has on each mission metric. Furthermore, we can find out through which cyber and physical quantities (e.g., response time and compute mass) this impact occurs. 

To study the different impact paths, we use a mixture of analytical models, benchmarks, and simulations. For our analytical models, we use detailed physics to show how compute impacts cyber and physical quantities and ultimately mission metrics such as mission time and energy. For example, through
derivation, we show how compute impacts response time, a cyber quantity, which impacts velocity, a physical quantity, which in turn impacts mission time. For our simulator and benchmarks, we address the lack of systematic benchmarks and infrastructure for research by developing MAVBench, a first of its kind platform for the holistic evaluation of aerial agents, involving a closed-loop simulation framework and a benchmark suite. MAVBench facilitates the integrated study of performance and energy efficiency of not only the compute subsystem in isolation but also the compute subsystem's dynamic and runtime interactions with the simulated MAV. 




MAVBench, which is a framework that is built on top of AirSim~\cite{Airsim_paper}, faithfully captures all of the interactions a real MAV encounters and ensures reproducible runs across experiments, starting from the software layers down to the hardware layers. Our simulation setup uses a hardware-in-the-loop configuration that can enable hardware and software architects to perform co-design studies to optimize system performance by considering the entire vertical application stack, including the Robotics Operating System (ROS). Our setup reports a variety of quality-of-flight (QoF) metrics, such as the performance, power consumption, and trajectory statistics of the MAV.

MAVBench includes an application suite covering a variety of popular applications of micro aerial vehicles: 
\bench{Scanning}, \bench{Package Delivery}, \bench{Aerial Photography}, \bench{3D Mapping} and \bench{Search and Rescue.} MAVBench applications are comprised of holistic end-to-end application dataflows found in a typical real-world drone application. These applications' dataflows are comprised of several state-of-the-art computational kernels, such as object detection~\cite{yolo16,hog}, occupancy map generation~\cite{octomap}, motion planning~\cite{ompl}, localization~\cite{orbslam2,vins-mono}, which we integrated together to create complete applications. 


MAVBench enables us to understand and quantify the energy and performance demands of typical MAV applications from the underlying compute subsystem perspective. More specifically, it allows us to study how compute impacts cyber and physical quantities along with the downstream effects of those impacts on mission metrics. It helps designers optimize MAV designs by answering the fundamental question of \emph{what is the role of compute in the operation of autonomous MAVs?}

Using the analytical models, benchmarks, and simulations, we quantitatively show that compute has a significant impact on MAV's mission time and energy. We bin the various impact paths mentioned above to three clusters and study them separately and then simultaneously (holistically). First, by studying each cluster independently, we isolate its effect to gain a better insight into its impact, as well as its progress along the impact path. Second, by studying them simultaneously, we illustrate the clusters aggregate impacts. The latter approach is especially valuable when the clusters have opposite impacts, and hence understanding compute's overall impact requires a holistic outlook.




Finally, we present two optimization case studies showing how our tool-sets combined with an understanding of the compute impact on the robot can be used to improve mission time and energy. In the first case study, we examine a sensor-cloud architecture for drones where the computation is distributed across the edge and the cloud to improve both mission time and energy. Such an architecture shows a reduction in the drone's overall mission time and energy by as much as 2X and 1.3X respectively when the cloud support is enabled. The second case study targets Octomap~\cite{octomap}, a computationally intensive kernel that is at the heart of many of the MAVBench applications, and demonstrates how a runtime dynamic knob tuning can reduce overall mission time and energy consumption to improve battery consumption by as much as 1.8X.

In summary, we make the following contributions:

\begin{itemize}
    \item We introduce an acyclic directed graph called the \csig to capture various impact paths that originate from compute in cyber-physical systems.
    \item We present various analytical models demonstrating these impacts for MAVs. 
    \item We provide an open-source, closed-loop simulation framework to capture these impacts. This enables hardware and software architects to perform performance and energy optimization studies that are relevant to compute subsystem design and architecture.
    \item We introduce an end-to-end benchmark suite, comprised of several workloads and their corresponding state-of-the-art kernels. These workloads represent popular real-world use cases of MAVs further aiding designers in their end-to-end studies.
    \item Combining our tool-sets and analytical models, we demonstrate the role of compute and its relationship with mission time and energy for unmanned MAVs.
    \item We use our framework to present optimization case studies that exploit compute's impact on performance and energy of MAV systems.
\end{itemize}

The rest of the paper is organized as follows. 
~\Sec{sec:background} provides a basic background about Micro Aerial Vehicles, the reasons for their prominence, and the challenges MAV system designers face.
\Sec{sec:MAV_cyber_physical_interaction} demonstrates the tight interaction between the cyber and physical processes of a MAV and introduces the ``cyber-physical interaction graph'' to capture how these two processes impact one another.
Architects simulators and benchmarks need to be updated to model such impacts. To this end, \Sec{sec:simulation} describes the MAVBench closed-loop simulation platform, and \Sec{sec:mavbench} introduces
the MAVBench benchmark suite and describes the computational kernels and full-system stack it implements.  \Sec{sec:evaluation} then describes our evaluation setup, and \Sec{sec:comp_mission_time},
\Sec{sec:comp_mission_energy}, and \Sec{sec:impact_holistic} use a combination of our analytical models, simulator, and benchmarks to dissect the impact of compute on MAVs. 
\Sec{sec:optimizations} presents two case studies exemplifying optimizations of the sort that system designers can exploit to improve mission time and energy, \Sec{sec:related} presents the related work, and finally, \Sec{sec:conclusion} summarizes and concludes the paper.

\section{Micro Aerial Vehicle Background}
\label{sec:background}

\begin{table}[b!]
\centering
\caption{UAVs by NATO Joint Air Competence Power~\cite{nato-uav-classification}.}
\label{nato-classification-drones}
\resizebox{.6\columnwidth}{!}{
\begin{tabular}{|c|c|c|c|}
\hline
Category                                                              & Weight (kg) & Altitude (ft) & Mission Radius (km) \\ \hline\hline
Micro                                                                 & \textless2                                             & \textless200                                                                    & 5                                                                           \\ \hline
Mini                                                                  & (2-20)                                                 & (200- 3000)                                                                     & 25                                                                          \\ \hline
Small                                                                 & (20-150)                                               & (3000-5000)                                                                    & 50                                                                          \\ \hline
Tactical                                                              & (150-600)                                              & (5000-10000)                                                                    & 2000                                                                        \\ \hline
Combat & \textgreater600                                        & \textgreater10000                                                               & Unlimited                                                                   \\ \hline
\end{tabular}
}
\end{table}

We provide a brief background on Micro Aerial Vehicles (MAVs), the most ubiquitous and growing segment of Unmanned Aerial Vehicles (UAVs). We then describe various subsystems that make up a MAV, and finally present the overall system level constraints facing MAVs.

\subsection{Micro Aerial Vehicles (MAVs)}
\label{sec:mav}

UAVs initially emerged as military weapons for missions in which having a human pilot would be a disadvantage~\cite{rs}. But since then there has been a recent proliferation of various other aerial vehicles for civilian applications including crop surveying, industrial fault detection, mapping, surveillance and aerial photography. There is no single established standard to categorize the wide range of UAVs. But Table~\ref{nato-classification-drones} shows one proposed classification guide provided by NATO. This classification is largely based on the weight of the UAV, and the mission altitude and range. 

In this paper, we focus on MAVs. A UAV is classified as a micro UAV if its weight is less than 2~\si{\kilo\gram}, and it operates within a radius of 5~\si{\kilo\meter}. MAVs' small size increases their accessibility and affordability by shortening their ``development and deployment time,'' and reducing the cost of ``prototyping and manufacturing''~\cite{Zhang2017207}. Also, their small size coupled with their ability to move flexibly empowers them with the agility and maneuverability necessary for these emerging applications.

MAVs come in different shapes and sizes. A key distinction is their wing type, ranging from fixed wing to rotary wing. Fixed wing MAVs, as their names suggest, have fixed winged airframes. Due to the aerodynamics of their wings, they are capable of gliding in the air, which improves their ``endurance'' (how long they last in the air). However, this also results in these MAVs typically requiring (small) runways for taking-off and landing. In contrast, rotor wing MAVs not only can take off and land vertically, but they can also move with more agility than their fixed-wing counterparts. 
They do not require constant forward airflow movement over their wings from external sources since they generate their own thrust using rotors. These capabilities enhance their benefits in constrained environments, especially indoors, where there are many tight spaces and obstacles. For many applications these benefits outweigh the cost of reduced endurance and as such rotor wing MAVs have become the dominant form of MAV. We focus on rotor based MAVs, specifically quadrotors. Nonetheless, the conclusions we draw from our studies apply other UAV categories as well.

\subsection {MAV Robot Complex}
\label{sec:MAV_subsystems}
\begin{figure}[t!]
\centering
\includegraphics[trim=0 0 0 -25, clip, width=.6\linewidth]{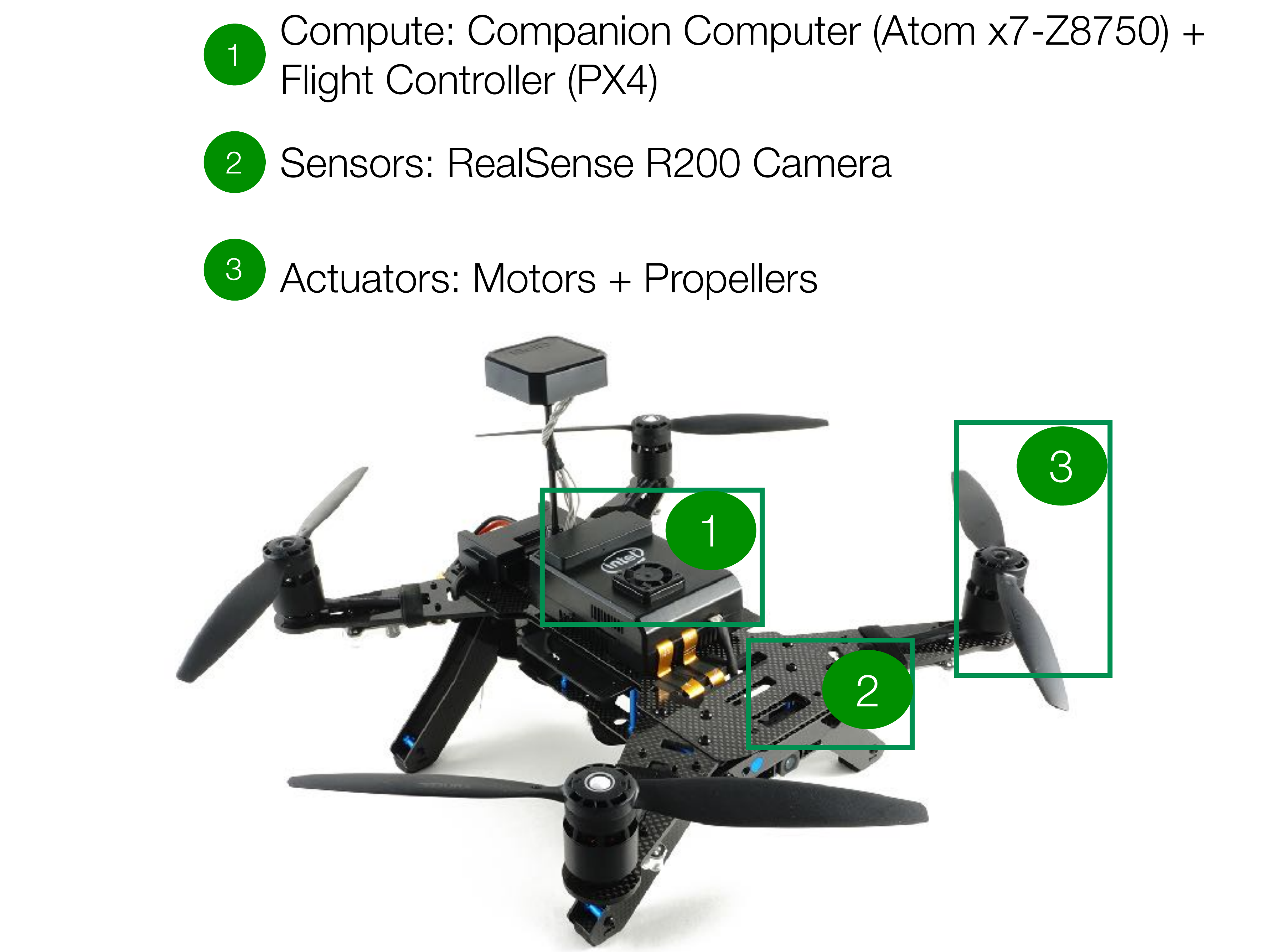}
\caption{MAV robot complex. The three main subsystems, i.e., compute, sensors and actuators, of an off-the-shelf Intel Aero MAV are shown. All other MAVs have a similar subsystem structure.}
\label{fig:drone_components}
\end{figure}

\begin{figure}[t!]
\centering
\includegraphics[trim=0 0 0 -25, clip, width=0.7\linewidth]{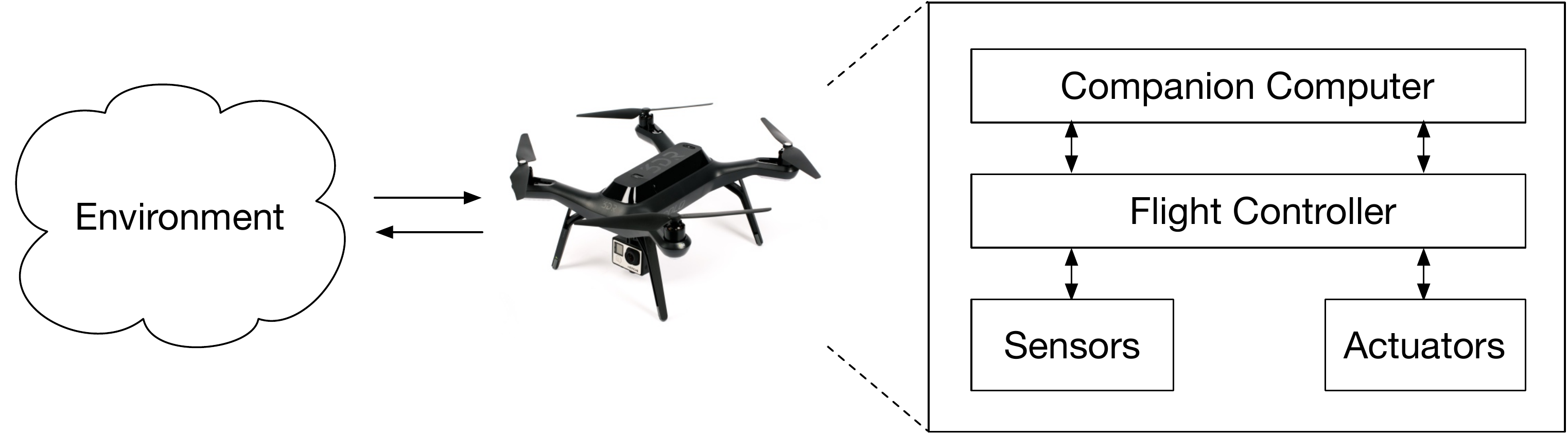}
\caption{Closed-loop data flow in a MAV. Information flows from sensors collecting environment data into the MAV's compute system, down into the actuators and back to the environment.}
\label{fig:Aerial_agent_data_flow}
\end{figure}

MAV's have three main subsystems that make up their robot complex: sensing, actuation, and compute, as shown in \Fig{fig:drone_components}. Similar to other cyber-physical systems, the design and development of MAVs requires an understanding of their composed and intertwined subsystems which we detail in this section. In these cyber-physical systems, the data flows in a (closed) loop, starting from the environment, going through the MAV and back to the environment, as shown in  \Fig{fig:Aerial_agent_data_flow}.

\paragraph{Sensors:} Sensors are responsible for capturing the state associated with the agent and its surrounding environment. 
To enable intelligent flights, MAVs must be equipped with a rich set of sensors capable of gathering various forms of data such as depth, position, and orientation. For example, \mbox{RGB-D} cameras can be utilized for determining obstacle distances and positions. The number and the type of sensors are highly dependent on the workload requirements and the compute capability of onboard processors which are used to interpret the raw data coming from the sensors.

\paragraph{Flight Controller (Compute):} The flight controller (FC) is an autopilot system responsible for the MAV's stabilization and conversion of high-level to low-level actuation commands. 
While they themselves come with basic sensors, such as gyroscopes and accelerometers, they are also used as a hub for incoming data from other sensors such as GPS and sonar. For command conversions, FCs take high-level flight commands such as``take-off" and lower them to a series of instructions understandable by actuators (e.g., current commands to electric motors powering the rotors). FCs use light-weight processors such as the ARM Cortex-M3 32-bit RISC core for the aforementioned tasks.    
 
\paragraph{Companion Computer (Compute):} The companion computer is a powerful compute unit, compared to the FC, that is responsible for the processing of the high level, computationally intensive tasks (e.g., computer vision). Not all MAVs come equipped with companion computers. Rather, these are typically an add-on option for more processing. NVIDIA's TX2 is a representative example with significantly more compute capability than a standard FC.

\paragraph{Actuators:} Actuators allow agents to react to their surroundings. They range from rather simple electric motors powering rotors to robotic arms capable of grasping and lifting objects. Similar to sensors, their type and number are a function of the workload and processing power on board.


\subsection{MAV Constraints}
\label{sec:constraints}


\begin{figure}[t!]
\centering
    \begin{subfigure}{.47\columnwidth}
    \centering
    \includegraphics[trim=0 0 0 0, clip, width=.7\columnwidth]{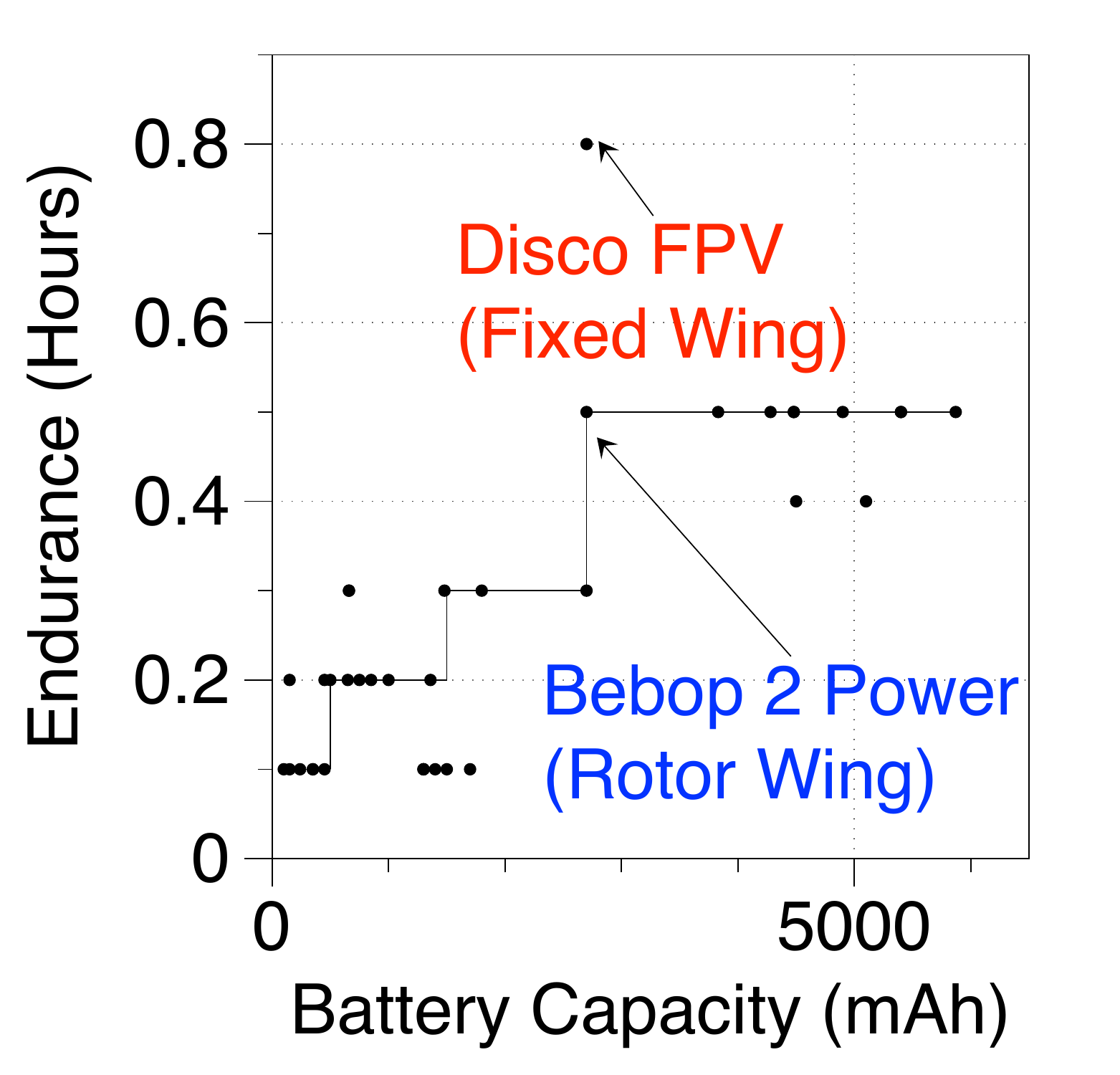}
   \caption{Endurance against battery capacity.}
    \label{fig:battery_capacity_vs_endurance}
    \end{subfigure}
    \begin{subfigure}{.47\columnwidth}
    \centering    
    \includegraphics[trim=0 0 0 0, clip, width=.7\columnwidth]{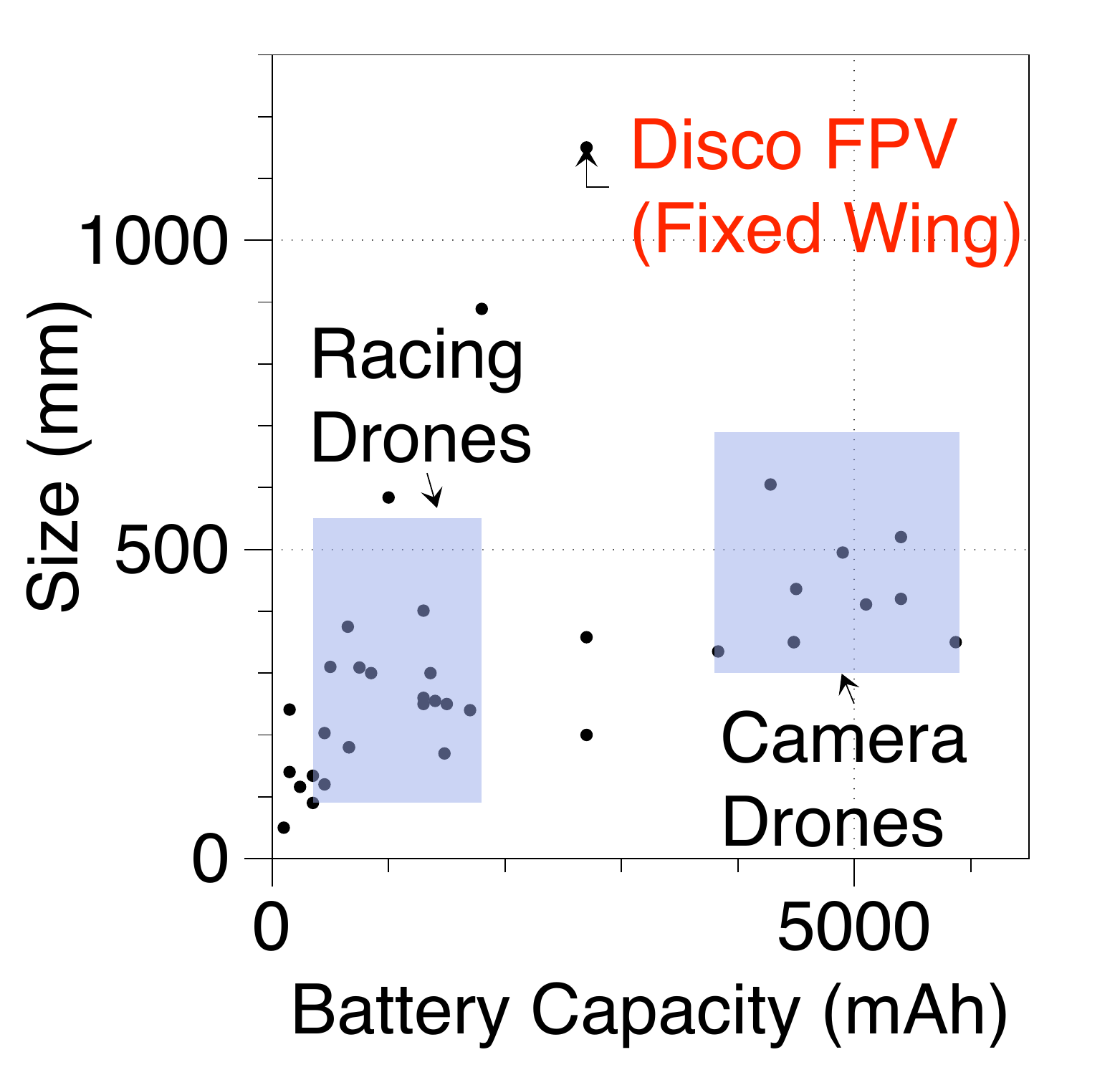}
    \caption{Drone size against total battery capacity.}
    \label{fig:battery_capacity_vs_size}
    \end{subfigure}
\caption{MAVs based on battery capacity and size. Endurance is important for MAVs to be useful in the real-world. However, their small size limits the amount of onboard battery capacity.}
\label{fig:tradeoff}
\end{figure}

A MAV's mechanical (propellers, payload, etc.) and electrical subsystems (motors and processors) constrain its operation and endurance, and as such present unique challenges for system architects and engineers. For example, when delivering a package, the payload size (i.e., the size of the package) affects the mechanical subsystem, requiring more thrust from the rotors and this, in turn, affects the electrical subsystem by demanding more energy from the battery source. Comprehending these constraints is crucial to understand how to optimize the system. The biggest of the constraints as they relate to computer system design are performance, energy, and weight.



\paragraph{Performance Constraints:} MAVs are required to meet various real-time constraints. For example, a drone flying at high speed looking for an object requires fast object detection kernels. Such a task is challenging in nature for large-sized drones that are capable of carrying high-end computing systems, and virtually impossible on smaller sized MAVs. Hence, the stringent real-time requirements dictate the compute engines that can be put on these MAVs.

\paragraph{Energy Constraints:} The amount of battery capacity on board plays an important role in the type of applications MAVs can perform. Battery capacity has a direct correlation with the endurance of these vehicles. To understand this relationship, we show the most popular MAVs available in the market and compare their battery capacity to their endurance. As \Fig{fig:battery_capacity_vs_endurance} shows, higher battery capacity translates to higher endurance. We see a step function trend, i.e., for classes of MAVs that has similar battery capacity, they have similar endurance. On top of this observation, we also see that for the same battery capacity, a fixed wing has longer endurance compared to rotor wing MAVs. For instance, in \Fig{fig:battery_capacity_vs_endurance}, we see that the Disco FPV (''Fixed wing'') has higher endurance compared to the Bebop 2 Power (''Rotor wing'') even though they have a similar amount of battery capacity.
We also note that the size of MAV also has a correlation with battery capacity as shown in \Fig{fig:battery_capacity_vs_size}.

\paragraph{Weight Constraints:} MAV weight, inclusive of its payload weight, can also have a significant impact on its endurance. Higher payload puts stress on the mechanical subsystems requiring more thrust to be generated by the rotors for hovering and maneuvering. This significantly reduces the endurance of MAVs. For instance, it has been shown that adding a payload of approximately 1.3~\si{\kilo\gram} reduces flight endurance by 4X~\cite{hasan2016sensorcloud}.

\section{A Cyber-Physical Perspective on MAVs}
\label{sec:MAV_cyber_physical_interaction}

MAVs are an integration of cyber and physical processes. 
A tight interaction of the two enables compute to control the physical actions of an autonomous MAV. 
Robot designers need to understand how such cyber and physical processes impact one another and ultimately, the robot's end-to-end behavior.
Furthermore, similar to cross-compute layer optimization approach widely adopted by the compute system designers,
robot designers can improve the robot's optimality by adopting a robot's cross-system (i.e., cross cyber and physical) optimization methodology and co-design.

To this end, we introduce the \csig, a directed acyclic graph that captures how different subsystems of a robot impact the mission metrics through various cyber and physical quantities.
We familiarize the reader with the graph (using a simple example) and move onto presenting what the graph looks like for a complicated MAV robot. Next, looking through the lens of this graph, we provide a brief example of how a subsystem such as compute can impact a mission metric, and finally discuss the need for new tools to investigate these impacts in details.

\subsection{Cyber-Physical Interaction Graph}

A \csig has four components to it. It has (1) a robot complex, (2) cyber-physical quantities, (3) impact functions, and (4) mission metrics. Subsystems in the robot complex have either cyber and/or physical quantities that impact one another that are captured in the graph, which can ultimately affect mission metrics such as mission time or energy consumption.

\begin{figure}[]
\centering
\includegraphics[trim=0 0 0 0, clip, width=1\linewidth]
{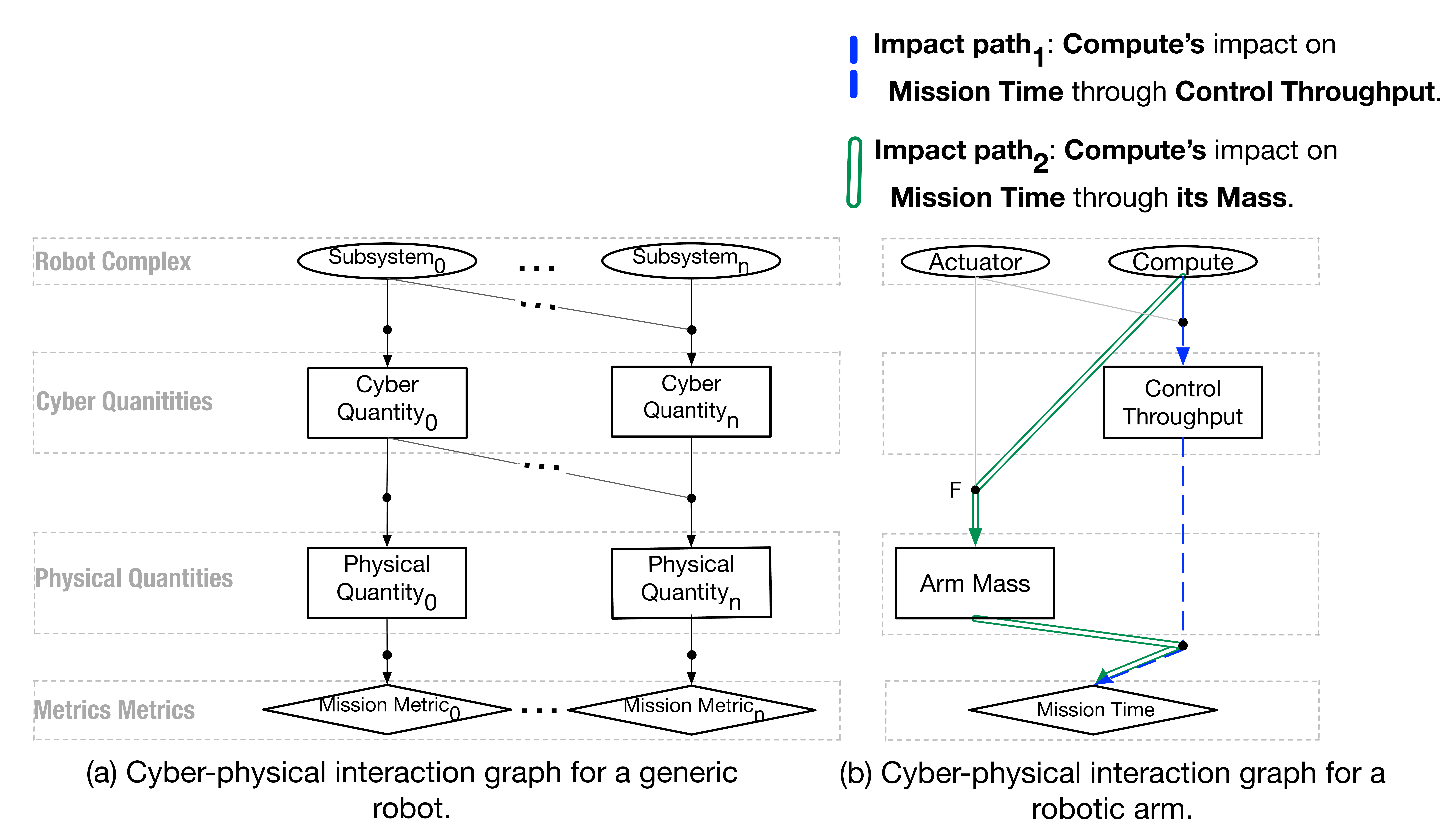}
\caption{Cyber-physical interaction graph. This graph captures how the various subsystems of a robot impact mission metrics through cyber and physical quantities that interact with one another.}
\label{fig:CIG_general}
\end{figure}

Figure ~\ref{fig:CIG_general}a shows a generic \csig. The graph consists of a set of edges and vertices. The subsystems making up the robot complex are denoted using ellipses. The mission
metrics specifying the metrics developers use to measure the mission's success are shown using diamonds. The cyber-physical quantities specifying various quantities that determine the behavior of the
robot are shown using rectangles. The impact functions, capturing the impact of one quantity on another and further on the mission metrics, are shown using contact points (filled black circles when two or more edges cross). The edges in the graph imply the existence and the direction of the impact. 

To investigate the impact of one vertex on another, such as compute and mission time, 
we need to examine all the paths originating from the first vertex (compute) and ending with the second vertex (mission time). We call each one of these paths ``impact paths.''

We use a toy example of a simple robot arm (Figure ~\ref{fig:CIG_general}b) to familiarize the reader with the graph.  
Our robotic arm has two subsystems, namely a compute and an actuation subsystem. These subsystems impact mission time, i.e., the time it takes for the
robot to relocate all the boxes, through a cyber quantity such as control throughput and a physical quantity such as arm's mass.
The green-color/double-sided and blue-color/coarse-grained-dashed paths show two paths that compute impacts mission time.
Intuitively speaking, through the blue-color/coarse-grained-dashed path, compute impacts controller's throughput and hence the robot's rotation speed. This, in return, impacts mission time.
Through the green-color/double-sided path, compute impacts the mass of the robot and hence dictating the speed and ultimately, the mission time. Note that the impact function, shown with the marker F in Figure ~\ref{fig:CIG_general}b, is simply an addition
function since the robot's overall mass is the aggregation of the compute and actuation subsystem mass.

\subsection{MAV Cyber-Physical Interaction Graph}
\blueDebug{par summary:SIG for our quad. Exemplify two paths with a short description. Explain direct/indirect interactions. outro to case studies}

\begin{figure}[!t]
\includegraphics[trim=0 0 0 0, clip, width=.8\linewidth]
{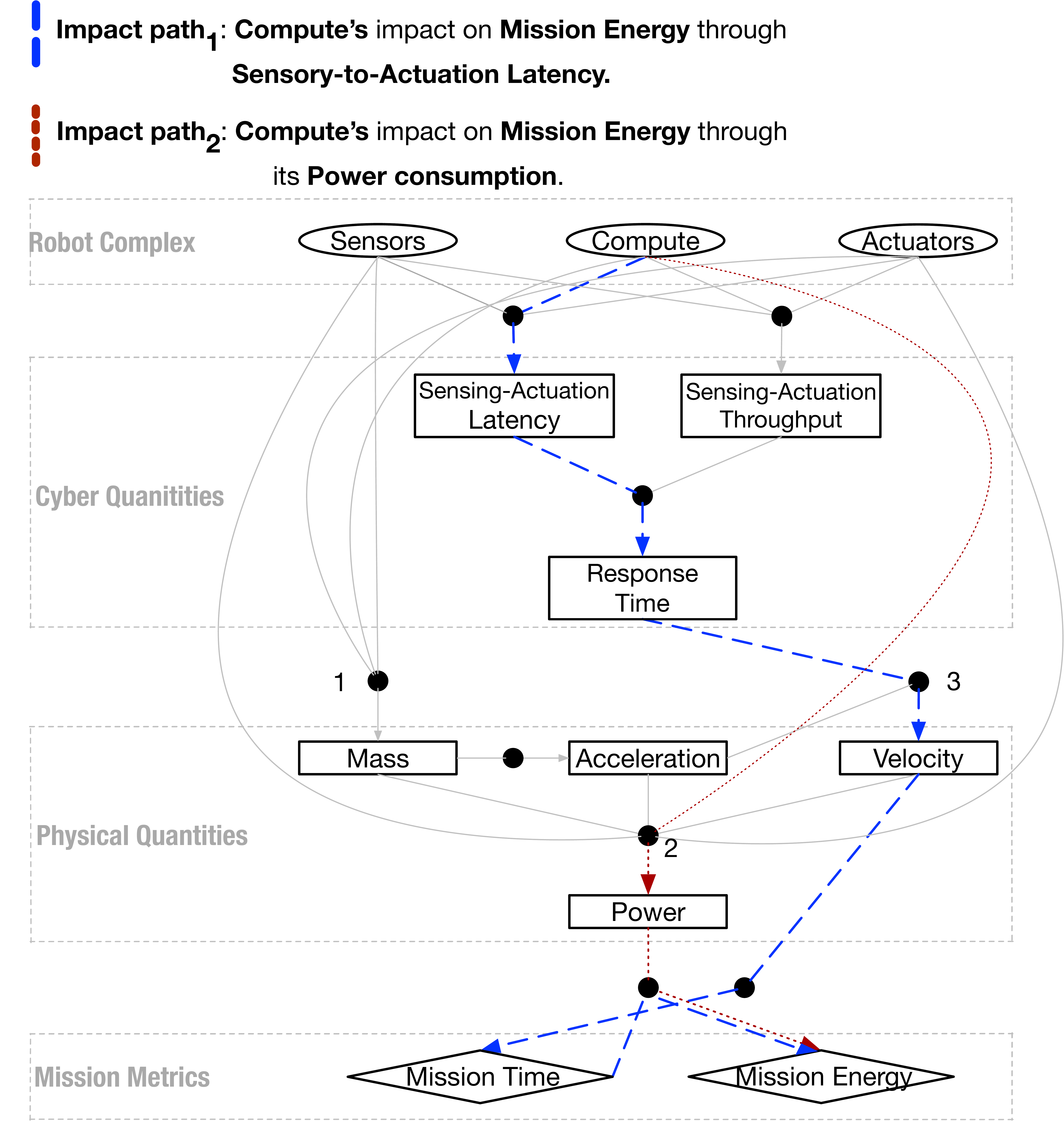}
\caption{Cyber-physical interaction graph for our quadrotor MAV with some path examples.}
\label{fig:CIG_MAV}
\end{figure}

We apply the \csig to our quadrotor MAV. The quadrotor consists of three subsystems, namely the sensory, actuation, and compute subsystem. Figure ~\ref{fig:CIG_MAV} illustrates the MAV's \csig broken down into the four major subcomponents.

We focus on three cyber quantities, i.e, \textit{sensing-to-actuation latency} (Sensing-Actuation Latency in the graph), \textit{sensing-to-actuation throughput} (Sensing-Actuation Throughput in the
graph), and \textit{Response Time}. Sensing-to-actuation latency is the time the drone takes to sample sensory data and process them to issue flight commands ultimately. 
Sensing-to-actuation throughput is the rate with which the drone can generate the aforementioned (and new) flight commands. Response time is the time the drone takes 
to respond to an event emerging (e.g., the emergence of an obstacle in the drone's field of view) 
in its surrounding environment. 

For the physical quantities, we focus on motion dynamic/kinematic related quantities such as mass, i,e, the total mass of the drone, it's acceleration, and velocity. This is because they impact our mission metrics. For instance, an increase in mass can decrease acceleration, which translates to more power demands from the rotors, and that ultimately increases the overall mission energy consumption. 

For mission metrics, we focus on time and energy. These metrics are chosen due to their importance to the mission success. Reducing mission time is of utmost importance for most applications such as package delivery, search and rescue, scanning, and others. Also, reduction in energy consumption is valuable as a drone that is out of battery is unable to finish its mission. 

The impact functions range from simple addition (marker 1 in Figure~\ref{fig:CIG_MAV}) to more complex linear functions (marker 2) to non-linear relations (marker 3). 

Note that although the MAV \csig presented in this paper does not contain all the possible cyber and physical quantities associated with a MAV, we have included the ones that have the most significant impact on our mission metrics.

\subsection{Examining the Role of Compute Using the Cyber-physical Interaction Graph}
\label{sec:role_of_comp_intro}

Compute plays a crucial role both in the overall mission time and total energy consumption of a MAV in different ways, which we refer to as impact paths. Figure \ref{fig:CIG_MAV}, highlights two paths that can influence mission energy. We briefly explain these to give the reader an intuition for how compute affects MAV's mission metrics, deferring the details until later for discussion.

Through one path,  the impact is positive (i.e., lowering the energy consumption and hence saving battery) while through the other, the impact is negative (i.e., increasing the energy
consumption). In the positive impact path, i.e., the blue-color/coarse-grained-dashed path, compute can reduce the mission energy. This is because a platform with more compute capability reduces a
cyber quantity, such as sensing-to-actuation latency and response time. This allows the drone to respond to its environment faster and in return, increase a physical quantity like its velocity. By flying faster, the drone finishes its mission faster and so reduces a mission metric such as its total mission energy. 

Looking through the lens of another path, the impact is negative (i.e., energy consumption increases). In the negative impact path, i.e., the red-color/fine-grained-dashed path, a more compute capable platform has a negative impact on the mission energy because it consumes more power. 

We count a total of nine impacts paths originating with compute and ending with mission time and energy. This paper quantitatively examines all such paths where the cyber and physical quantities
impact one another dictating the drone's behavior. At first in sections ~\ref{sec:comp_mission_time} and ~\ref{sec:comp_mission_energy},
we investigate them in isolation to gain a better insight into the underlying concepts, and then in section ~\ref{sec:impact_holistic}, we put them all together for a holistic examination. 
Overall, we see that an increase in compute can positively impact (reduce) mission time and energy
by improving cyber quantities such as sensing-to-actuation throughput and latency; however,
an increase in compute can negatively impact the mission time and energy through increasing physical quantities such as quad's mass and power.

Investigating the cyber and physical interactions of the sorts mentioned above requires new tools. This is due to the numerous differences between cyber-physical systems and their more traditional counterparts (i.e., desktops, servers, smartphones, etc.).
Such differences need to be appreciated, and the architects' tool sets need to be adjusted accordingly. In this paper, we mainly focus on two major difference, namely (1) continuous interaction of the system with its complex and unpredictable surrounding environment, an aspect that is void in traditional systems, and (2) a closed-loop data-flow. 

To enable various system design research and development, we provide a simulator (Section~\ref{sec:simulation}) and a benchmark suite (Section ~\ref{sec:mavbench}) to model the MAV-environment close interactions. Furthermore, the environments' complexity is captured with high fidelity using a game engine. And finally, the closed-loop data flow nature of these systems are modeled using hardware in the loop simulator. 
In the next two sections, we discuss each of these tools in detail. It is worth noting that although this paper mainly focuses on Micro Aerial Vehicles (MAVs), the generality of our simulation framework allows for the investigation of other autonomous machines (e.g., AirSim now supports cars as well). With this, we hope to systematically bootstrap a collaboration between the robotics and system design community---an opportunity for domain-specific architecture specialization. 


\section{Closed-loop Simulation}
\label{sec:simulation}
\setstcolor{red}
In this section, we present a closed-loop simulation environment for simulating and studying MAVs. We show how our setup captures MAV robot complex, i.e., MAV subsystems and their components and further their interactions in a closed-loop setup. We describe the knobs that our simulator supports to enable exploratory studies for cyber-physical co-design.
We also describe how the simulator models mission metrics such as energy consumption, in addition to functionality and performance.

\subsection{Simulation Setup}
\label{sec:setup}
\begin{figure}[t!]
\centering
\includegraphics[trim= 10 10 10 10, clip, width=0.75\columnwidth]{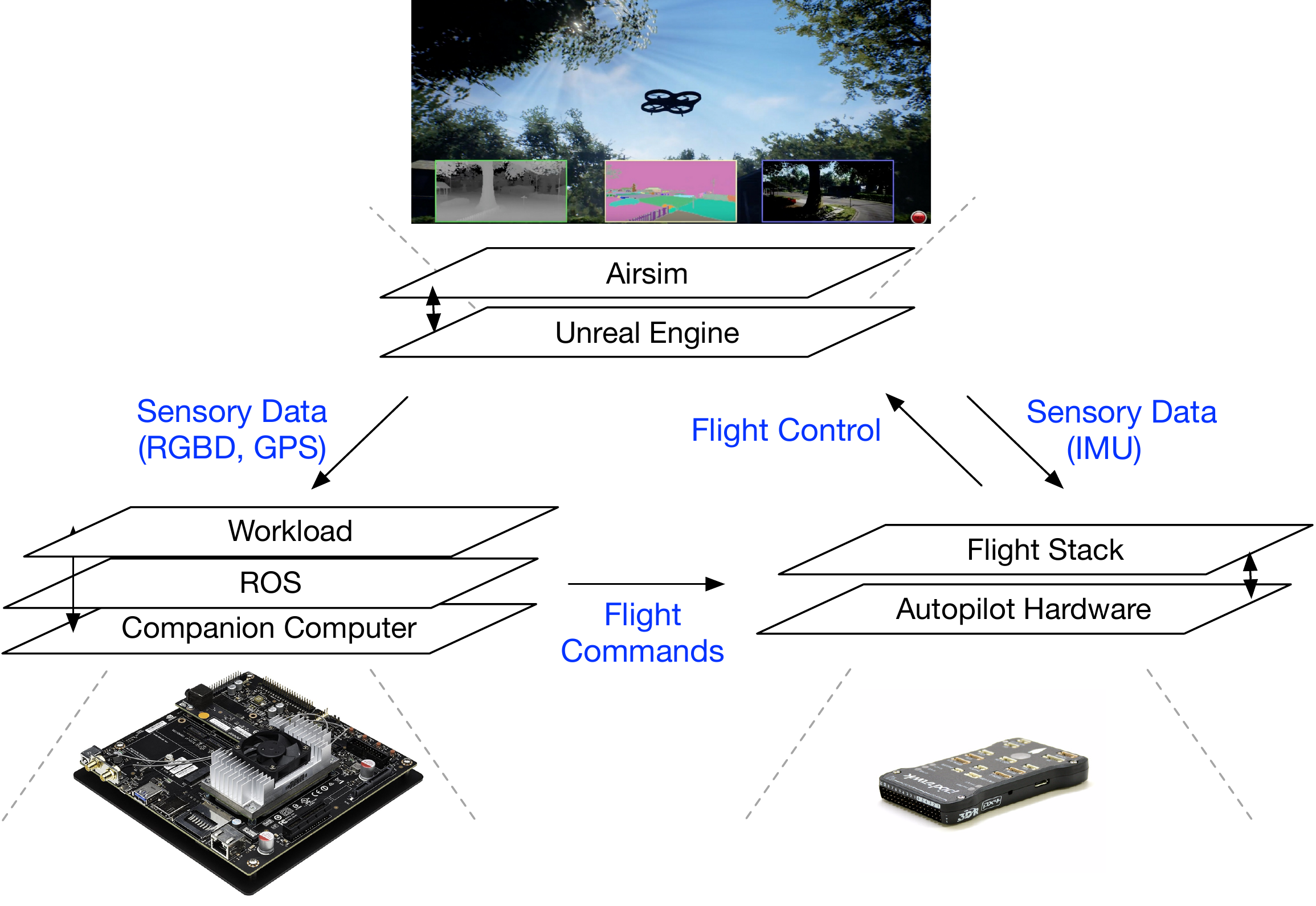}
\caption{Architectural overview of our closed-loop simulation. 
}
\label{fig:end-to-end}
\end{figure}

Closed-loop operation is an integral component of autonomous MAVs. As described previously in \Sec{sec:background}, in such systems, the data flow in a (closed) loop, starting from the environment, going through the MAV and back to the environment, as shown in \Fig{fig:Aerial_agent_data_flow}. The process involves sensing the environment (Sensors), interpreting it and making decisions (Compute), and finally navigating within or modifying the
environment (Actuators) in a loop. In this section, 
we show how our simulation setup, shown in \Fig{fig:end-to-end}, maps to the various components corresponding to a MAV robot complex. Furthermore, we discuss the simulator's ability to capturing various cyber and physical quantities. 

\paragraph{Environments, Sensors and Actuators:} Environments, sensors and actuators are simulated with the help of a game engine called Unreal~\cite{GameEngi70:online}. With a physics engine at its heart, it ``provides the ability to perform accurate collision detection as well as simulate physical interactions between objects within the world''~\cite{PhysicsS8:online}. Unreal provides a rich set of environments such as mountains, jungles, urban setups, etc. to simulate.

To simulate MAV's dynamics and kinematics, we used AirSim, an open-source Unreal based plug-in from Microsoft~\cite{Airsim:online,Airsim_paper}. Through AirSim we can study the impact of drone's physical quantities such as velocity and acceleration. We limit our sensors and actuators to the ones realistically deployable by MAVs, such as \mbox{RGB-D} cameras and IMUs. Unreal and Airsim run on a powerful computer (host) capable of physical simulation and rendering. Our setup uses an Intel Core i7 CPU and a high-end NVIDIA GTX 1080 Ti GPU.  

\paragraph{Flight Controller:} AirSim supports various flight controllers that can be either hardware-in-the-loop or completely software-simulated. For our experiments, we chose the default software-simulated flight controller provided by AirSim. However, AirSim also supports other FCs, such as the Pixhawk~\cite{Pixhawk:online}, shown in black in \Fig{fig:end-to-end} which runs the PX4~\cite{PX4Archi7:online} software stack. AirSim supports any FC which can communicate using MAVLINK, a widely used micro aerial vehicle message marshaling library~\cite{mavlinkm68:online}. 

\paragraph{Companion Computer:} We used an NVIDIA Jetson TX2~\cite{TX2}, a high-end embedded platform from Nvidia with 256 Pacal CUDA cores GPU and a Quad ARM CPU; however, the flexibility of our setup allows for swapping this embedded board with others such as x86 based Intel Joule~\cite{joule:online}. TX2 communicates with Airsim and also FC via Ethernet. Note that the choice of the companion computer influences both cyber and physical quantities such as response time and compute mass. 

\paragraph{ROS:} Our setup uses the popular Robot Operating System (ROS) for various purposes such as low-level device control and inter-process communication~\cite{ROSorgPo80:online}.
Robotic applications typically consist of many concurrently-running processes that are known as ``nodes.'' For example, one node might be responsible for navigation, another for localizing the agent and a third for object detection. ROS provides peer-to-peer communication between nodes, either through blocking ``service'' calls, or through non-blocking FIFOs (known as the Publisher/Subscriber paradigm). 

\paragraph{Workloads:} Our workloads runs within the ROS runtime on TX2. Briefly, we developed five distinct workloads, each representing a real world usecase: agricultural scanning, aerial photography, package delivery, 3D mapping and search and rescue. They are extensively discussed in \Sec{sec:benchmarks}.

\paragraph{Putting It All Together:} To understand the flow of data, we walk the reader through a simple workload where the MAV is tasked to detect an object and fly toward it. The object (e.g., a person) and its environment (e.g., urban) are modeled in the Unreal Engine. As can be seen in \Fig{fig:end-to-end}, the MAV's sensors (e.g., accelerometer and \mbox{RGB-D} Camera), modeled in Airsim, feed their data to the flight controller (e.g., physics data to PX4) and the companion computer (e.g., visual and depth to TX2) using MAVLink protocol. The kernel (e.g., object detection), running within the ROS runtime environment on the companion computer, is continuously invoked until the object is detected. Once so, flight commands (e.g., move forward) are sent back to the flight controller, where they get converted to a low-level rotor instruction stream flying the MAV closer to the person. 

\subsection{Simulation Knobs and Extensions}
\label{sec:knobs}

With the help of Unreal and AirSim, our setup exposes a wide set of knobs. Such knobs enable the study of agents with different characteristics targeted for a range of workloads and conditions. For different environments, the Unreal market provides a set of maps free or ready for purchase. Furthermore, by using Unreal programming, we introduce new environmental knobs, such as (static) obstacle density, (dynamic) obstacle speed, and so on. In addition, Unreal and AirSim allow for the MAV and its sensors to be customized. For example, the cameras' resolution, their type, number, and positions all can be tuned according to the workloads' need.   

Our simulation environment can be extended. For the compute on edge, the TX2 can be replaced with other embedded systems or even micro-architectural simulators, such as gem5. Sensors and actuators can also be extended, and various noise models can be introduced for reliability studies. 

\subsection{Energy Simulation and Battery Model}
\label{sec:energy}
We extended the AirSim simulation environment with an energy and a battery model to collect mission energy data in addition to mission time. Our energy model is a function of the velocity and acceleration of the MAV~\cite{energyaware}.  The higher the velocity or acceleration, the higher the amount of energy consumption. Velocity and acceleration values are sampled continuously, their associated power calculated and integrated for capturing the total energy consumed by the agent.

\newcommand{\norm}[1]{\left\lVert#1\right\rVert}

We used a parametric power estimation model proposed in \cite{3DR-energy-model}. The formula for estimating power $P$ is described below:%
\begin{equation}
\begin{aligned}
P = \begin{bmatrix}
        \beta_{1} \\
        \beta_{2} \\
        \beta_{3}
    \end{bmatrix}^{T}
    \begin{bmatrix}
        \norm{\vec{v}_{xy}} \\
        \norm{\vec{a}_{xy}} \\
        \norm{\vec{v}_{xy}}\norm{\vec{a}_{xy}}
    \end{bmatrix}
    +
    \begin{bmatrix}
        \beta_{4} \\
        \beta_{5} \\
        \beta_{6}
    \end{bmatrix}^{T}
    \begin{bmatrix}
        \norm{\vec{v}_{z}} \\
        \norm{\vec{a}_{z}} \\
        \norm{\vec{v}_{z}}\norm{\vec{a}_{z}}
    \end{bmatrix}
    +
    \begin{bmatrix}
        \beta_{7} \\
        \beta_{8} \\
        \beta_{9}
    \end{bmatrix}^{T}
    \begin{bmatrix}
        m \\
        \vec{v}_{xy} \cdot \vec{w}_{xy} \\
        1
    \end{bmatrix}
\end{aligned}
\label{eqn:power}
\end{equation}

In the Equation~\ref{eqn:power}, $\beta_{1}$, ..., $\beta_{9}$ are constant coefficients determined based on the simulated drone. $\vec{v}_{xy}$ and $\vec{a}_{xy}$ are the horizontal speed and acceleration vectors whereas  $\vec{v}_{z}$ and $\vec{a}_{z}$ are the corresponding vertical values. $m$ is the mass and $\vec{w}_{xy}$  is the vector of wind movement. 

We have a battery model that implements a coulomb counter approach~\cite{coulomb-counter}. The simulator calculates how many coulombs (product of current and time) have passed through the drone's battery over every cycle. This is done by calculating the power and the voltage associated with the battery. The real-time voltage is modeled as a function of the percentage of the remaining coulomb in the battery as described in ~\cite{battery-model}. \Sec{sec:comp_mission_energy}
 presents experimental results for a 3DR Solo MAV.

\subsection{Simulation Fidelity and Limitations}
\label{sec:accuracy}

The fidelity of our end-to-end simulation platform is subject to different sources of error, as it is with any simulation setup. The major obstacle is the \textit{reality gap}---i.e., the difference between the simulated experience and the real world. This has always posed a challenge for robotic systems. The discrepancy results in difficulties where the system developed via simulation does not function identically in the real world. 
To address the reality gap, we iterate upon our simulation components and discuss their fidelity and limitations. Specifically, this involves (1) simulating the environment, (2) modeling the drone's sensors and flight mechanics, and last but not least (3) evaluating the compute subsystem itself.

First, the Unreal engine provides a high fidelity environment. By providing a rich toolset for lighting, shading, and rendering, photo-realistic virtual worlds can be created. In prior work~\cite{unrealcv}, authors examine photorealism by running a Faster-RCNN model trained on PASCAL in an Unreal generated map. The authors show that object detection precision can vary between 1 and 0.1 depending on the elevation and the angle of the camera. Also, since Unreal is open-sourced, we programmatically emulate a range of real-world scenarios. For example, we can set the number of static obstacles and vary the speed of the dynamic ones to fit the use case.   

Second, AirSim provides high fidelity models for the MAV, its sensors, and actuators. Embedding these models into the environment in a real-time fashion, it deploys a physics engine running with 1000~\si{\hertz}.  As the authors discuss in~\cite{Airsim_paper}, the high precision associated with the sensors, actuators, and their MAV model, allows them to simulate a Flamewheel quadrotor frame equipped with a Pixhawk~v2 with little error. 
Flying a square-shaped trajectory with sides of length 5~\si{\meter} and a circle with a radius of 10~\si{\meter},  AirSim achieves  0.65~\si{\meter} and 1.47~\si{\meter} error, respectively. Although they achieve high precision, the sensor models, such as the ``camera lens models,'' ``degradation of GPS signal due to obstacles,'' ``oddities in camera,'' etc. can benefit from further improvements. 

Third, as for the compute subsystem itself, our hardware has high fidelity since we use off-the-shelf embedded platforms for the companion computer and flight controller. As for the software, ROS is widely used and adopted as the \textit{de facto} middleware software in the robotics research community.

\section{Benchmark Suite} 
\label{sec:mavbench}

To quantify the energy and performance demands of typical MAV applications and understand the cyber-physical interactions, we created a set of workloads that we compiled into a benchmark suite.
By combining this suite with our simulation setup, we get to study the robot's end-to-end behavior from both cyber and physical perspective, and further investigate various compute optimization techniques for MAV applications.
Our benchmarks run on top of our closed-loop simulation environment. 


Each workload is an \textit{end-to-end} application that allows us to study the kernels' impact on the whole application as well as to investigate the interactions and dependencies between kernel. 
By providing holistic end-to-end applications instead of only focusing on individual kernels, MAVBench allows for the examination of kernels' impacts and their optimization at the application level. This is a lesson learned from Amdahl's law, which recognizes that the true impact of a component's improvement needs to be evaluated globally rather than locally.

The MAVBench workloads have different computational kernels, as shown in Table~\ref{kernel_makeup}. MAVBench aims at being comprehensive by (1) selecting applications that target different robotic domains (robotics in hazardous areas, construction, etc.) and (2) choosing kernels (e.g., point cloud, RRT) common across a range of applications, not limited to our benchmark-suite. The computational kernels (OctoMaps, RTT, etc.) that we use in the benchmarks are the building blocks of many robotics applications, and hence, they are platform agnostic. We present a high-level software pipeline associated (though not exclusive) to our workloads. Then, we provide functional summaries of the workloads in MAVBench, their use cases, and mappings from each workload to the high-level software pipeline. We describe in detail the prominent computational kernels that are incorporated into our workloads. Finally, we provide a short discussion regarding the Quality-of-Flight (QoF) metrics with which we can evaluate MAV applications success and further the role of compute.

 \begin{figure*}[t]
\centering
\includegraphics[height=1.65in, keepaspectratio]{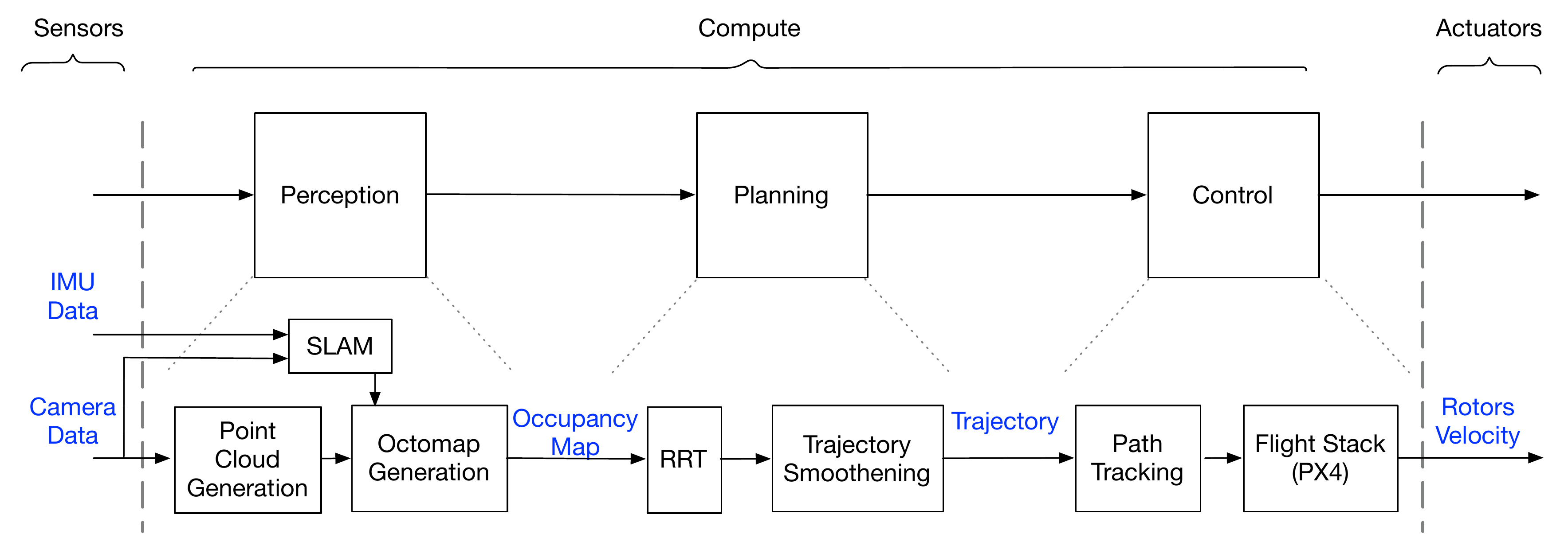}
\caption{High-level application pipeline for a typical MAV application. The upper row presents a universal pipeline that all our MAVBench applications follow, which involves \emph{perception}, \emph{planning} and \emph{control}. The lower row presents how a specific workload in MAVBench (e.g. package delivery) maps to the universal high-level application pipeline.}
\label{fig:software_pipeline}
\end{figure*}

\subsection{Workloads and Their Data Flow} \label{sec:benchmarks}
The benchmark suite consists of five workloads, each equipped with the flexibility to configure its computational kernel composition (described later in \Sec{sec:kernels}).
The following section sheds light on the high-level data flow governing all the applications, each application's functional summary, and finally, the inner workings of these workloads as per the three-stage high-level application pipeline. 

There are three fundamental processing stages in each application: \emph{Perception}, \emph{Planning} and \emph{Control}. In the perception stage, the sensory data is processed to extract relevant states from the environment and the drone. This information is fed into the next two stages (i.e., planning and control). Planning ``plans'' flight motions and forwards them to the actuators in the control subsystem. \Fig{fig:software_pipeline} summarizes this high-level software pipeline, which each of our workloads embodies.

\textbf{Perception:}
It is defined as ``the task-oriented interpretation of sensor data''~\cite{Handbook_robotic}. Inputs to this stage, such as sensory data from cameras or depth sensors, are fused to develop an elaborate model in order to extract the MAV's and its environment's relevant states (e.g., the positions of obstacles around the MAV). This stage may include tasks such as Simultaneous Localization and Mapping (SLAM) that enables the MAV to infer its position in the absence of GPS data.

\textbf{Planning:}
 Planning generates a \textit{collision-free} path to a target using the output of the perception (e.g., an occupancy map of obstacles in the environment). In short, this step involves first generating a set of possible paths to the target, such as by using the probabilistic roadmap (PRM) algorithm and then choosing an optimal one among them using a path-planning algorithm, such as A*.

\textbf{Control:}
This stage is about following the desired path, which is absorbed from the previous stage while providing a set of guarantees such as feasibility, stability, and robustness~\cite{tech_problem}. In this stage, the MAV's kinematics and dynamics are considered, such as by smoothening paths to avoid high-acceleration turns, and then, finally, the flight commands are generated (e.g., by flight controllers such as the PX4) while ensuring the aforementioned guarantees are still respected.

\begin{figure*}[t]
    \centering
    \begin{subfigure}[t]{1.385in}
        \centering
        \includegraphics[width=\textwidth, height=1in]{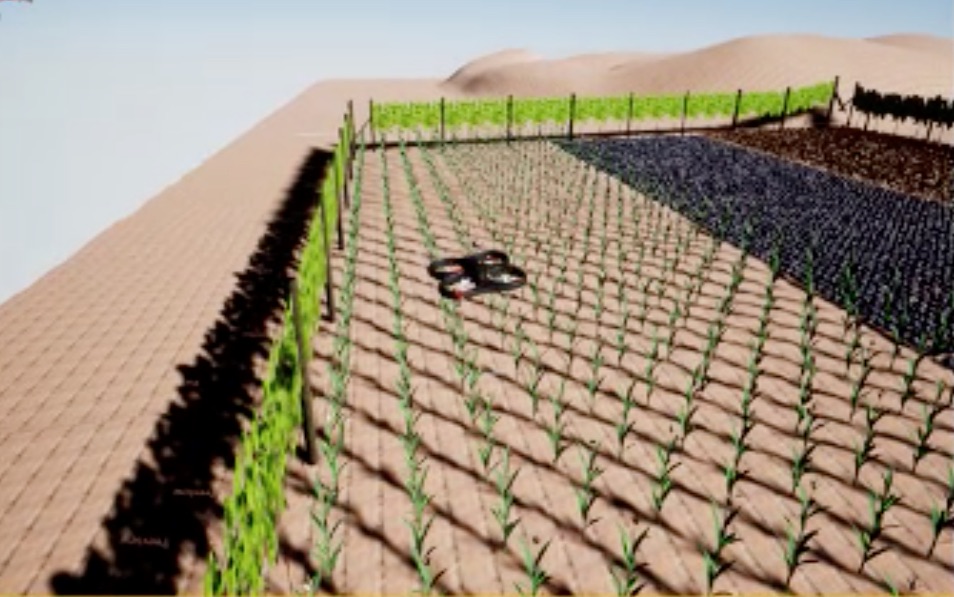}
        \caption{Scanning.}\label{fig:benchmarks:scanning}
    \end{subfigure}
    \hfill
    \begin{subfigure}[t]{1.385in}
        \centering
        \includegraphics[width=\textwidth, height=1in]{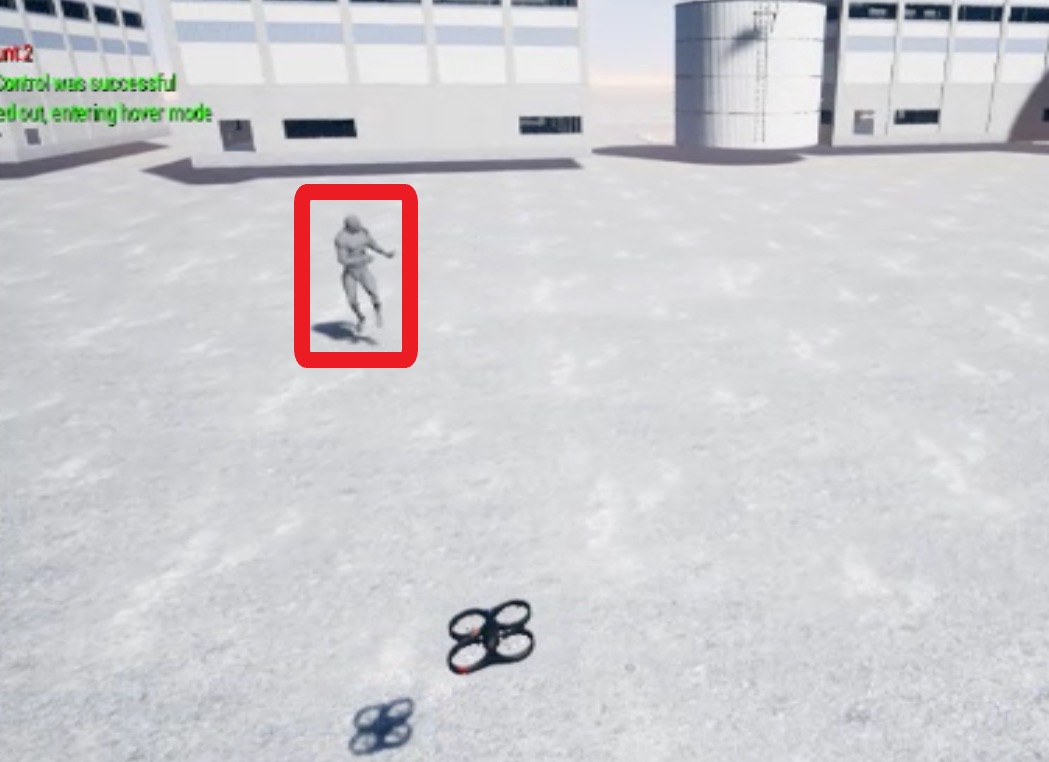}
        \caption{Aerial Photography.}\label{fig:benchmarks:aerial-photo}
    \end{subfigure}
    \hfill
    \begin{subfigure}[t]{1.385in}
        \centering
        \includegraphics[width=\textwidth, height=1in]{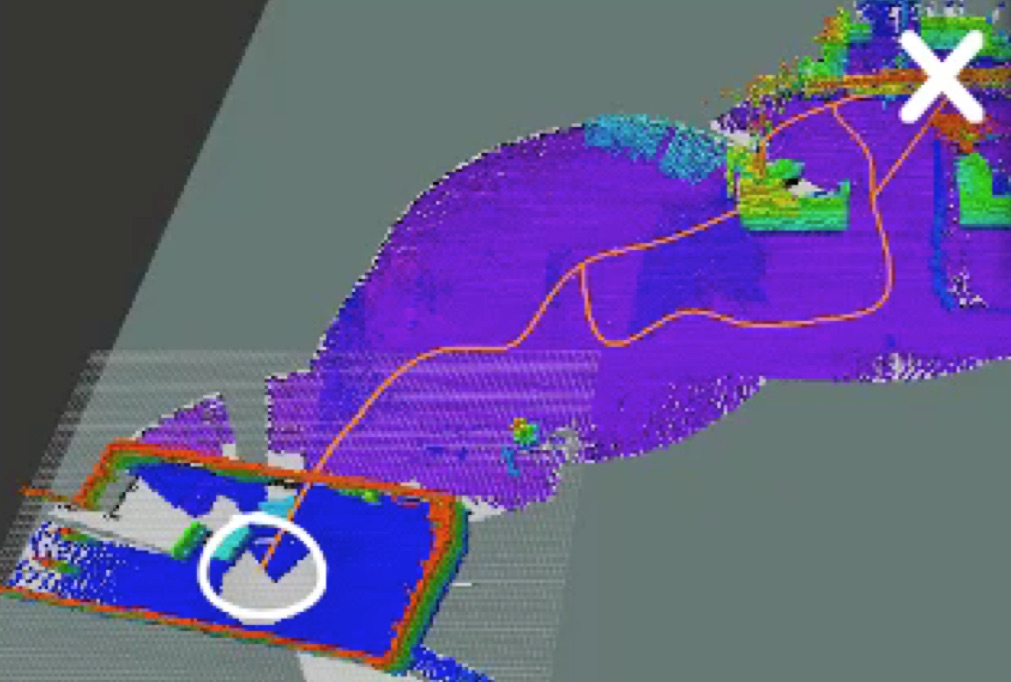}
        \caption{Package Delivery.}\label{fig:benchmarks:package-delivery}
    \end{subfigure}
    \hfill
    \begin{subfigure}[t]{1.385in}
        \centering
        \includegraphics[width=\textwidth, height=1in]{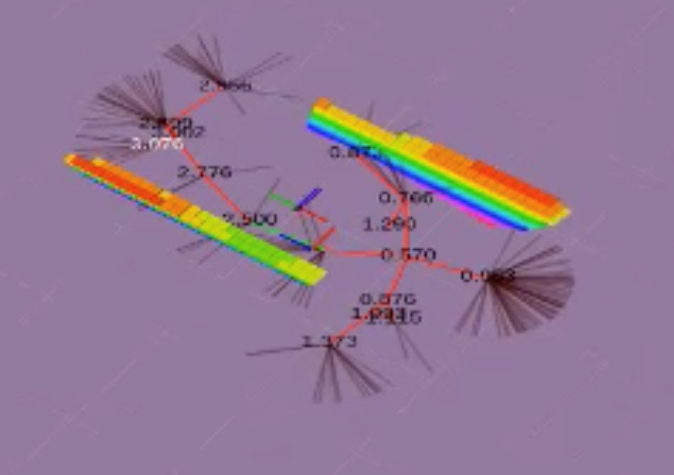}
        \caption{3D Mapping.}\label{fig:benchmarks:3D-mapping}
    \end{subfigure}
    \hspace{28pt}
    \begin{subfigure}[t]{1.385in}
        \includegraphics[width=\textwidth, height=1in]{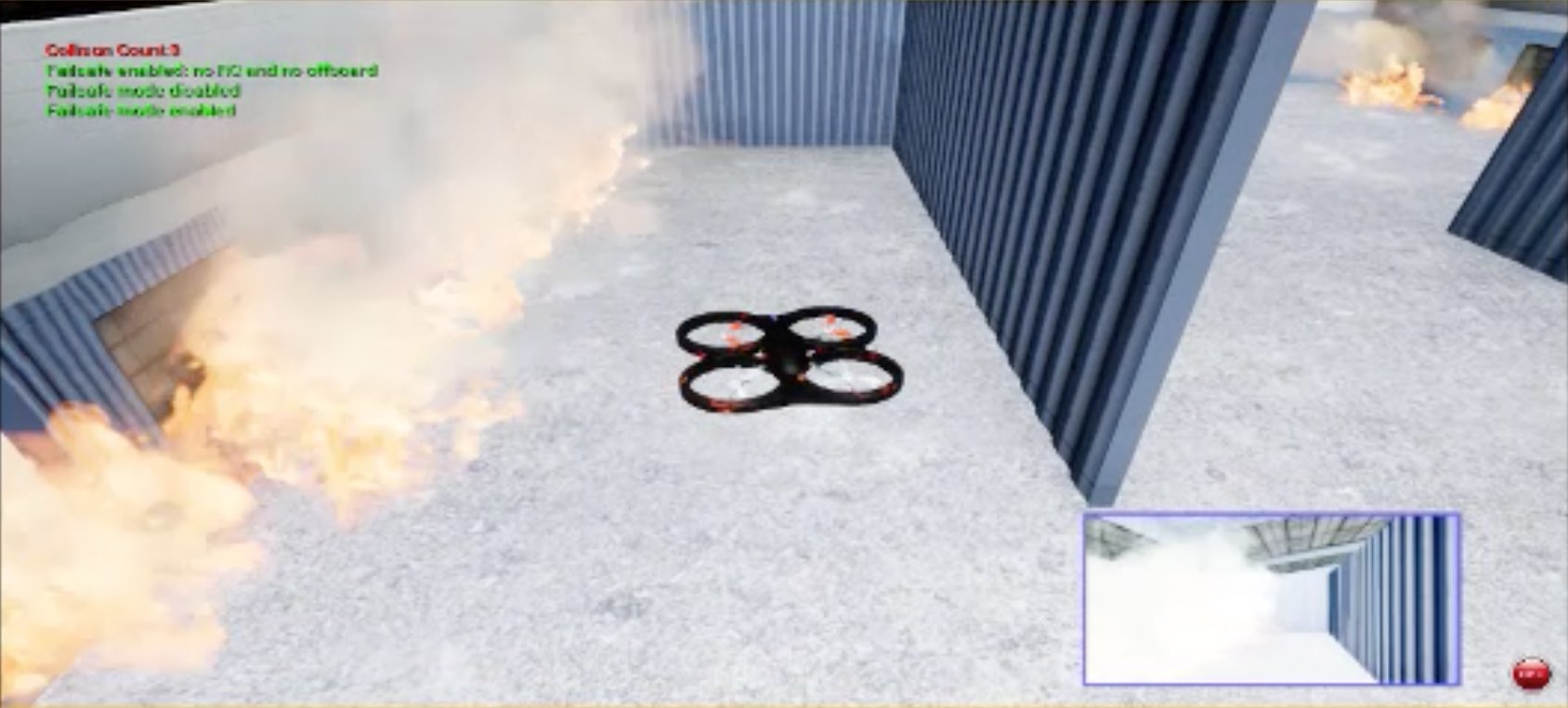}
        
        \caption{Search and Rescue.}\label{fig:benchmarks:search-and-rescue}
    \end{subfigure}
    \caption{MAVBench workloads. Each workload is an end-to-end application targeting both industry and research use cases. All figures are screenshots of a MAV executing a workload within its simulated environment. Fig.~\ref{fig:benchmarks:package-delivery} shows a MAV planning a trajectory to deliver a package. Fig.~\ref{fig:benchmarks:3D-mapping} shows a MAV sampling its environment in search of unexplored areas to map.}
    \label{fig:bench_screenshot}
\end{figure*}

\begin{figure}[t!]
    \centering
    \vspace{-54pt}
    \begin{subfigure}[t]{.725\columnwidth}
        \centering
        \includegraphics[width=\columnwidth]{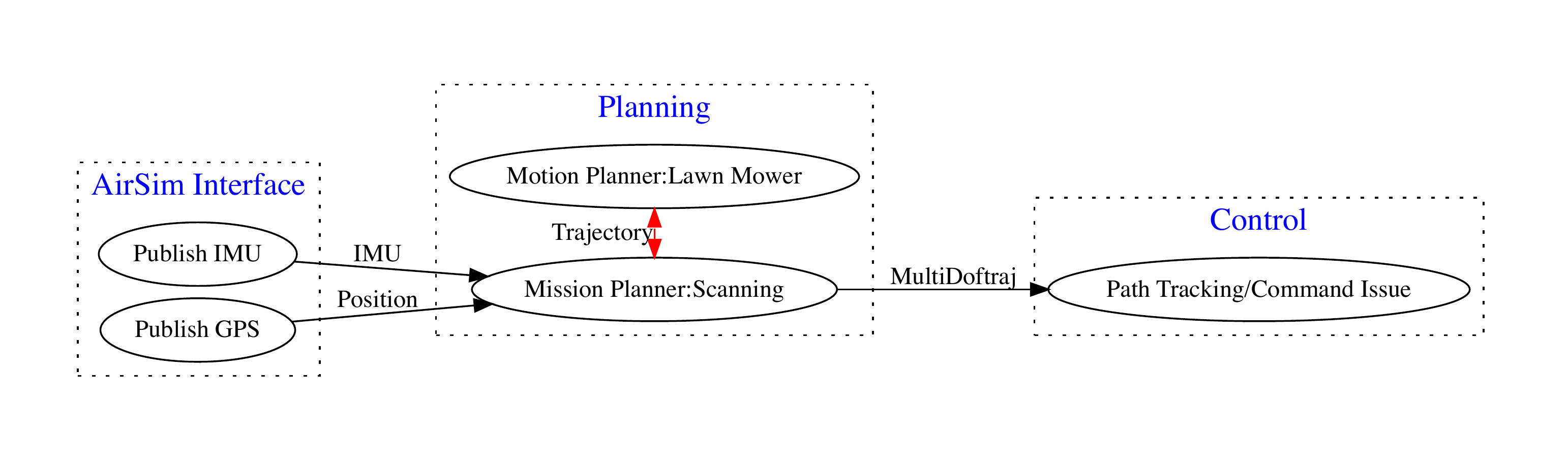}
        \vspace{-35pt}
    \caption{Scanning.}
    \label{fig:benchmarks:data-flow:scanning}
    \end{subfigure}
    \begin{subfigure}[t]{.725\columnwidth}
        \centering
        \includegraphics[width=\columnwidth]{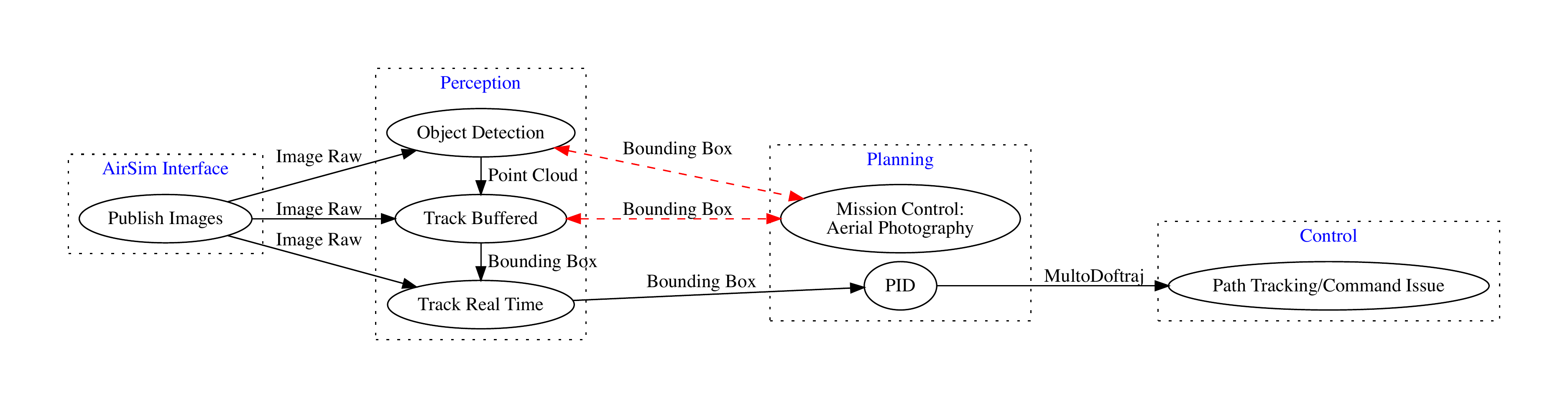}
        \vspace{-25pt}
        \caption{Aerial Photography.}\label{fig:benchmarks:data-flow:aerial_photography}
    \end{subfigure}
    \begin{subfigure}[t]{.725\columnwidth}
    \centering
        \includegraphics[width=\columnwidth] {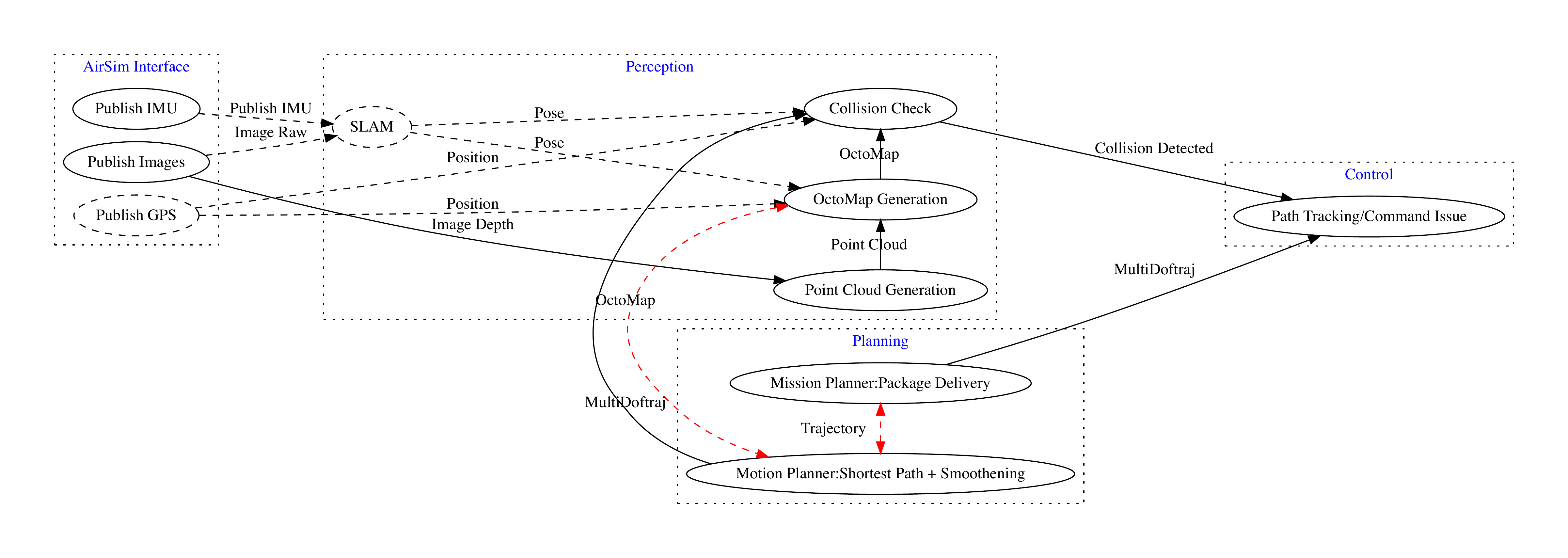}
        \vspace{-20pt}
        \caption{Package Delivery.}\label{fig:benchmarks:data-flow:package_deilvery}
    \end{subfigure}
    \begin{subfigure}[t]{.725\columnwidth}
        \centering
        \includegraphics[width=\columnwidth]{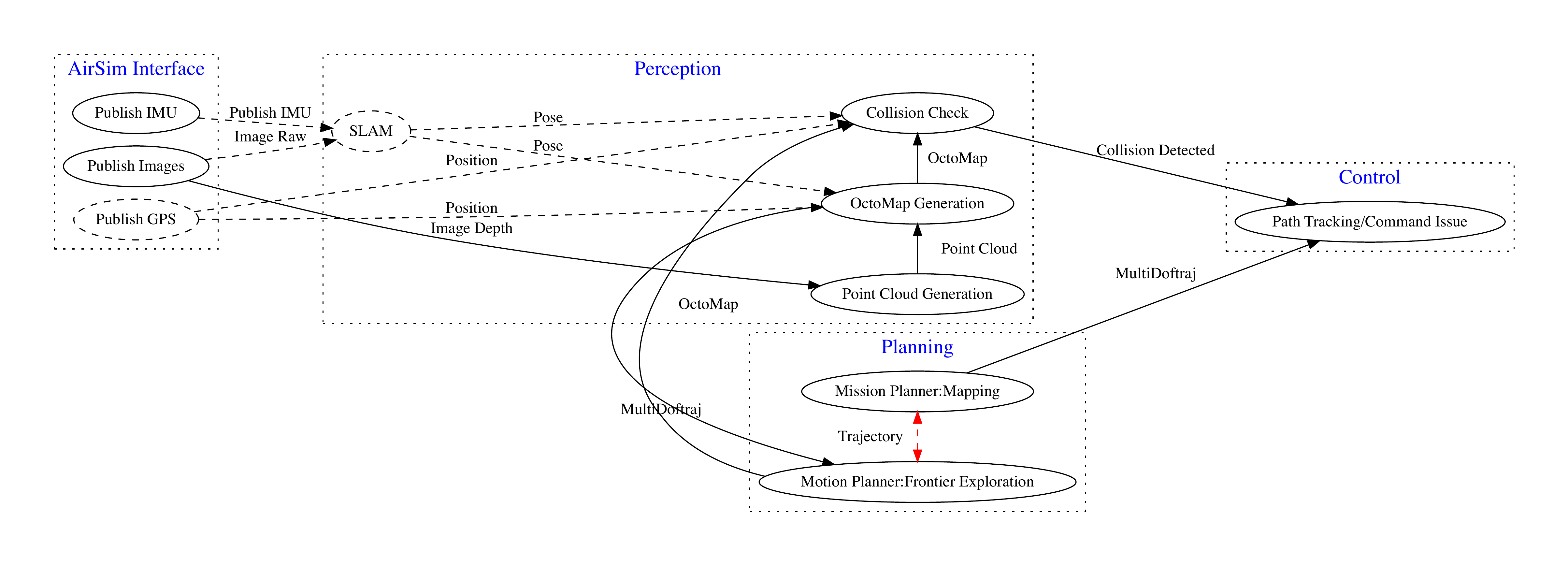}
        \vspace{-20pt}
        \caption{3D Mapping.}
        \label{fig:benchmarks:data-flow:mapping}
    \end{subfigure}
    \begin{subfigure}[t]{.725\columnwidth}
    \vspace{-5pt}
        \centering
        \includegraphics[width=\columnwidth]{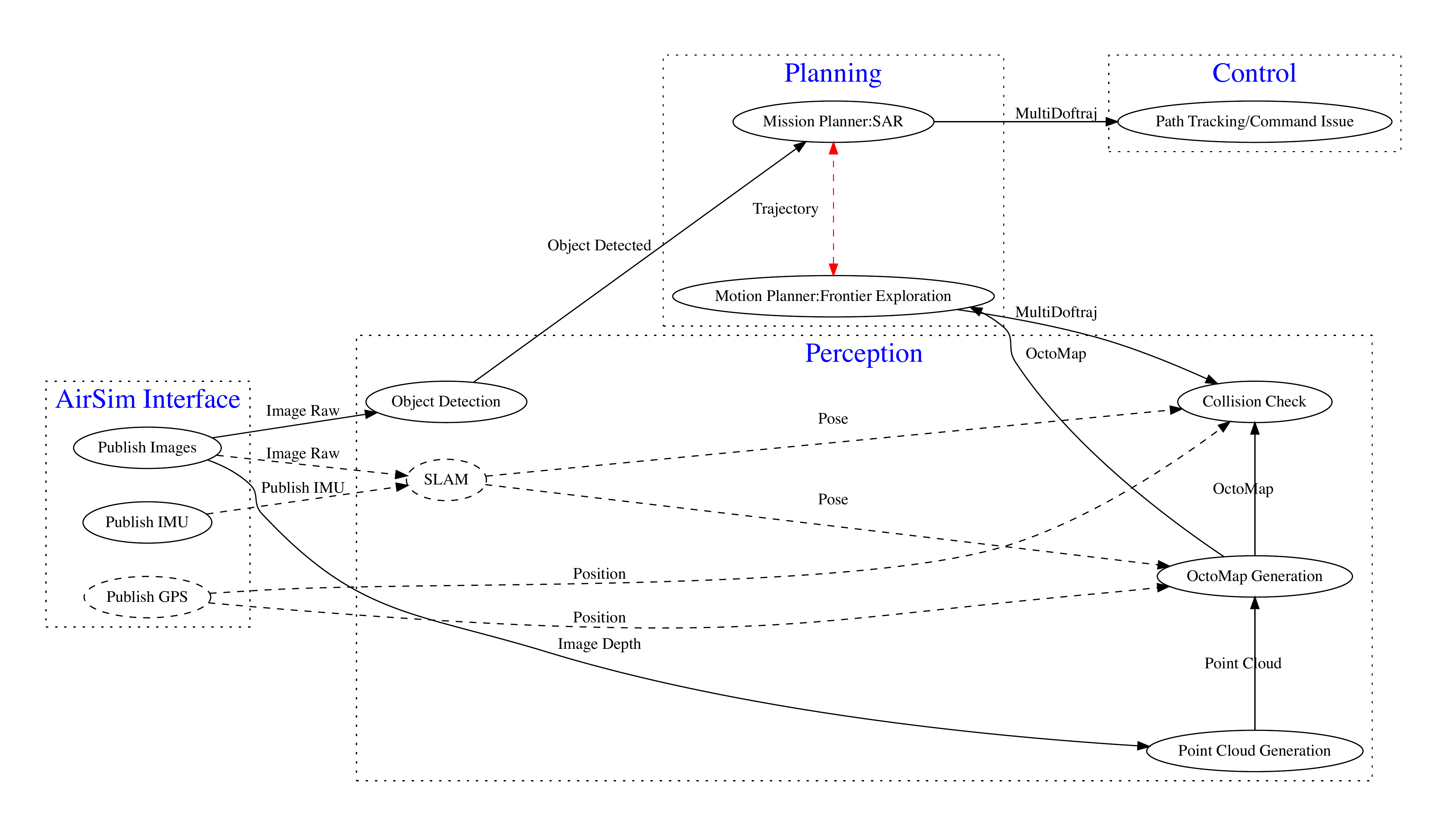}
        \vspace{-25pt}
        \caption{Search and Rescue.}\label{fig:benchmarks:data-flow:sar}
    \vspace{1pt}
    \end{subfigure}
    \caption{\footnotesize Application dataflows. Circles and arrows denote nodes and their communications respectively. Subscriber/publisher communication paradigm is denoted with filled black arrows whereas client/server with dotted red ones. Dotted black arrows denote various localization techniques.}
    \label{fig:benchmarks_data_flow}
\end{figure}

\Fig{fig:bench_screenshot} presents screenshots of our workloads. The application dataflows are shown in \Fig{fig:benchmarks_data_flow}. Note that all the workloads follow the perception, planning, and control pipeline mentioned previously. For the ease of the reader, we have also labeled the data flow with these stages accordingly.

\textbf{Scanning:} In this simple though popular use case, a MAV scans an area specified by its width and length while collecting sensory information about conditions on the ground. It is a common agricultural use case. For example, a MAV may fly above a farm to monitor the health of the crops below. To do so, the MAV first uses GPS sensors to determine its location (Perception). Then, it plans an energy efficient ``lawnmower path'' over the desired coverage area, starting from its initial position (Planning). Finally, it closely follows the planned path (Control). While in-flight, the MAV can collect data on ground conditions using onboard sensors, such as cameras or LIDAR.


\textbf{Aerial Photography:} Drone aerial photography is an increasingly popular use of MAVs for entertainment, as well as businesses. In this workload, we design the MAV to follow a moving target with the help of computer vision algorithms. The MAV uses a combination of object detection and tracking algorithms to identify its relative distance from a target (Perception). Using a PID controller, it then plans motions to keep the target near the center of the MAV's camera frame (Planning), before executing the planned motions (Control).

\textbf{Package Delivery:} In this workload, a MAV navigates through an obstacle-filled environment to reach some arbitrary destination, deliver a package and come back to its origin. Using a variety of sensors such as RGBD cameras or GPS, the MAV creates an occupancy map of its surroundings (Perception). Given this map and its desired destination coordinate, it plans an efficient collision-free path. To accommodate for the feasibility of maneuvering, the path is further smoothened to avoid high-acceleration movements (Planning), before finally being followed by the MAV (Control). While flying, the MAV continuously updates its internal map of the surroundings to check for new obstacles and re-plans its path if any such obstacles obstruct its planned trajectory.

\textbf{3D Mapping:} With use cases in mining, architecture, and other industries, this workload instructs a MAV to build a 3D map of an unknown polygonal environment specified by its boundaries. To do so, as in package delivery, the MAV builds and continuously updates an internal map of the environment with both ``known'' and ``unknown'' regions (Perception). Then, to maximize the highest area coverage in the shortest time, the map is sampled, and a heuristic is used to select an energy efficient (i.e., short) path with a high exploratory promise (i.e., with many unknown areas along the edges) (Planning). Finally, the MAV follows this path (Control), until the area has been mapped.


\textbf{Search and Rescue:} MAVs are promising vehicles for search-and-rescue scenarios where victims must be found in the aftermath of a natural disaster. For example, in a collapsed building due to an earthquake, they can accelerate the search since they are capable of navigating difficult paths by flying over and around obstacles. In this workload, a MAV is required to explore an unknown area while looking for a target such as a human.  For this workload, the \textit{3D Mapping} application is augmented with an object detection machine-learning-based algorithm in the perception stage to constantly explore and monitor its environment until a human target is detected.

\subsection{Benchmark Kernels} \label{sec:kernels}

The workloads incorporate many computational kernels that can be grouped under the three pipeline stages described earlier in Section~\ref{sec:benchmarks}. Table~\ref{kernel_makeup} shows the kernel make up of MAVBench's workloads and their corresponding time profile (measured at 2.2 GHz, 4 cores enabled mode of Jetson TX2). MAVBench is equipped with multiple implementations of each computational kernel. For example, MAVBench comes equipped with both YOLO and HOG detectors that can be used interchangeably in workloads with object detection. The user can determine which implementations to use by setting the appropriate parameters. Furthermore, our workloads are designed with a ``plug-and-play'' architecture that maximizes flexibility and modularity, so the computational kernels described below can easily be replaced with newer implementations designed by researchers in the future.

\renewcommand{\arraystretch}{1.15}
\begin{table*}[]
\centering
\caption{MAVBench applications and their kernel make up time profile in $ms$. The application suite, as a whole, exercises a variety of different computational kernels across the perception, planning and control stages, depending on their use case. Furthermore, within each of the kernel computational domain, applications have the flexibility to choose between different kernel implementations.}
\label{kernel_makeup}
\resizebox{\columnwidth}{!}{
\begin{tabular}{c|c|c|c|c|c|c|c|c|c|c|c|c|c|}
\cline{2-14}
\multirow{3}{*}{}                                                                           & \multicolumn{8}{c|}{\textbf{Perception}}                                                                                                                                                                                                                                                                                                                                                                                                                                                                      & \multicolumn{4}{c|}{\textbf{Planning}}                                                                                                                                                                                                                                                                    & \textbf{Control}                                                                                 \\ \cline{2-14} 
                                                                                            & \multirow{2}{*}{\textit{\begin{tabular}[c]{@{}c@{}}Point Cloud\\ Generation\end{tabular}}} & \multirow{2}{*}{\textit{\begin{tabular}[c]{@{}c@{}}Occupancy Map\\ Generation\end{tabular}}} & \multirow{2}{*}{\textit{\begin{tabular}[c]{@{}c@{}}Collision\\ Check\end{tabular}}} & \multirow{2}{*}{\textit{\begin{tabular}[c]{@{}c@{}}Object\\ Detection\end{tabular}}} & \multicolumn{2}{c|}{\textit{\begin{tabular}[c]{@{}c@{}}Object\\ Tracking\end{tabular}}} & \multicolumn{2}{c|}{\textit{Localization}} & \multirow{2}{*}{\textit{PID}} & \multirow{2}{*}{\textit{\begin{tabular}[c]{@{}c@{}}Smoothened\\ Shortest Path\end{tabular}}} & \multirow{2}{*}{\textit{\begin{tabular}[c]{@{}c@{}}Frontier\\ Exploration\end{tabular}}} & \multirow{2}{*}{\textit{\begin{tabular}[c]{@{}c@{}}Smoothened \\Lawn Mowing\end{tabular}}} & \multirow{2}{*}{\textit{\begin{tabular}[c]{@{}c@{}}Path Tracking/\\ Command Issue\end{tabular}}} \\ \cline{6-9}
                                                                                            &                                                                                            &                                                                                              &                                                                                     &                                                                                      & \textit{Buffered}                          & \textit{Real Time}                         & \textit{GPS}        & \textit{SLAM}        &                               &                                                                                              &                                                                                          &                                                                                 &                                                                                                  \\ \hline
\multicolumn{1}{|c|}{\textbf{Scanning}}                                                     &                                                                                            &                                                                                              &                                                                                     &                                                                                      &                                            &                                            &                     &                      &                               &                                                                                              &                                                                                          & 89                                                                              & 1                                                                                                \\ \hline
\multicolumn{1}{|c|}{\textbf{\begin{tabular}[c]{@{}c@{}}Aerial\\ Photography\end{tabular}}} &                                                                                            &                                                                                              &                                                                                     & 307                                                                                  & 80                                         & 18                                         & 0                   &                      & 0                             &                                                                                              &                                                                                          &                                                                                 & 1                                                                                                \\ \hline
\multicolumn{1}{|c|}{\textbf{\begin{tabular}[c]{@{}c@{}}Package\\ Delivery\end{tabular}}}   & 2                                                                                          & 630                                                                                          & 1                                                                                   &                                                                                      &                                            &                                            & 0                   & 55                   &                               & 182                                                                                          &                                                                                          &                                                                                 & 1                                                                                                \\ \hline
\multicolumn{1}{|c|}{\textbf{\begin{tabular}[c]{@{}c@{}}3D\\ Mapping\end{tabular}}}         & 2                                                                                          & 482                                                                                          & 1                                                                                   &                                                                                      &                                            &                                            &              0       & 46                   &                               &                                                                                              & 2647                                                                                     &                                                                                 & 1                                                                                                \\ \hline
\multicolumn{1}{|c|}{\textbf{\begin{tabular}[c]{@{}c@{}}Search and\\ Rescue\end{tabular}}}  & 2                                                                                          & 427                                                                                          & 1                                                                                   & 271                                                                                  &                                            &                                            &               0      & 45                   &                               &                                                                                              & 2693                                                                                     &                                                                                 & 1                                                                                                \\ \hline
\end{tabular}
}
\end{table*}
\renewcommand{\arraystretch}{1}

\paragraph{Perception Kernels:} These are the computational kernels that allow a MAV application to interpret its surroundings.

\textit{Object Detection:} Detecting objects is an important kernel in numerous intelligent robotics applications. So, it is part of two MAVBench workloads: \textit{Aerial Photography} and \textit{Search and Rescue}. MAVBench comes pre-packaged with the YOLO~\cite{yolo16} object detector, and the standard OpenCV implementations of the HOG~\cite{hog} and Haar people detectors.

\textit{Tracking:}
It attempts to follow an instance of an object as it moves across a scene. 
This kernel is used in the \textit{Aerial Photography} workload. MAVBench comes pre-packaged with a C++ implementation~\cite{kcf-c++} of a KCF~\cite{kcf} tracker.

\textit{Localization:} MAVs must determine their position. There are many ways that have been devised to enable localization, using a variety of different sensors, hardware, and algorithmic techniques. MAVBench comes pre-packaged with multiple localization solutions that can be used interchangeably for benchmark applications. Examples include a simulated GPS, visual odometry algorithms such as ORB-SLAM2~\cite{orbslam2}, and VINS-Mono~\cite{vins-mono} and these are accompanied with ground-truth data that can be used when a MAVBench user wants to test an application with perfect localization data.


\textit{Occupancy Map Generation:} Several MAVBench workloads, like many other robotics applications, model their environments using internal 3D occupancy maps that divide a drone's surroundings into occupied and unoccupied space. Noisy sensors are accounted for by assigning probabilistic values to each unit of space.
In MAVBench we use OctoMap~\cite{octomap} as our occupancy map generator since it provides updatable, flexible and compact 3D maps.

\paragraph{Planning Kernels:} Our workloads comprise several motion-planning techniques, from simple ``lawnmower" path planning to more sophisticated sampling-based path-planners, such as RRT~\cite{rrt} or PRM~\cite{prm} paired with the A*~\cite{astar} algorithm. We divide MAVBench's path-planning kernels into three categories: \textit{shortest-path planners}, \textit{frontier-exploration planners}, and \textit{lawnmower path planners}. The planned paths are further smoothened using the \textit{path smoothening} kernel. 


\textit{Shortest Path:} Shortest-path planners find collision-free flight trajectories that minimize the MAV's traveling distance. MAVBench comes pre-packaged with OMPL~\cite{ompl}, the Open Motion Planning Library, consisting of many state-of-the-art sampling-based motion planning algorithms. These algorithms provide collision-free paths from an arbitrary start location to an arbitrary destination. 

\textit{Frontier Exploration:} 
Some applications incorporate collision-free motion-planners that aim to efficiently ``explore'' all accessible regions in an environment, rather than simply moving from a single start location to a single destination as quickly as possible.
For these applications, MAVBench comes equipped with the official implementation of the exploration-based ``next best view planner''~\cite{nbvplanner}.

\textit{Lawnmower:} Some applications do not require complex, collision-checking path planners, e.g., agricultural MAVs fly over farms in a simple, lawnmower pattern, where the high-altitude of the MAV means that obstacles can be assumed to be nonexistent. For such applications, MAVBench comes with a simple path-planner that computes a regular pattern for covering rectangular areas.

\textit{Path Smoothening:} The motion planners discussed earlier return piecewise trajectories that are composed of straight lines with sharp turns. However, sharp turns require high accelerations from a MAV, consuming high amounts of energy (i.e., battery capacity). Thus, we use this kernel to convert these piecewise paths to smooth, polynomial trajectories that are more efficient for a MAV to follow.

\paragraph{Control Kernels:} The control stage of the pipeline enables the MAV to closely follow its planned motion trajectories in an energy-efficient, stable manner. 

\textit{Path Tracking:} MAVBench applications produce trajectories that have specific positions, velocities, and accelerations for the MAV to occupy at any particular point in time. However, due to mechanical constraints, the MAV may drift from its location as it follows a trajectory, due to small but accumulated errors. So, MAVBench includes a computational kernel that guides MAVs to follow trajectories while repeatedly checking and correcting the error in the MAV's position.

\subsection{Quality-of-Flight (QoF) Metrics}
\label{sec:QoF}

Metrics are key for quantitive evaluation and comparison of different systems. In traditional computing systems, we use Quality-of-Service (QoS), Quality-of-Experience (QoE) etc. to evaluate computer system performance for servers and mobile systems, respectively. Similarly, various figures of merits can be used to measure a drone's mission quality. These metrics otherwise called as mission metrics measure mission success and also throughout this paper are used to gauge and quantify compute impact on the drone's behavior. While some of these metrics are universally applicable across applications, others are specific to the application under inquiry. On the one hand, for example, a mission's overall time and energy consumption are almost universally of concern. On the other hand, the discrepancy between a collected and ground truth map or the distance between the target's image and the frame center are specialized metrics for 3D mapping and aerial photography respectively. MAVBench platform collects statistics of both sorts; however, this paper mainly focuses on time and energy due to their universality and applicability to our goal of cyber-physical co-design. 

\section{Evaluation Setup}
\label{sec:evaluation}

\begin{figure}[t!]
\vspace{-12pt}
\begin{subfigure}{\linewidth}
\centering
\includegraphics[trim=0 0 0 0, clip, width=.46\columnwidth]
{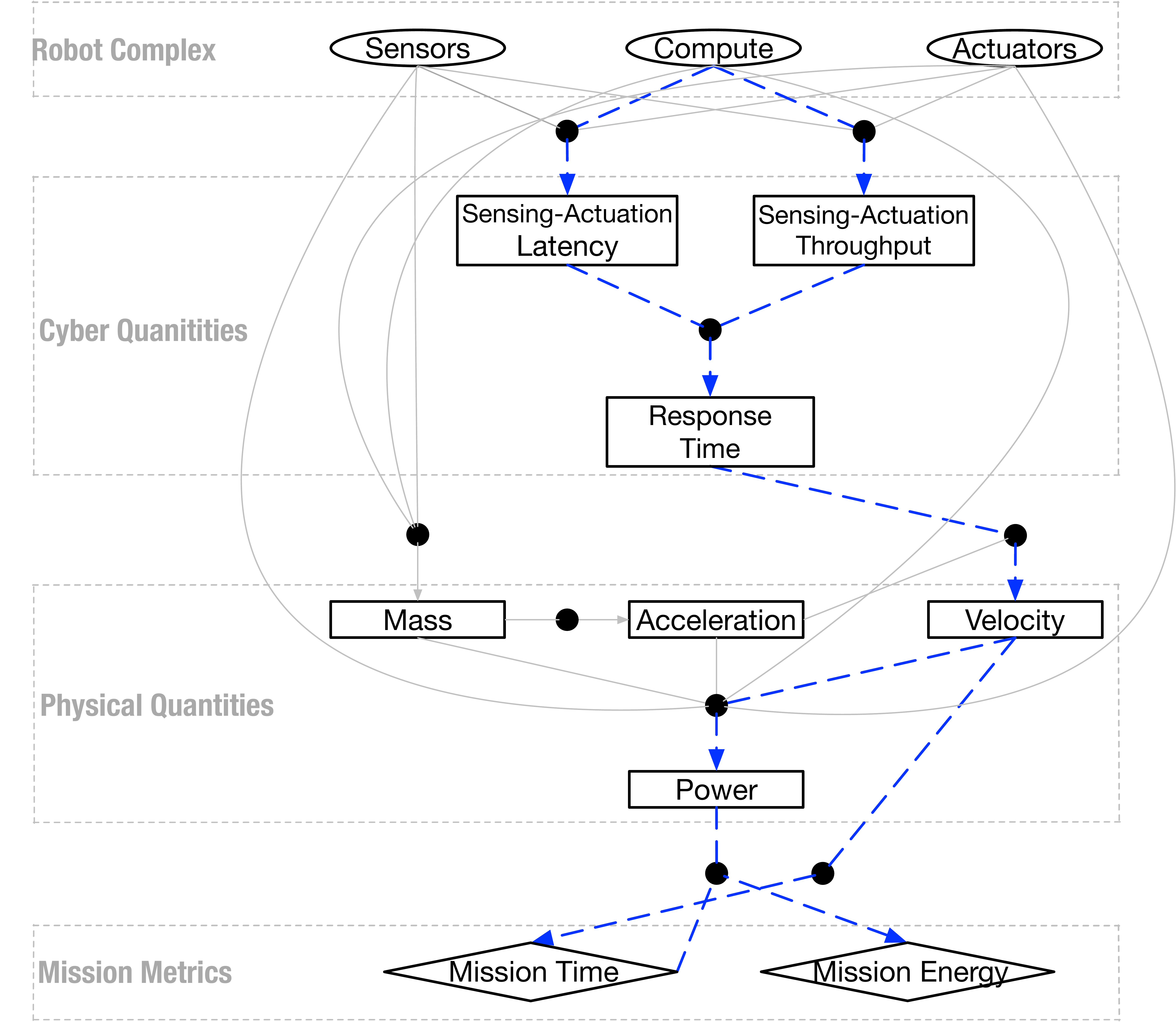}
\caption{Performance cluster. Impact paths influencing mission-time/energy through latency/throughput.}
\label{fig:CIG_performance_paths}
\end{subfigure}
\begin{subfigure}{\linewidth}
\centering
\includegraphics[trim=0 0 0 0, clip, width=.46\columnwidth]
{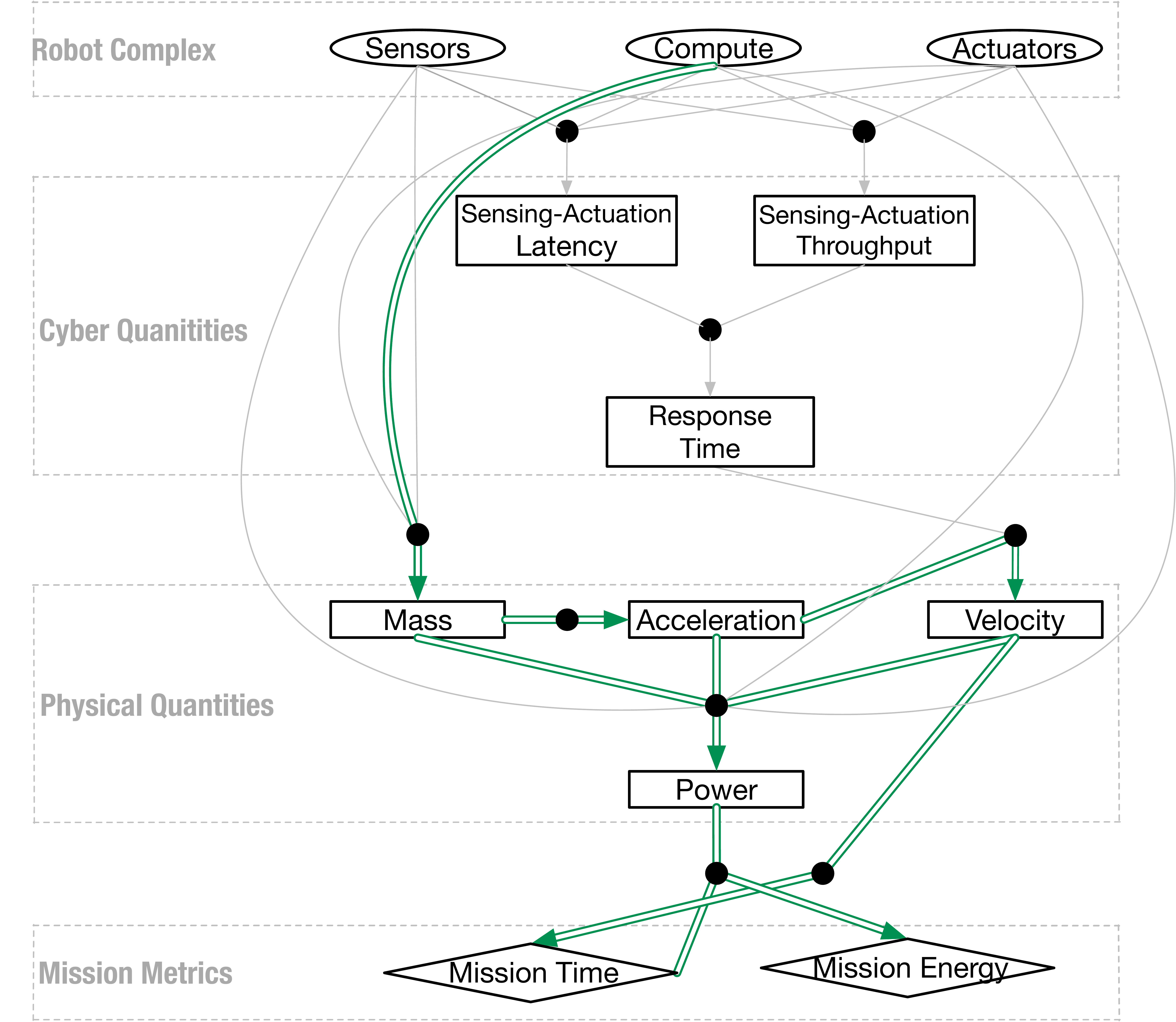}
\caption{Mass cluster. Impact paths influencing mission time and energy through compute mass.}
\label{fig:CIG_mass_paths}
\end{subfigure}
\begin{subfigure}{\linewidth}
\centering
\includegraphics[trim=0 0 0 0, clip, width=.46\columnwidth]
{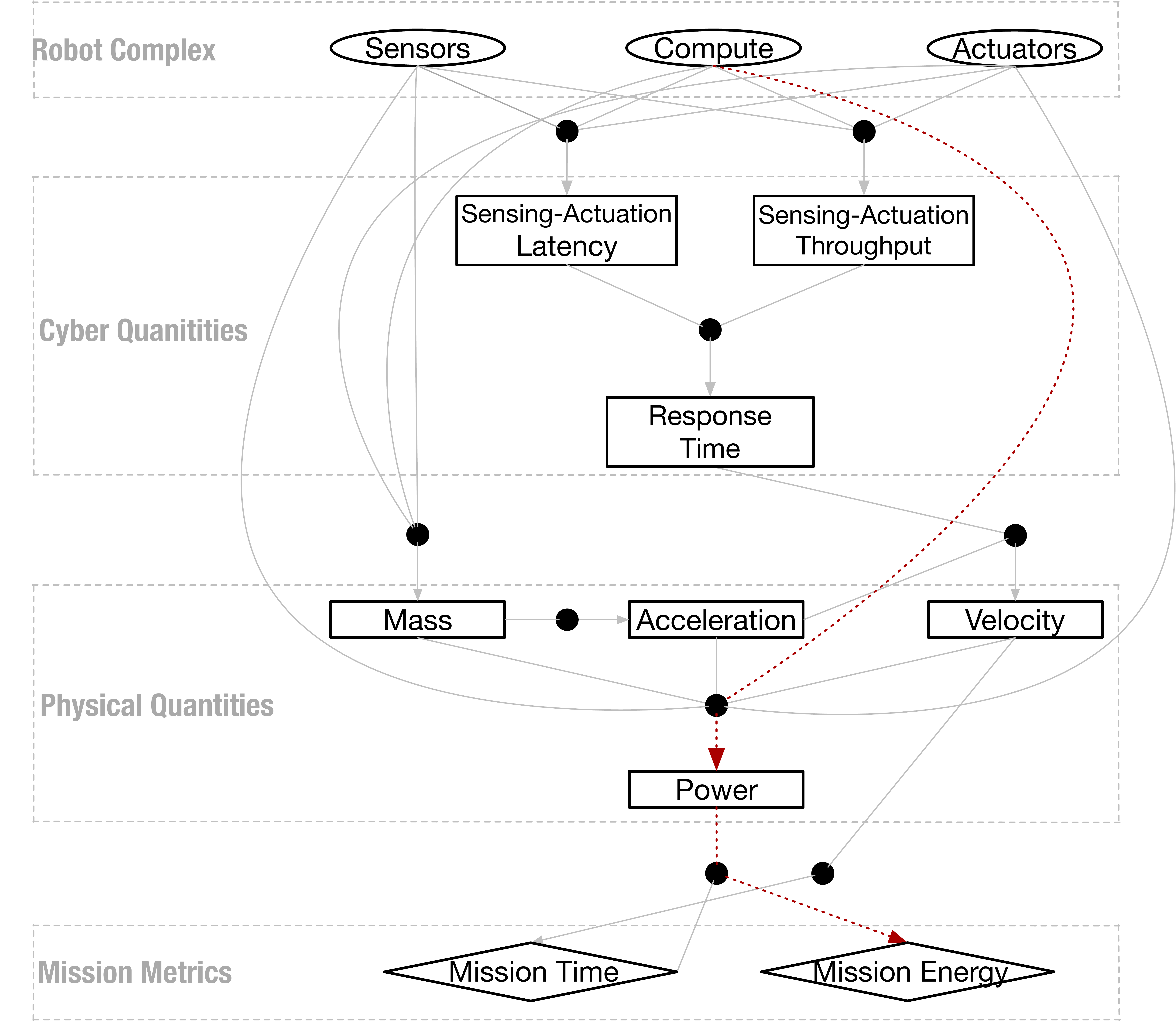}
\caption{Power cluster. Impact paths influencing mission time and energy through compute power.}
\label{fig:CIG_power_paths}
\end{subfigure}
\caption{Three impact clusters, performance, mass, and power,  impacting mission time and energy. Each cluster with all the paths contained in it are shown with a different color. Having the
cyber-physical interaction graph with different clusters enables the cyber-physical co-design advocated in this paper.}
\label{fig:CIG_all}
\end{figure}

We want to study how for a cyber-physical mobile machine such as a MAV, \textit{the fundamentals of compute relate to the fundamentals of motion}. To this end, we combine theory, system modeling, and micro and end-to-end benchmarking using MAVBench. The next three sections detail our experimental evaluation and in-depth studies. We deploy our \csig to investigate paths that start from compute and end with mission time or energy. To assist the reader in the semantic understanding of the various impacts, we bin the impact paths into three clusters:
\begin{enumerate}
\item Performance impact cluster: Impact paths that originate from compute performance (i.e., sensing-to-actuation latency and throughput) which are shown in blue-color/coarse-grained-dashed lines in Figure~\ref{fig:CIG_performance_paths}.

\item Mass impact cluster: Impact paths that originate from compute mass which are shown in green-color/double-sided lines in Figure~\ref{fig:CIG_mass_paths}.

\item Power impact cluster: Impact paths that originate from compute power which are shown in red-color/fine-grained-dashed lines in Figure~\ref{fig:CIG_power_paths}.
\end{enumerate}

At first, we study the impact of each cluster on mission time (Section ~\ref{sec:comp_mission_time}) and energy (Section ~\ref{sec:comp_mission_energy}) separately. This allows us to isolate their effect in order to gain better insights into their inner workings. Then we combine all clusters together and study them holistically in order to understand their aggregate impact (Section~\ref{sec:impact_holistic}). 

In the \textbf{compute performance and power studies}, we conduct a series of sensitivity analysis using core and frequency scaling on an NVIDIA TX2. The TX2 has two sets of cores, a \textit{Dual-Core NVIDIA Denver 2}
and a \textit{Quad-Core ARM Cortex-A57}. We turned off the Denver cores during our experiments to ensure that the indeterminism caused by process to core mapping variations across runs would not affect our results. We profile and present the average velocity, mission, and energy values of various operating points for our end-to-end applications. 

In the \textbf{compute mass and holistic studies}, we use four different compute platforms with different compute capabilities and mass ranging from a lower-power TX2 to high-performance, power-hungry
Intel Core-i9. These studies model a mission where the drone is required to traverse a \SI{1}{\km} path to deliver a package. We collect sensing-to-actuation latency and throughput values by running a package delivery application as a micro benchmark for 30 times on each platform. Mission time is calculated using the velocity and the path length while the power and energy are calculated using our experimentally verified models provided in Section~\ref{sec:energy}. 
\section{compute impact on mission time}
\label{sec:comp_mission_time}
In this section, we take a deep dive exploring how compute impacts mission time through a combination of analytical models, simulation, and micro and end-to-end benchmarking.
Briefly, compute impacts mission time through both cyber and physical quantities. It impacts cyber quantities such as sensing-to-actuation latency, throughput and ultimately response time, and also impacts
physical quantities such as drone's mass, velocity, and acceleration. Such impacts percolate down to the bottom of the \csig influencing mission metrics such as mission time. 
This section studies each impact cluster separately to isolate their effect so that we gain better insights into their inner working. First, we explain the impact paths in the performance cluster (Figure~\ref{fig:CIG_performance_paths}, blue-color/coarse-grained-dashed paths), and
then, we explain the paths in the mass cluster (Figure~\ref{fig:CIG_mass_paths}, green-color/double-sided paths).

\subsection{Compute Performance Impact on Mission Time}
\label{sec:comp_perf_impact_on_mission_time}
Compute reduces mission time through performance cluster by impacting physical quantities, such as the drone's average velocity (performance cluster shown in Figure~\ref{fig:CIG_performance_paths} with the  blue-color/coarse-grained-dashed paths). A MAV's average velocity is a function of its response time, i.e., how quickly it can respond to a new event, such as the emergence of an obstacle in its environment. By improving response time, compute allows the drone to fly faster while being safe (i.e., with no collisions), and flying faster in return reduces the mission time. To achieve a high average velocity throughout the mission, the drone needs to be capable of reaching a high velocity (maximum velocity) and also quickly arrive at it (high acceleration). We discuss compute-maximum velocity relationship and leave the compute-acceleration discussion to the next section.   

 \begin{figure}[t!]
    \centering
    \includegraphics[trim=0 0 0 0, clip, width=.4\columnwidth]{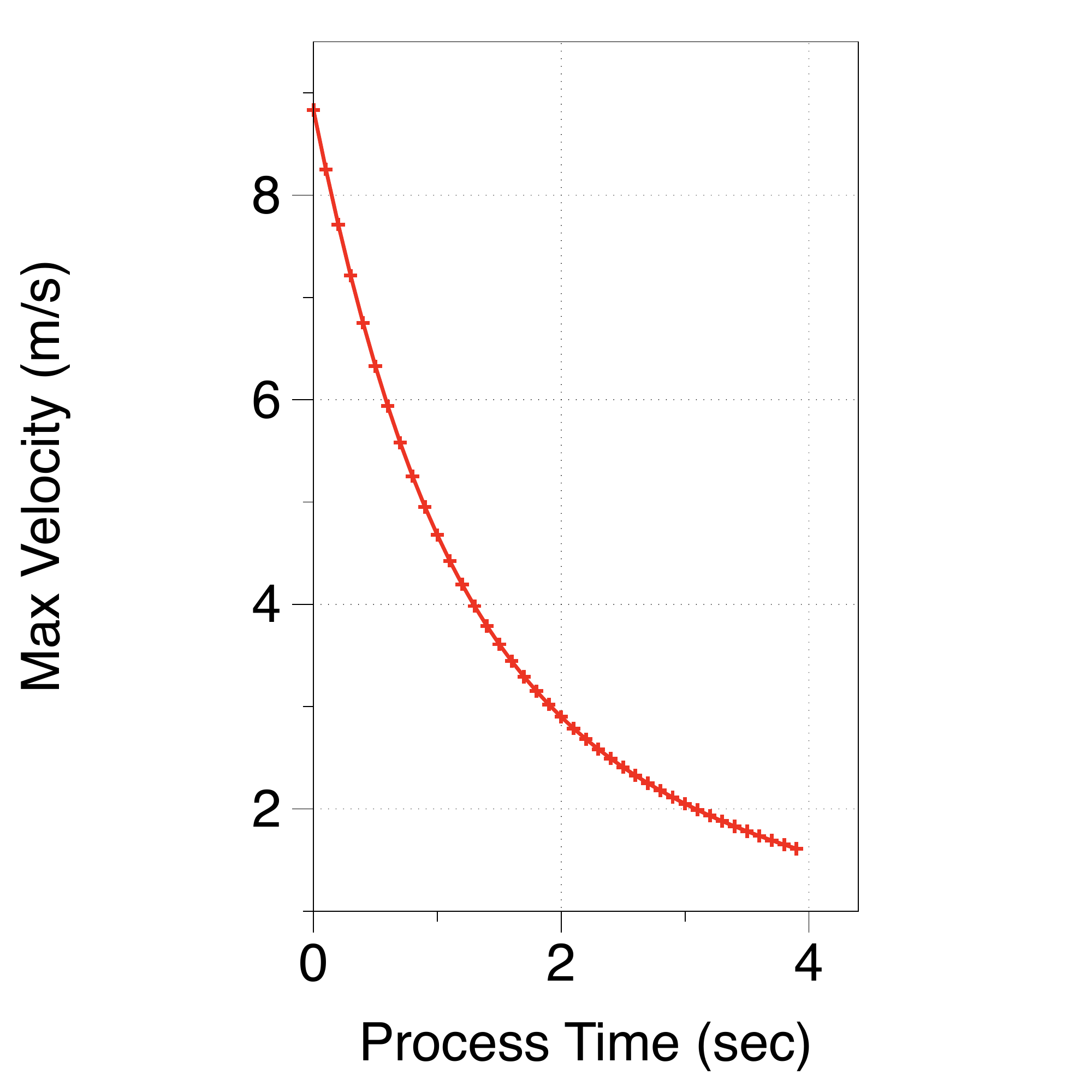}
    \caption{Theoretical max velocity and response time relationship.}
    \label{fig:process-time-velocity}
    \end{figure}



\paragraph{Improving Maximum Velocity By \textit{Reducing Response Time:}}
Drone's maximum velocity is not only mechanically bounded but also computationally bounded. Equation~\ref{eq:runtime-compute-bound} shows this where response time, a cyber quantity determined by compute, impacts velocity, a physical quantity. The variables $\delta{t}_{response}$, $d$, $a_{max}$ and $v_{max}$ denote response time, distance from obstacle,
maximum acceleration limit of the drone and maximum allowed velocity, respectively. Applying Equation~\ref{eq:runtime-compute-bound} for out simulated DJI Matric 100 drone,  Figure~\ref{fig:process-time-velocity} shows that, in theory, the drone's maximum velocity takes a value between 1.57 m/s to 8.83 m/s given a response time ranging from 0 to 4~seconds.

\begin{equation}
\label{eq:runtime-compute-bound}
\boxed{
v_{max} = a_{max}(~\sqrt[]{\delta{t_{response}}^2 + 2\frac{d}{a_{max}}} - \delta{t_{response}})}
\end{equation}


To help explain the relationship between compute and velocity, we step through a typical obstacle avoidance task whose maximum velocity obeys this equation. At a high level, a MAV periodically takes snapshots of its environment and then spends some processing time responding to the emerging obstacles in its path ($\delta{t}_{response}$). However, if the motion planner fails to find a trajectory that circumvents the obstacle, the drone needs to decelerate immediately ($a_{max}$) to avoid running into the obstacle. In the worst case, the drone needs to be able to decelerate from its maximum speed ($v_{max}$).

\begin{figure}[t!]
\centering
    \begin{subfigure}{\columnwidth}
    \centering
   \includegraphics[trim=0 0 0 0, clip, width=1.0\columnwidth]{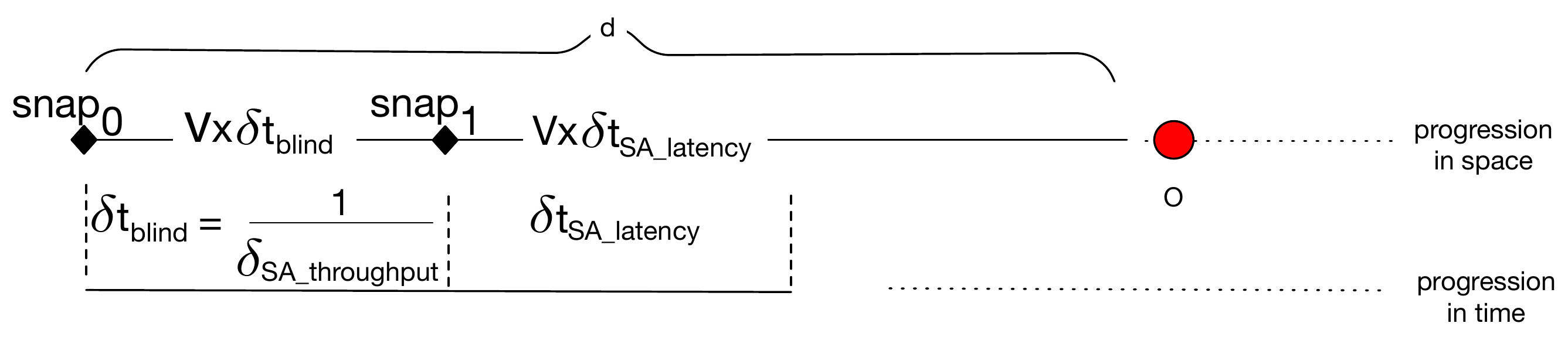}
    
    \end{subfigure}
\caption{Obstacle avoidance in action, a bird's-eye view. Note the progression in time as a result of cyber quantities such as sensing-to-actuation latency, and the progression in space as the result of physical quantities such as $v$.}
\label{fig:vmax_gen}
\end{figure}

Figure~\ref{fig:vmax_gen} shows the progression of this task for two snapshots, $snap_{0}$ and $snap_{1}$. We call the rate with which these snapshots occur sensing-to-actuation throughput (denoted by $\delta_{SA\_throughput}$). Between the two snapshots, i.e, inverse of the throughput, the drone is blind (Equation~\ref{eq:blind}). This is because \textbf{no} new snapshot are taken, and hence the drone is unaware of any changes in the environment during this period. In the
worst-case scenario, an obstacle (O) can be hiding within the blind space caused by $\delta{t}_{blind}$. This reduces the 
distance between the drone and the obstacle by 
$v_{max}$*$\delta{t}_{blind}$ (Equation~\ref{eq:distance_to_obs_after_blind}). After this blind period, at point $snap_{1}$, the second snapshot is taken and the drone spends sensing-to-actuation latency ($\delta{t}_{SA\_latency}$), 
to perceive ($\delta{t}_{pr}$), plan ($\delta{t}_{pl}$) and control ($\delta{t}_{c}$), traversing the PPC pipeline, to formulate and follow a trajectory to circumvent the obstacle (Equation \ref{eq:reaction_latency}).
Equation~\ref{eq:distance_to_obs_after_reaction_formulation} shows the distance between the drone and the obstacle
after this traversal.

\begin{equation}
\label{eq:blind}
\delta{t}_{blind} = \frac{1}{\delta_{SA\_throughput}}
\end{equation}

\begin{equation}
\begin{aligned}
\label{eq:distance_to_obs_after_blind}
Distance\;to\;Obstacle\;After\;Blind\;Time &=  d - v_{max}*\delta{t}_{blind} \\ &= d - v_{max}*\frac{1}{\delta_{SA\_throughput}} 
\end{aligned}
\end{equation}

\begin{equation}
\label{eq:reaction_latency}
\delta{t}_{SA\_latency} = \delta{t}_{pr} + \delta{t}_{pl} + \delta{t}_{c}
\end{equation}

\begin{equation}
    \label{eq:distance_to_obs_after_reaction_formulation}
\begin{aligned}
Distance\;to\;Obstacle\;After\;PPC\;Traversal  =  d - v_{max}*\frac{1}{\delta_{SA\_throughput}} - v_{max}*\delta{t}_{SA\_latency}
\end{aligned}
\end{equation}
 
At this point, if the drone fails to generate a plan, it must decelerate and ideally come to a halt before running into the obstacle in its current path. 
Equation \ref{eq:stop-distance} shows the distance that it takes for a moving body to come to a complete stop. Setting \ref{eq:distance_to_obs_after_reaction_formulation} and \ref{eq:stop-distance}
equal to one another and solving for $v$ results in Equation~\ref{eq:runtime-general}, the absolute maximum velocity with which the drone is allowed to fly and still be able to guarantee a collision-free mission. This equation shows the relationship between two cyber quantities, i.e.,
$\delta{t}_{SA\_latency}$ and
$\delta_{SA\_throughput}$, and a physical quantity, i.e., $v$.\footnote{If we pair this equation with Equation~\ref{eq:runtime-compute-bound}, we see that for a drone to be able to respond to an obstacle in the worst case scenario, it needs to spend a total of $\delta{t}_{SA\_latency}$ plus inverse of
$\delta_{SA\_throughput}$ which indeed is the response time (Equation~\ref{eq:response_time}) of the MAV to an emerging event (obstacle).}

\begin{equation}
\label{eq:stop-distance}
Stopping\;Distance= \frac{v^2}{2*a_{max}} 
\end{equation}

\begin{equation}
\label{eq:runtime-general}
\boxed{
v_{max} = a_{max}\left(~\sqrt[]{\left(\delta{t}_{SA\_latency} + \frac{1}{\delta_{SA\_throughput}}\right)^2 + 2\frac{d}{a_{max}}}
- \left(\delta{t}_{SA\_latency} + \frac{1}{\delta_{SA\_throughput}}\right)\right)}
\end{equation}

\begin{equation}
\label{eq:response_time}
\delta{t}_{response} = \delta{t}_{SA\_latency} + \delta_{SA\_throughput}
\end{equation}

\begin{figure}[t!]
\centering
    \begin{subfigure}{.49\columnwidth}
    \includegraphics[trim=0 0 0 0, clip, width=1.0\columnwidth]{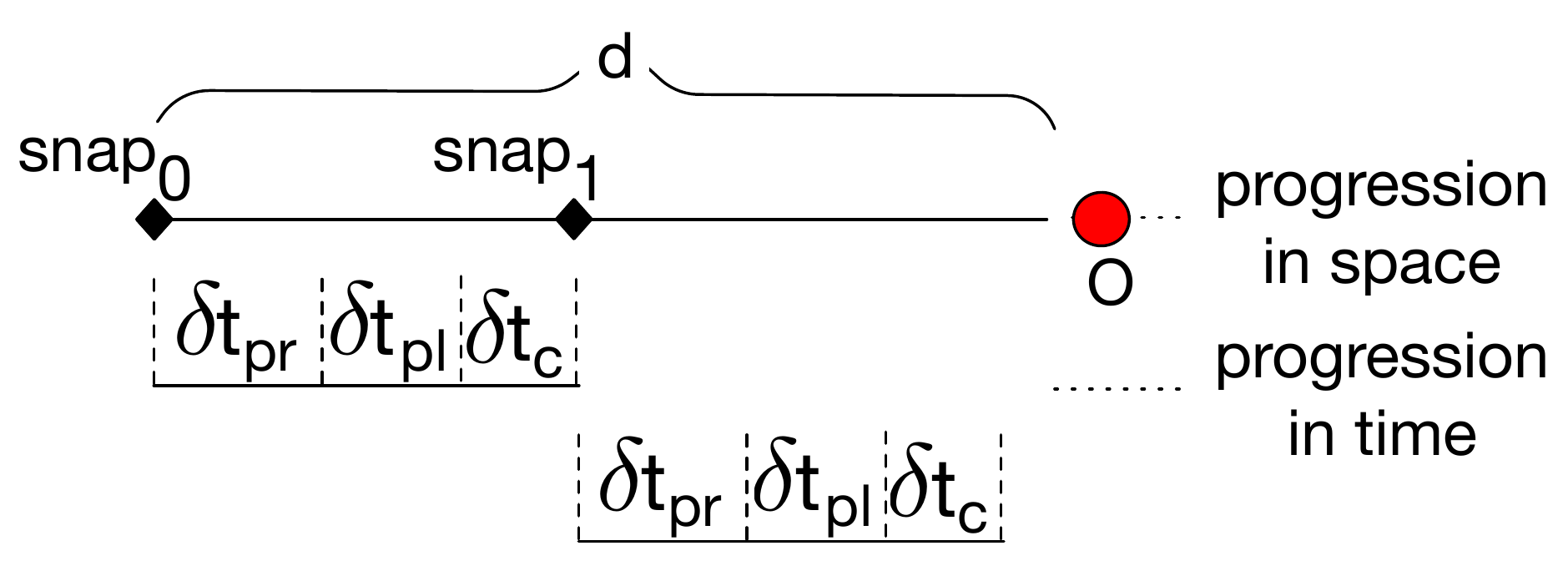}
    \caption{Sequential paradigm.}
    \label{fig:vmax_seq}
    \end{subfigure}
    \begin{subfigure}{.49\columnwidth}
   \includegraphics[trim=0 0 0 0, clip, width=1.0\columnwidth]{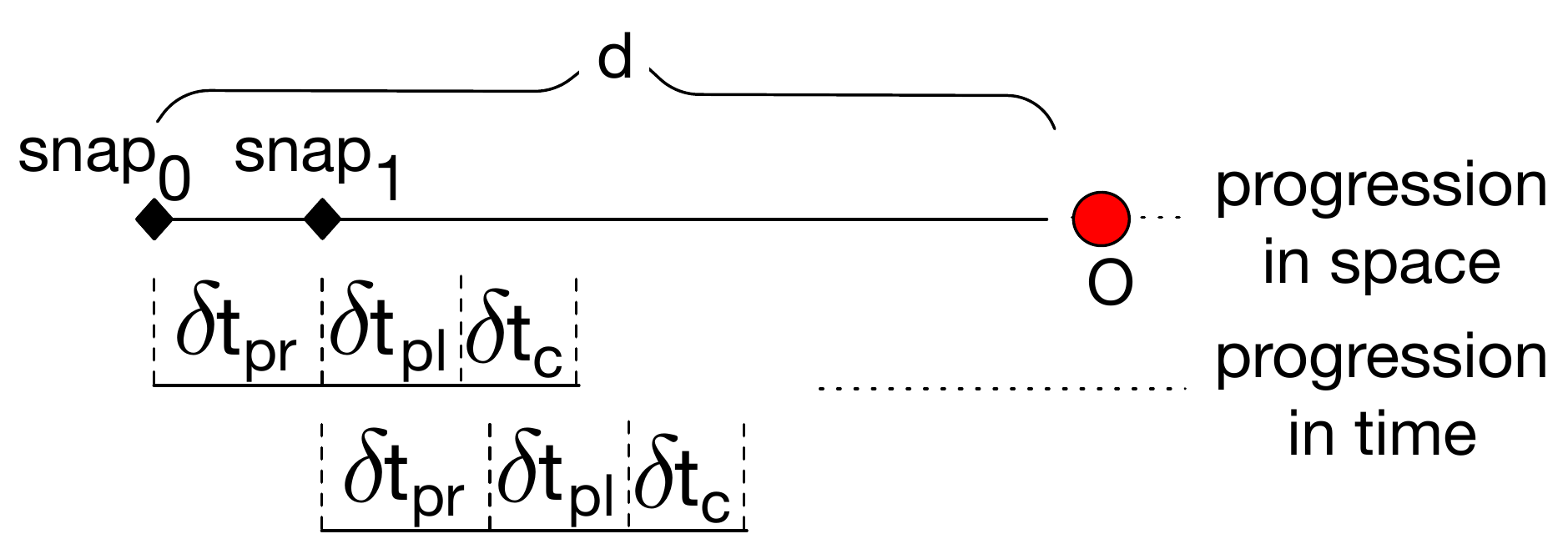}
    \caption{Pipelined paradigm.}
    \label{fig:vmax_pipe}
    \end{subfigure}
       \caption{Obstacle avoidance with the PPC pipeline. Latency associated with each stage is denoted with $\delta$. Two different design paradigms are presented.}
      \label{fig:comp_stop_distance}
\end{figure}

Investigating how system design choices impact $\delta_{SA\_throughput}$ and  $\delta{t}_{SA\_latency}$ (and hence response time and velocity)
demands computer and system architects' attention. For example, consider the sequential versus a pipelined design paradigm.
In the sequential processing paradigm, while the drone is going through one iteration of the PPC pipeline, no new snapshots are taken (Figure~\ref{fig:vmax_seq}). This means that the
sensing-to-actuation throughput is the inverse of sensing-to-actuation latency (Equation~\ref{eq:reaction_throughput}).
\begin{equation}
\label{eq:reaction_throughput}
\delta_{SA\_throughput} = \frac{1}{\delta{t}_{SA\_latency}}
\end{equation} 
 
This implies that we can rewrite Equations
~\ref{eq:distance_to_obs_after_blind},~\ref{eq:distance_to_obs_after_reaction_formulation}, 
~\ref{eq:runtime-general} and
~\ref{eq:response_time} as such:

\begin{equation}
\label{eq:distance_to_obs_after_blind_seq}
Distance\;to\;Obstacle\;After\;Blind\;Time  = d - v_{max}*\delta{t}_{SA\_latency}
\end{equation}

\begin{equation}
    \label{eq:distance_to_obs_after_reaction_formulation_seq}
\begin{aligned}
Distance\;to\;Obstacle\;After\;SA\;Traversal  =  
d -  v_{max}*2*\delta{t}_{SA\_latency}
\end{aligned}
\end{equation}

\begin{equation}
\label{eq:response_time_seq}
\delta{t}_{response} = 2*\delta{t}_{SA\_latency}
\end{equation}

resulting in a $v_{max}$ of:
\begin{equation}
\label{eq:runtime-compute_seq}
\boxed{
v_{max} = a_{max}\left(~\sqrt[]{4*\delta{t}_{SA\_latency}^2 + 2\frac{d}{a_{max}}} - 2*\delta{t}_{SA\_latency}\right)}
\end{equation}
\setlength{\belowcaptionskip}{-1ex}

However, in a pipelined processing paradigm (Figure~\ref{fig:vmax_pipe}), perception, planning and control stages overlap with one another. Hence, it is possible for us to reduce the $\delta_{SA\_throughput}$ and thereby cut down $\delta{t}_{blind}$ to the minimum of latency of each stage (Equation~\ref{eq:throughput}). Note that in this design $\delta{t}_{SA\_latency}$ stays intact.
Using the pipeline approach, the velocity is calculated using Equation~\ref{eq:runtime-pipelined}. 

\begin{equation}
\label{eq:throughput}
\begin{aligned}
 \delta{t}_{blind} = \frac{1}{\delta_{SA\_throughput}} = \frac{1}{Min(\frac{1}{\delta{t}_{pr}},\frac{1}{\delta{t}_{pl}}, \frac{1}{\delta{t}_{c}})} = 
 Max(\delta{t}_{pr},\delta{t}_{pl}, \delta{t}_{c}) 
 \end{aligned}
\end{equation}

\setlength{\belowcaptionskip}{-1ex}

\begin{equation}
\label{eq:runtime-pipelined}
\boxed{
\begin{aligned}
v_{max} = a_{max}(& ~\sqrt[]{(\delta{t}_{SA\_latency} +  Max(\delta{t}_{pr}, \delta{t}_{pl}, \delta{t}_{c})^2 + 2\frac{d}{a_{max}}}  - \\ 
&(\delta{t}_{SA\_latency} +  Max(\delta{t}_{pr}, \delta{t}_{pl}, \delta{t}_{c})))
\end{aligned}
}
\end{equation}

There is a tradeoff in opting between the sequential versus pipeline paradigms. However, the choice is not straightforward. Opting for one or the other requires a rigorous and thorough investigation by system designers. For example, simply pipelining the design does not necessarily improve the velocity. This is because the response time is equal to the addition of $\delta{t}_{SA\_latency}$ (see above) and inverse of $\delta_{SA\_throughput}$. Therefore, if the pipelined design increases $\delta{t}_{SA\_latency}$ (e.g., due to the communication overhead between parallel processes), the overall response time might increase. 

\begin{figure}[t!]
\centering
    \begin{subfigure}{.4\columnwidth}
    \centering
   \includegraphics[trim=0 0 0 0, clip, width=1.0\columnwidth]{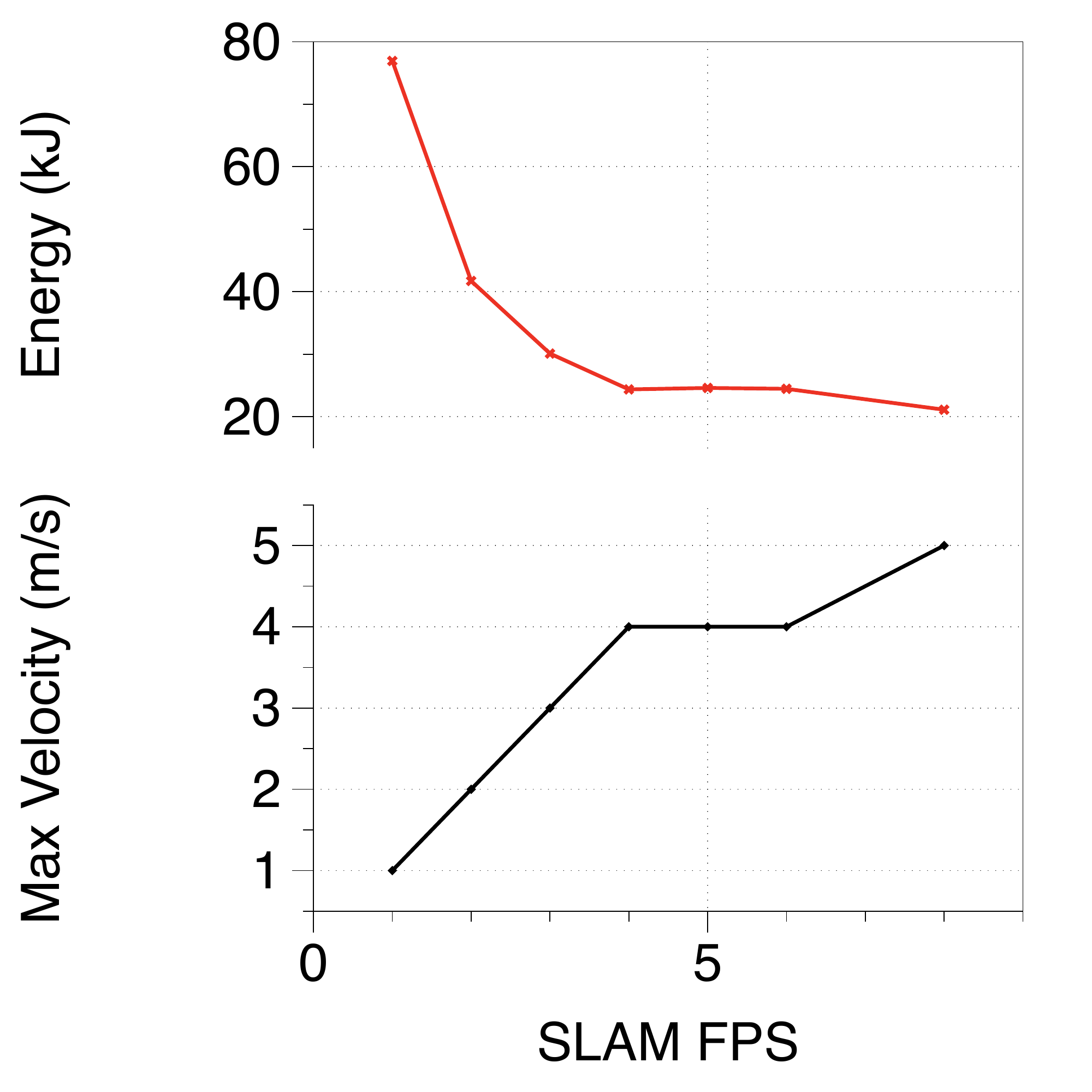}
    
    \end{subfigure}
\caption{Relationship between SLAM throughput (FPS) and maximum velocity and energy of UAVs.}
\label{fig:slam-velocity-energy}
\end{figure}

\paragraph{Improving Max Velocity by \textit{Reducing Perception Latency:}} Another way to improve velocity is to reduce perception processing time. 
The faster a drone wants to fly, the faster it must process its sensory feed
to extract the MAV's and its environment's relevant states. In other words, \textit{faster flights require faster perception}. This can be seen with perception related compute intensive kernels such as
Simulateneous Localization and Mapping (SLAM)~\cite{SLAM_survey}. SLAM localizes a MAV by tracking sets of features in the environment. Since a faster flight results in more rapid changes in the MAV's environment, fast flight can be problematic for this kernel leading to catastrophic effects such as permanent loss or a flight time increase (for example by backtracking due to re-localization). Minimizing or avoiding localization-related failures is highly favorable, if not necessary.

To examine the relationship between the compute, maximum velocity and localization failure, we evaluated a micro-benchmark in which the drone was tasked to follow a predetermined circular path of the radius 25~meters. For the localization kernel, we used ORB-SLAM2~\cite{orbslam2} and to emulate different compute powers, we inserted a sleep into the kernel. We swept velocities and sleep times and bounded the failure rate to 20\%. As \Fig{fig:slam-velocity-energy} shows, increasing FPS values from 1 to 8, which is enabled by more
compute, allows for an increase in maximum velocity from \SI{1}{\meter/\second} to \SI{5}{\meter/\second} (for a bounded failure rate), which shows that the maximum velocity is affected by perception latency.

\begin{figure*}[t!]
\centering
\includegraphics[width=\columnwidth]{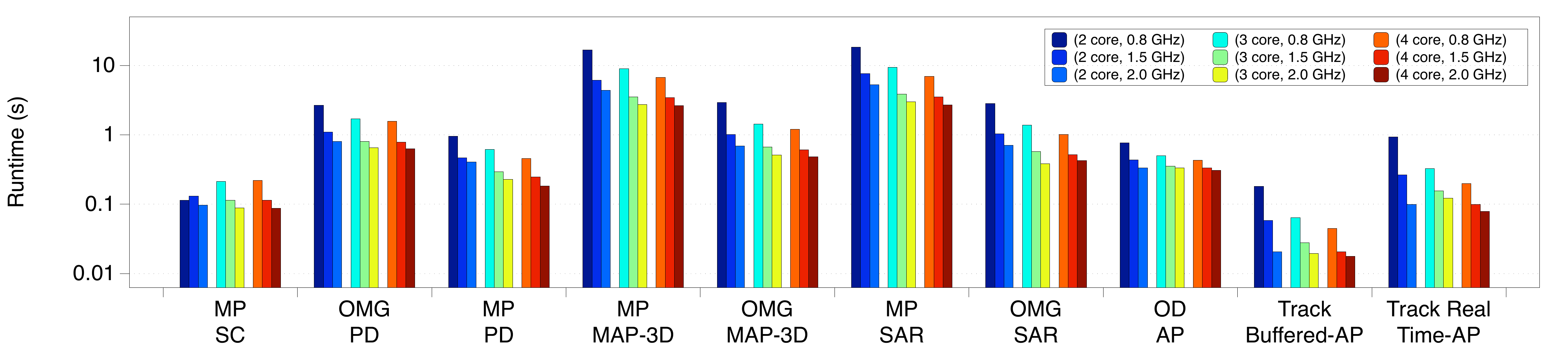}
\caption{Kernel breakdown for MAVBench. The abbreviations are as follows: \emph{OD-}Object detection, \emph{MP-}Motion Planning, \emph{OMG-}OctoMap Generation for kernels and \emph{SC-}Scanning, \emph{PD-}Package Delivery,  \emph{MAP-}3D Mapping, \emph{SAR-}Search and Rescue, and \emph{AP-}Aerial Photography for applications. The $x$-axis lists the kernel-application names and $y$-Axis represents the runtime in seconds. Each bar graph represents one of the configurations used in the hardware. The cores are varied from 2 to 4 and the frequency goes from from 0.8 GHz , 1.5 GHz or 2.2 GHz.}
\label{fig:kernel-breakdown}
\end{figure*}

Expanding on the microbenchmark insight from \Fig{fig:slam-velocity-energy}, 
we conducted a series of performance sensitivity analysis using processor core count and
frequency scaling.
We study the effect of compute on all of the MAVBench applications.
Average velocity and mission times of various operating points are profiled and presented as heat maps (Figures~\ref{fig:benchmarks_mission_velocity} and \ref{fig:benchmarks_mission_time}) for a DJI Matrice 100 drone. \emph{In general, compute can improve mission time by as much as 5X.} 

red\textit{Scanning:} In this application, we observe trivial differences for velocity and endurance across all three operating points (\Fig{fig:benchmarks:OPA:scanning:velocity}, \Fig{fig:benchmarks:OPA:scanning:time}) despite seeing a 3X boost in the motion planning kernel, i.e. lawn mower planning, which is its bottleneck (\Fig{fig:kernel-breakdown}). This is because, for this application, planning is only done once at the beginning of the mission and amortized over the rest of the mission time. For example, the overhead of planning for a five-minute flight is less than .001\%.

\textit{Package Delivery}:  As compute scales with the number of cores and/or frequency values, we observe a reduction of up to 84\% for the mission time (\Fig{fig:benchmarks:OPA:pd:time}). With frequency scaling, this improvement is due to the speed up of the sequential bottlenecks, i.e., motion planning and OctoMap generation kernel. On the other hand, there does not seem to be a clear trend with regard to core scaling, specifically between three and four cores. We conducted investigations and determined that the anomalies are caused by the non-real-time aspects of ROS, AirSim, and the TCP/IP protocol used for the communication between the companion computer and the host. Overall, we achieve up to 2.9X improvement in OctoMap generation which leads to maximum velocity improvement. It is important to note that although we also gain up to 9.2X improvements for the motion planning kernel, the low number of re-plannings and its short computation time relative to the entire mission time render its impact trivial. Overall the aforementioned improvements translate to up to 4.8X improvement in the average velocity. Therefore, mission time and MAV's total energy consumption are reduced.  
{ 
    \begin{figure}[t!] \centering \begin{subfigure}[t!]{.18\columnwidth} \centering \includegraphics[width=\columnwidth]{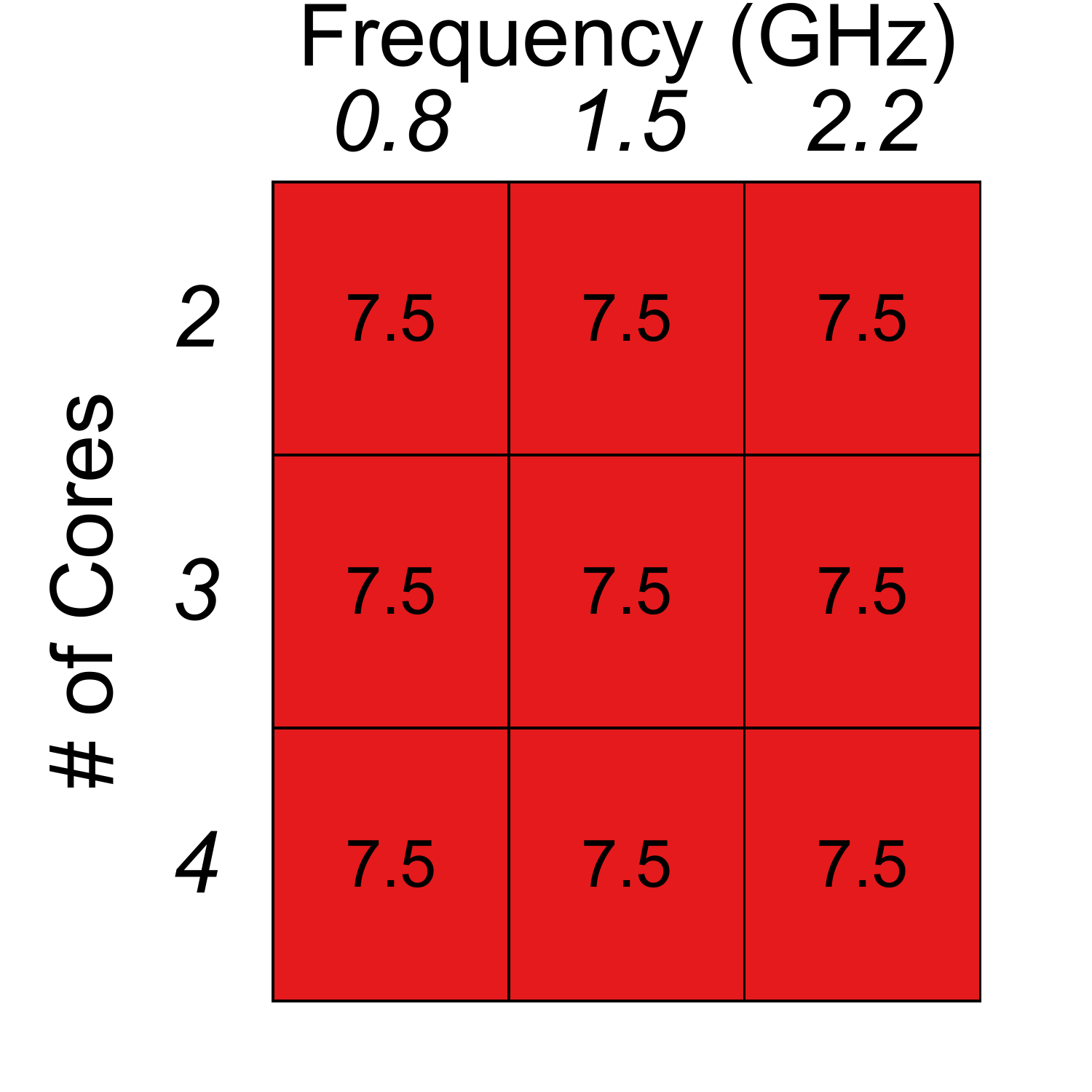} 
    \caption{\scriptsize Scanning.}
    \label{fig:benchmarks:OPA:scanning:velocity} \end{subfigure}%
    \begin{subfigure}[t!]{.18\columnwidth} \centering \includegraphics[width=\columnwidth]{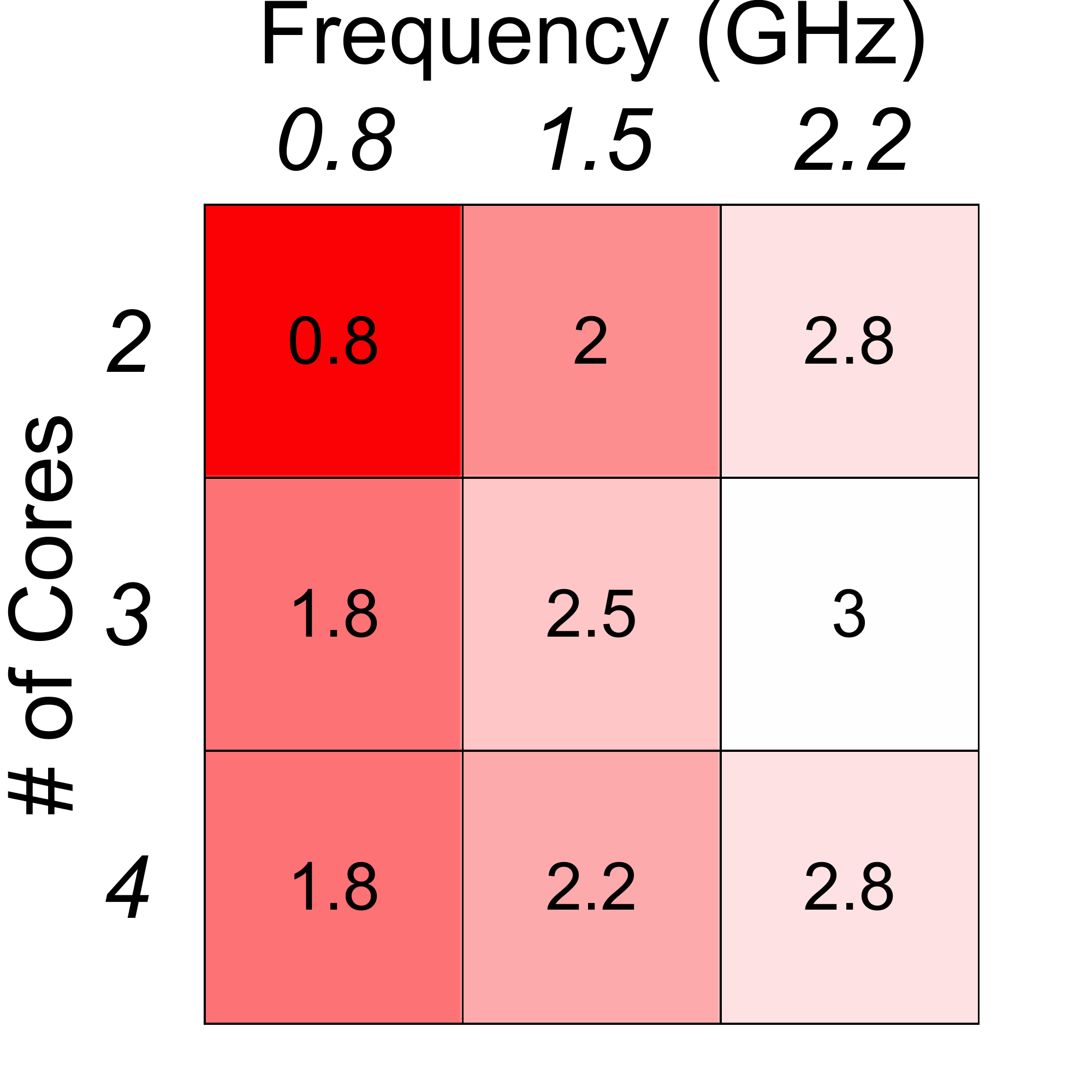}
\caption{\scriptsize Package Delivery.} \label{fig:benchmarks:OPA:pd:velocity} \end{subfigure}%
    \begin{subfigure}[t!]{.18\columnwidth} \centering
\includegraphics[width=\columnwidth]{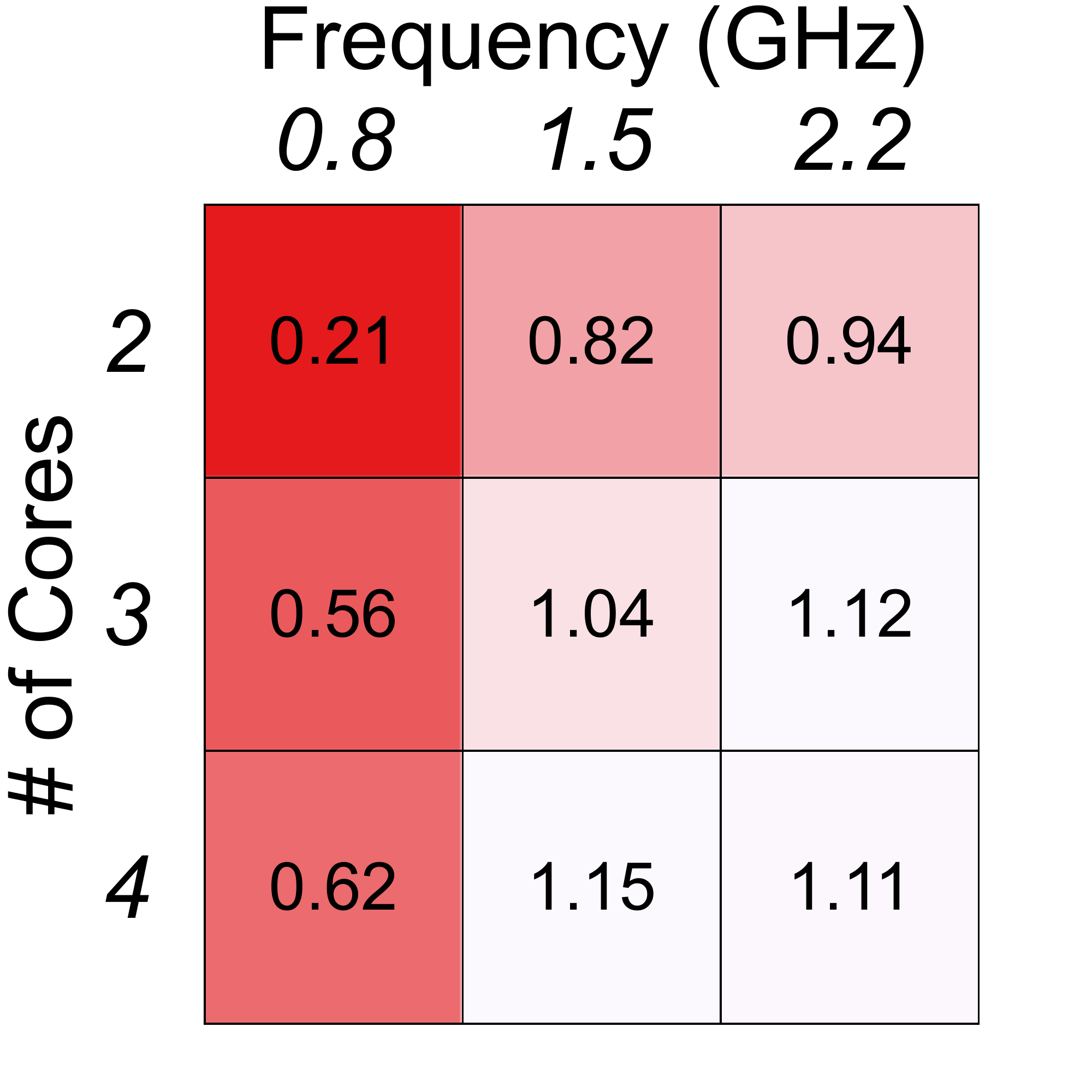}
\caption{\scriptsize 3D Mapping.} \label{fig:benchmarks:OPA:mapping:velocity} \end{subfigure} %
\begin{subfigure}[t!]{.18\columnwidth}
\centering \includegraphics[width=\columnwidth]{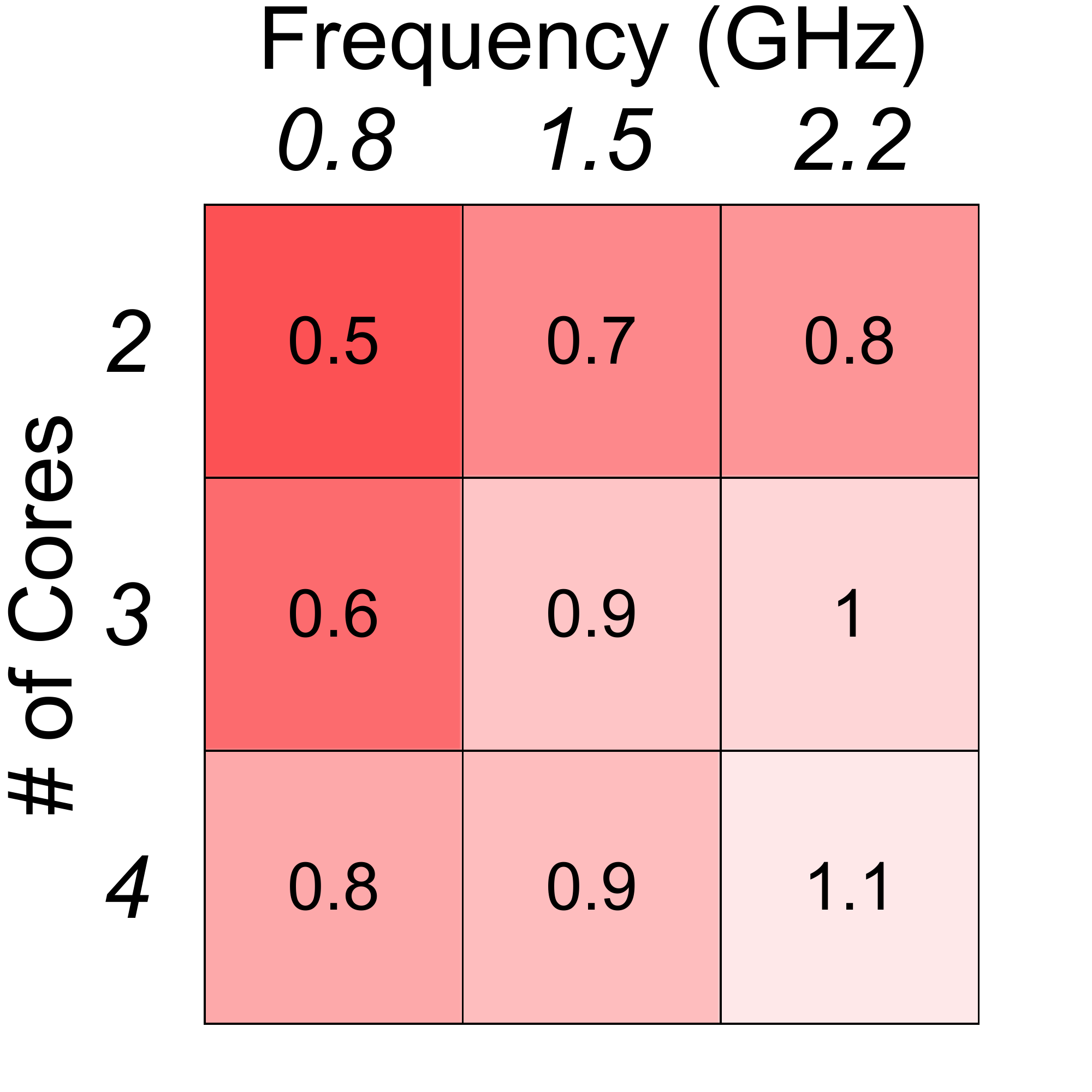} 
\caption{\scriptsize Search \& Rescue.} \label{fig:benchmarks:OPA:sar:velocity} \end{subfigure}%
\begin{subfigure}[t!]{.18\columnwidth} \centering \includegraphics[width=\columnwidth]{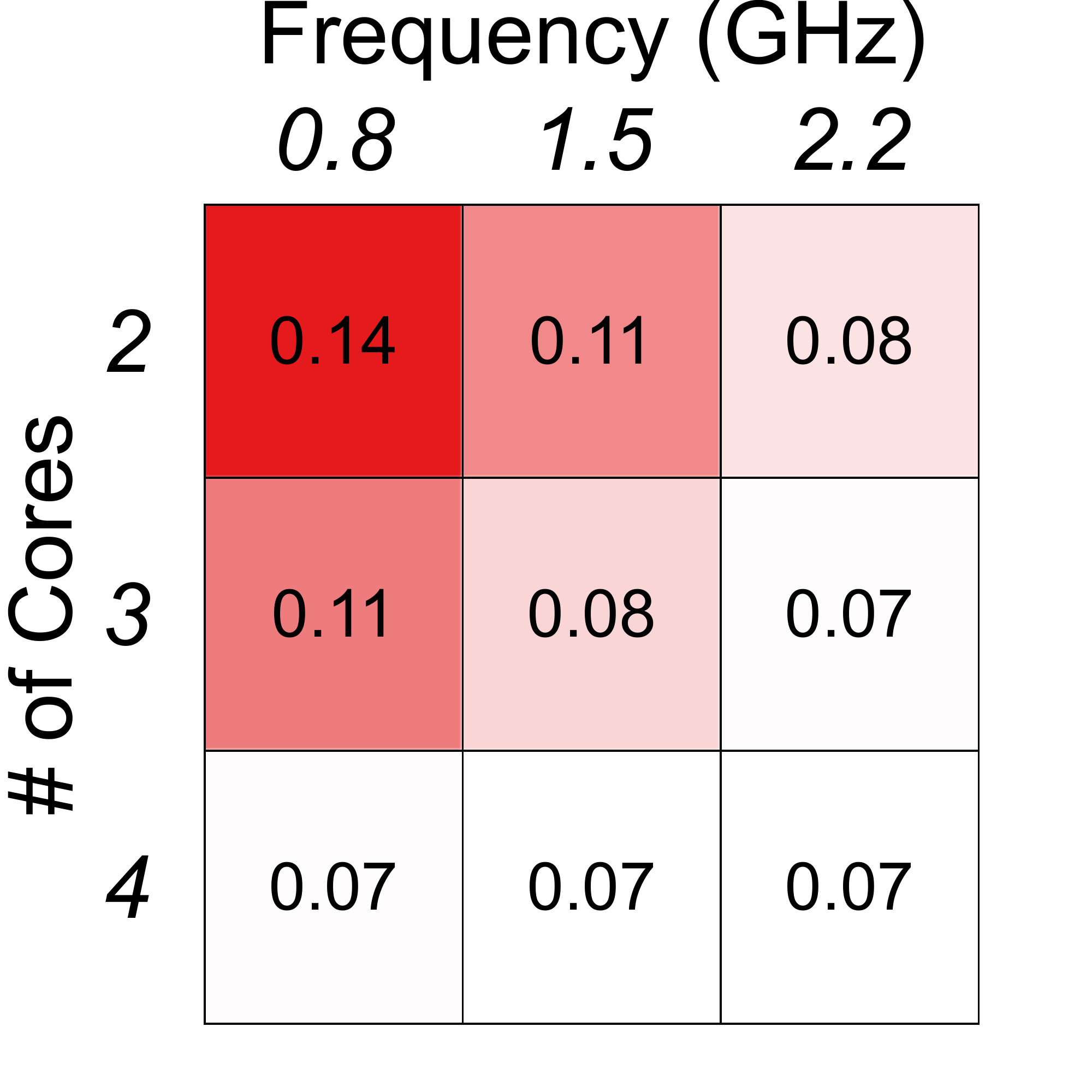}
    \caption{\scriptsize Aer Photography.}
     \label{fig:benchmarks:OPA:ap:velocity}
    \end{subfigure}
\caption{Core/frequency sensitivity analysis of mission average velocity for various benchmarks.}
\label{fig:benchmarks_mission_velocity}
\end{figure}

   \begin{figure}[t!]
   \centering
   \begin{subfigure}[t!]{.18\columnwidth}
    \centering 
    \includegraphics[width=\columnwidth]{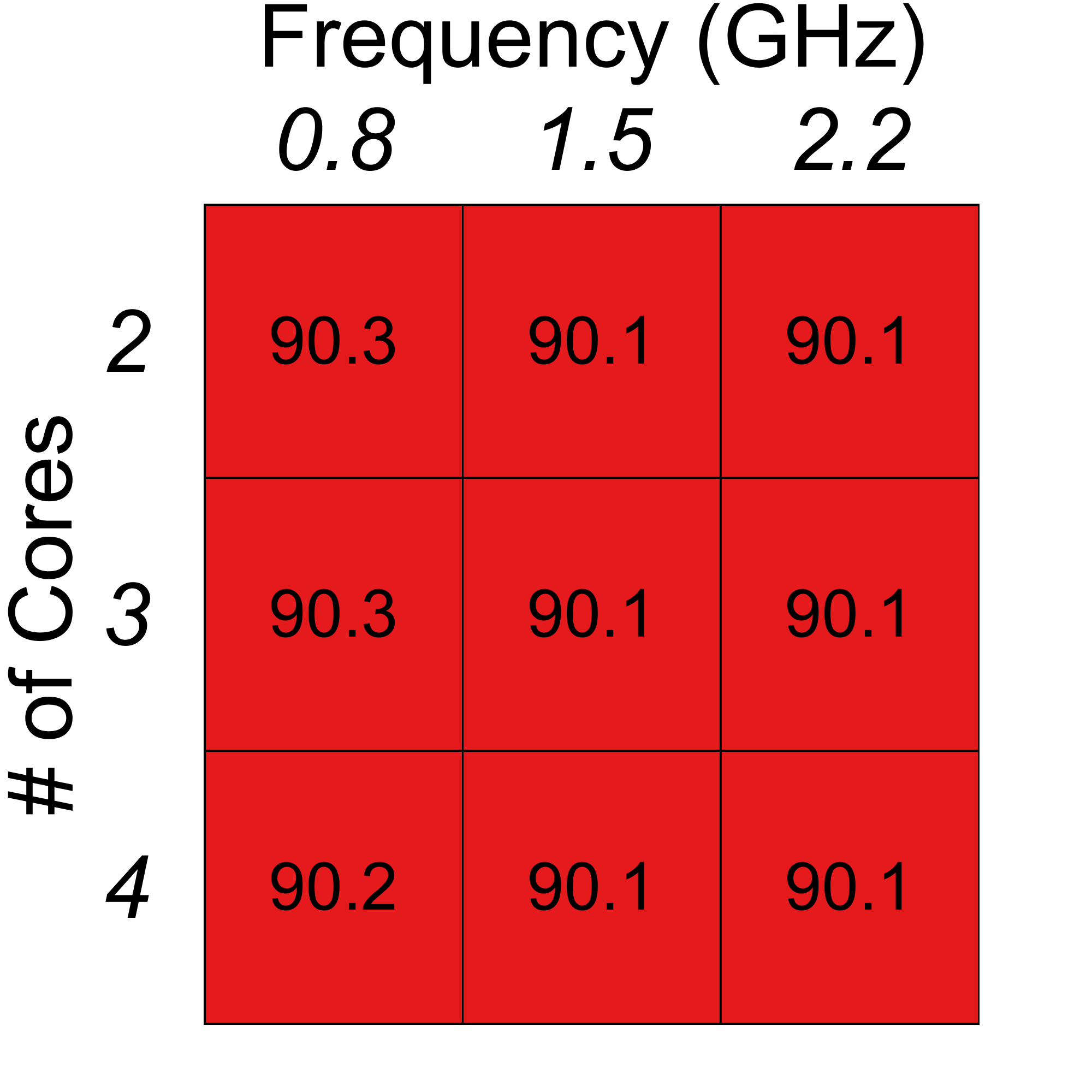}
    \caption{\scriptsize Scanning.}
    \label{fig:benchmarks:OPA:scanning:time}
    \end{subfigure}%
    \begin{subfigure}[t!]{.18\columnwidth}
    \centering
    \includegraphics[width=\columnwidth]{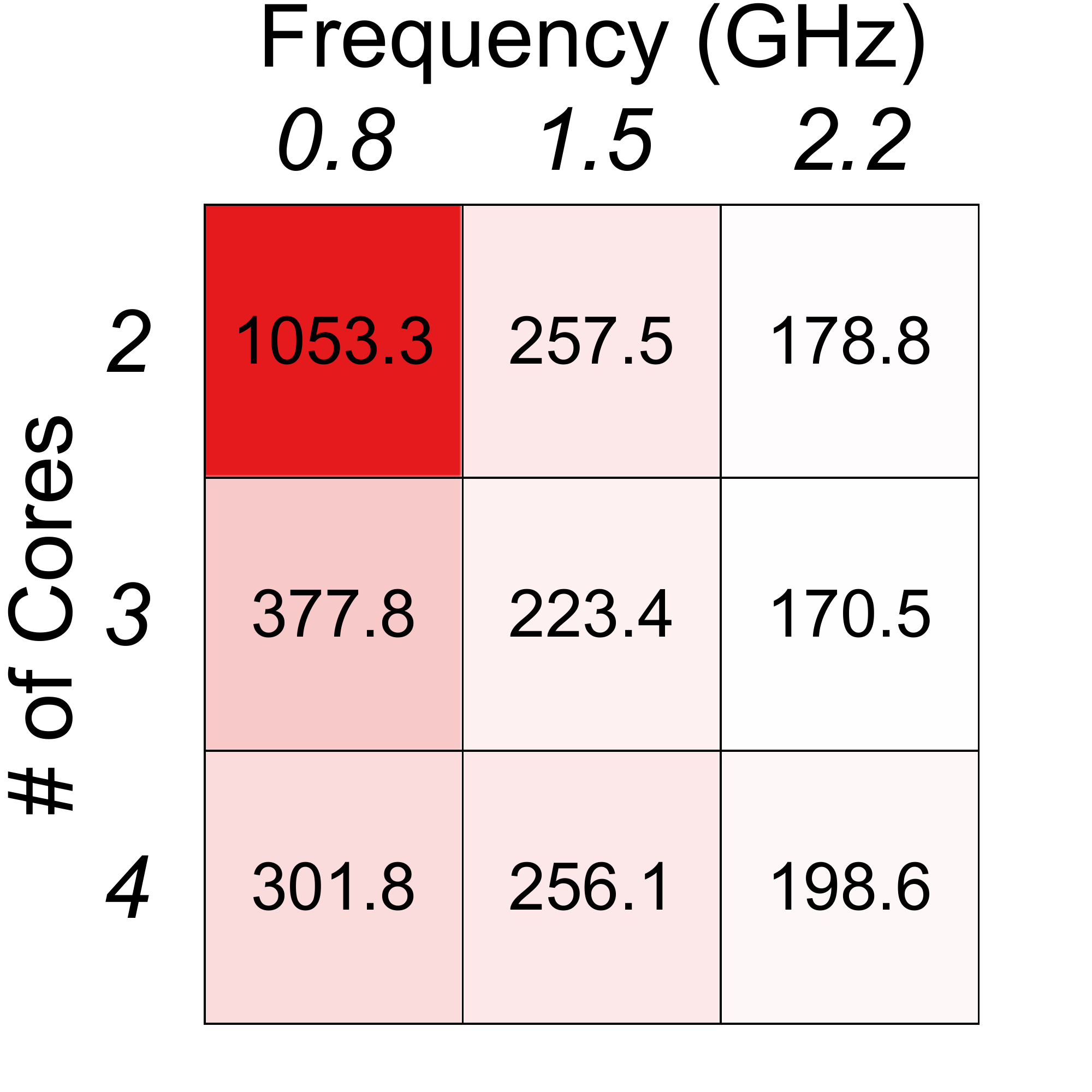}
    \caption{\scriptsize Package Delivery.}
    \label{fig:benchmarks:OPA:pd:time}
    \end{subfigure}%
    \begin{subfigure}[t!]{.18\columnwidth}
    \centering
    \includegraphics[width=\columnwidth]{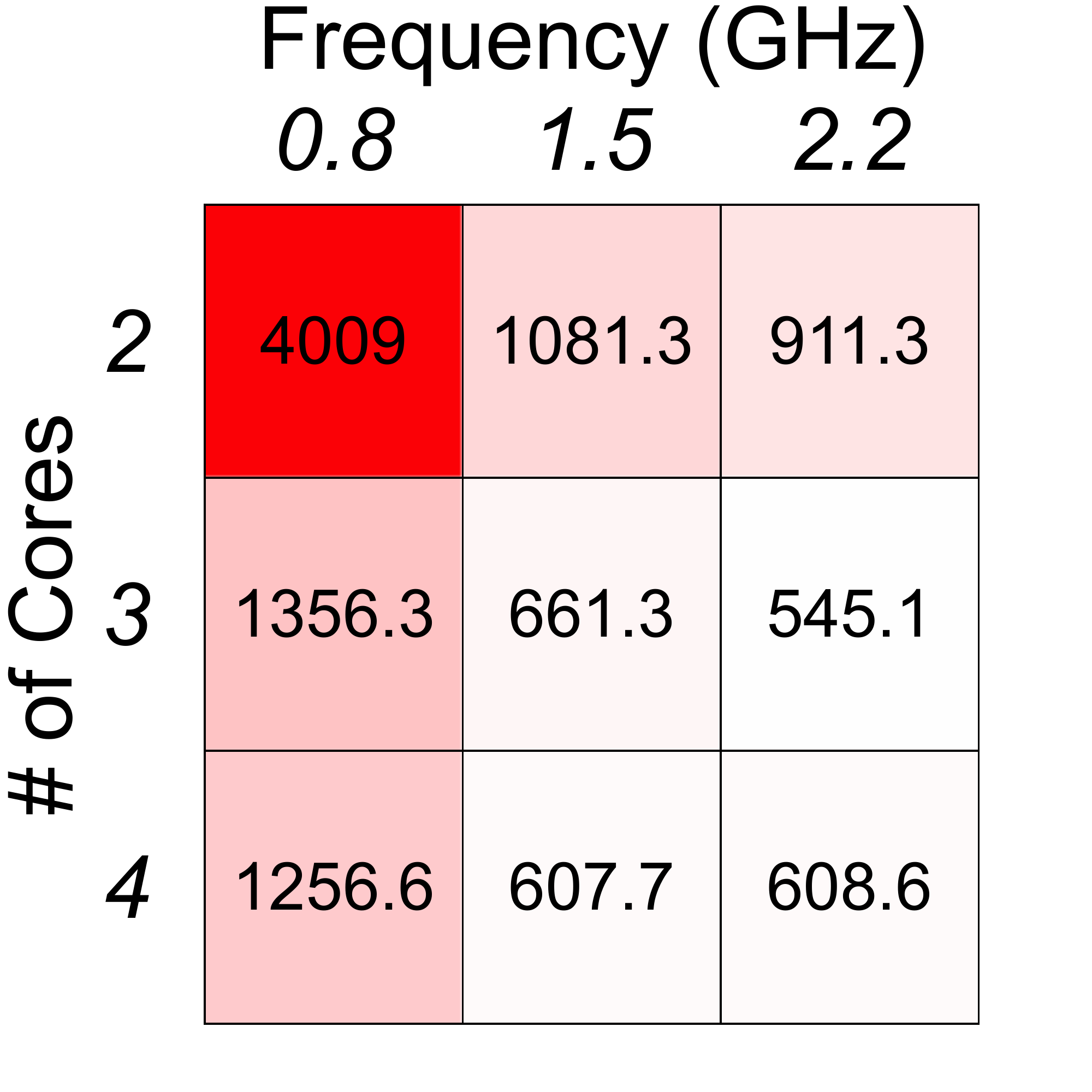}
    \caption{\scriptsize 3D Mapping.}
    \label{fig:benchmarks:OPA:mapping:time}
    \end{subfigure}%
     \begin{subfigure}[t!]{.18\columnwidth}
    \centering
    \includegraphics[width=\columnwidth]{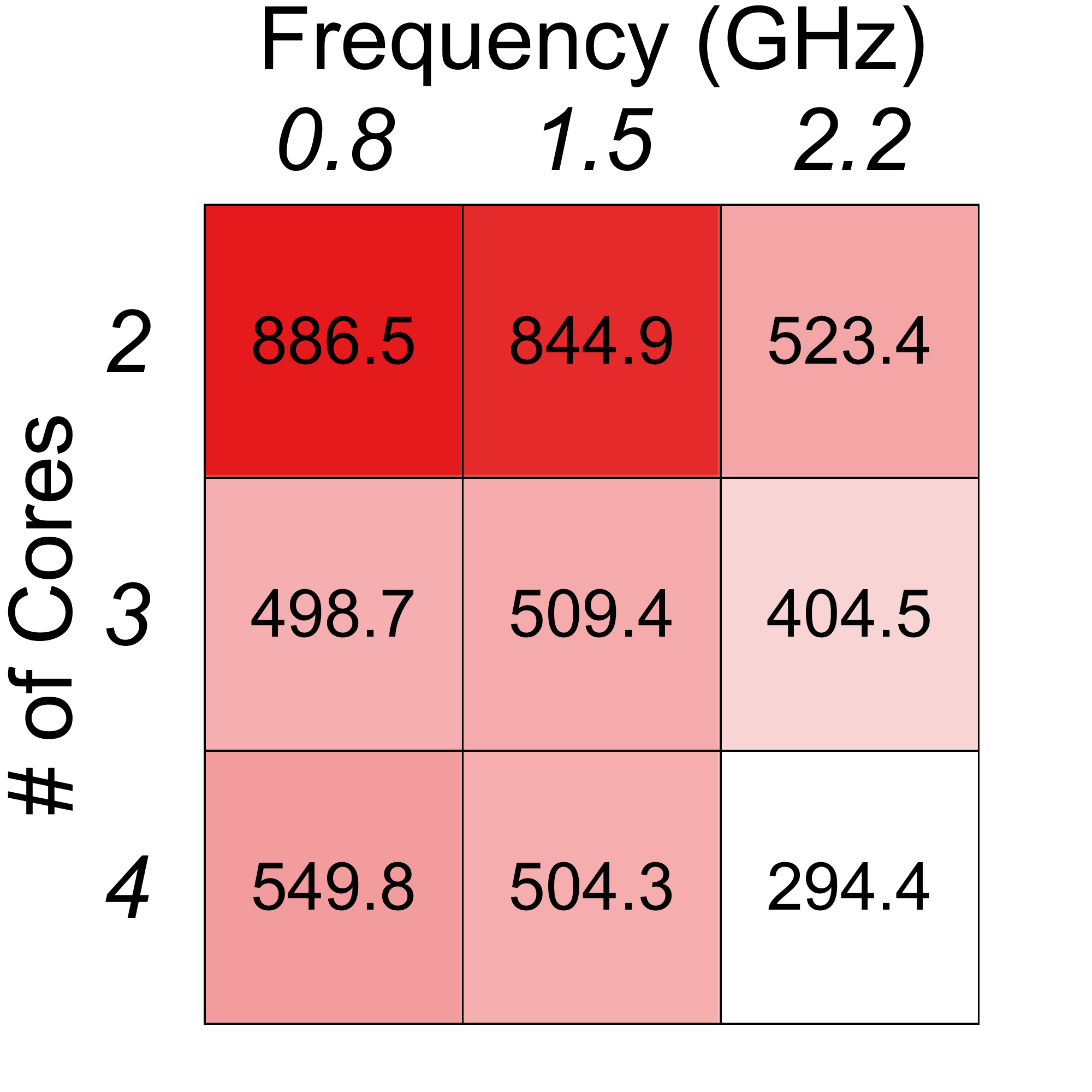}
    \caption{\scriptsize Search \& Rescue.}
    \label{fig:benchmarks:OPA:sar:time}
    \end{subfigure}%
    \begin{subfigure}[t!]{.18\columnwidth}
    \centering
    \includegraphics[width=\columnwidth]{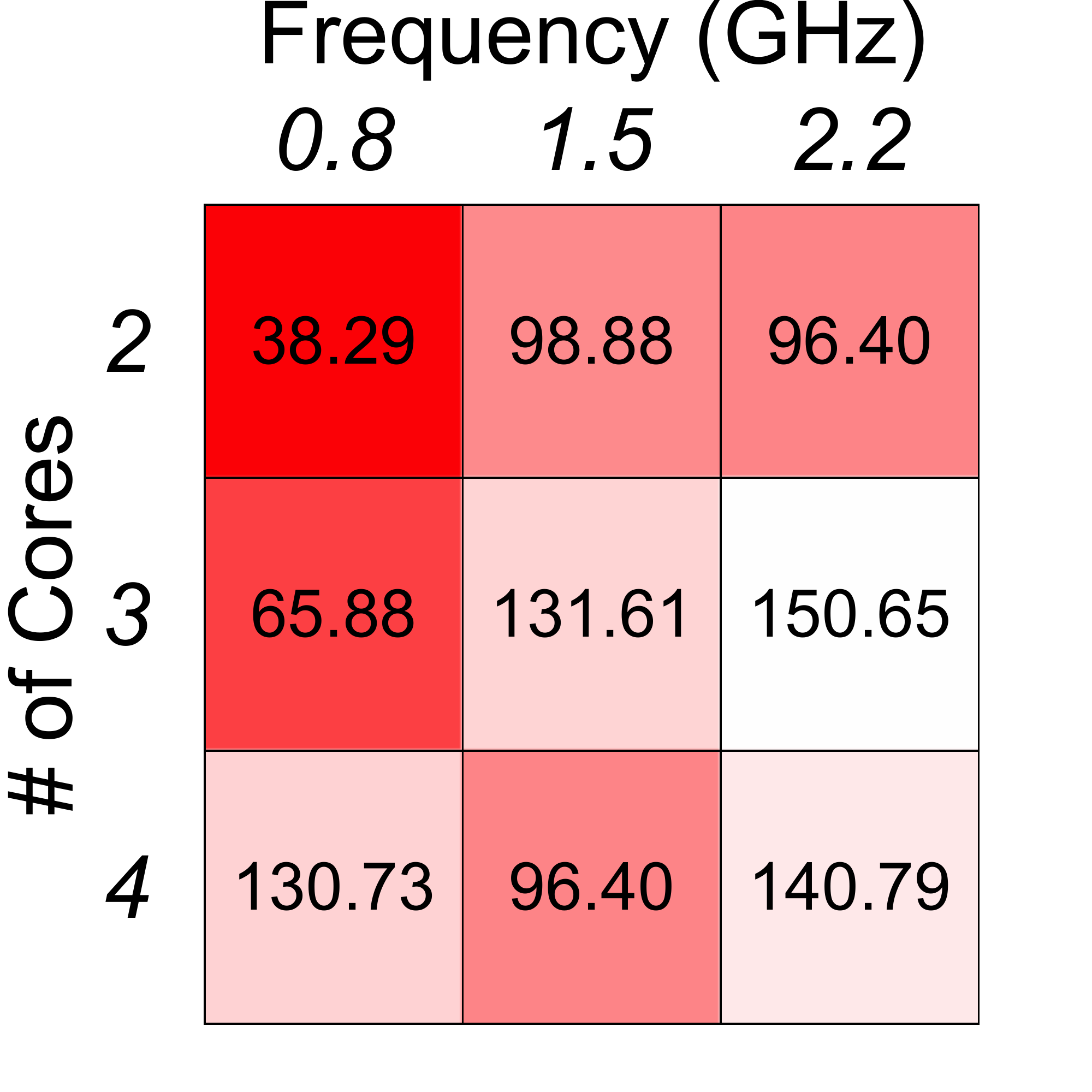}
    \caption{\scriptsize Aer Photography.}
    \label{fig:benchmarks:OPA:ap:time}
    \end{subfigure}
    \caption{Core/frequency sensitivity analysis of mission time for various benchmarks.}
    \label{fig:benchmarks_mission_time}
\end{figure}
}

\textit{3D Mapping:} As compute scales, mission time reduces by as much as 86\% (\Fig{fig:benchmarks:OPA:mapping:velocity}, \Fig{fig:benchmarks:OPA:mapping:time}). The concurrency present in this application (all nodes denoted by circles with a filled arrow connection or none at all in \Fig{fig:benchmarks:data-flow:mapping} run in parallel) justifies the performance boost from core scaling. The sequential bottlenecks, i.e., motion planning and OctoMap generation explains the frequency scaling improvements. We achieve up to 6.3X improvement in motion planning (\Fig{fig:kernel-breakdown}) which leads to hover time reduction. We achieve a 6X improvement in OctoMap generation which leads to a maximum velocity improvement. Combined the improvements translate to a 5.3X improvement in average velocity. Improving average velocity reduces mission time.

\textit{Search and Rescue:} As compute scales, we see a reduction of up to 67\% for the mission time (\Fig{fig:benchmarks:OPA:sar:velocity}, \Fig{fig:benchmarks:OPA:sar:time}). Similar to the case of 3D Mapping, more compute allows for the reduction of hover time and an increase in maximum velocity which contribute to the overall reduction in mission time and energy. In addition, a faster object detection kernel prevents the drone from missing sampled frames during any motion. We achieve up to 1.8X, 6.8X, and 6.6X speedup for the object detection, motion planning and OctoMap generation kernels, respectively. In aggregate, these improvements translate to a 2.2X improvement in the MAV's average velocity.

\textit{Aerial Photography:} As compute scales, we observe an improvement of up to 53\% and 267\% for \emph{error} and mission time, respectively (\Fig{fig:benchmarks:OPA:ap:velocity}, \Fig{fig:benchmarks:OPA:ap:time}). Note that this application is a special case. In aerial photography, as compared to other applications, higher mission time is more desirable than a lower mission time. The drone only flies while it can track the person; hence, a longer mission time means that the target has been tracked for a longer duration. In addition to maximizing the mission time, error minimization is also desirable for this application. We define error as the distance between the person's bounding box (provided by the detection kernel) center to the image frame center.  Clock and frequency improvements translate to 2.49X and 10X speedup for the detection and tracking kernels and that allows for longer tracking with a lower error.


\subsection{Compute Mass Impact on Mission Time}
\label{sec:comp_mass_impact_on_mission_time}
\begin{figure}[t!]
\begin{subfigure}[t]{0.45\textwidth}
\centering
\includegraphics[width=0.6\columnwidth]{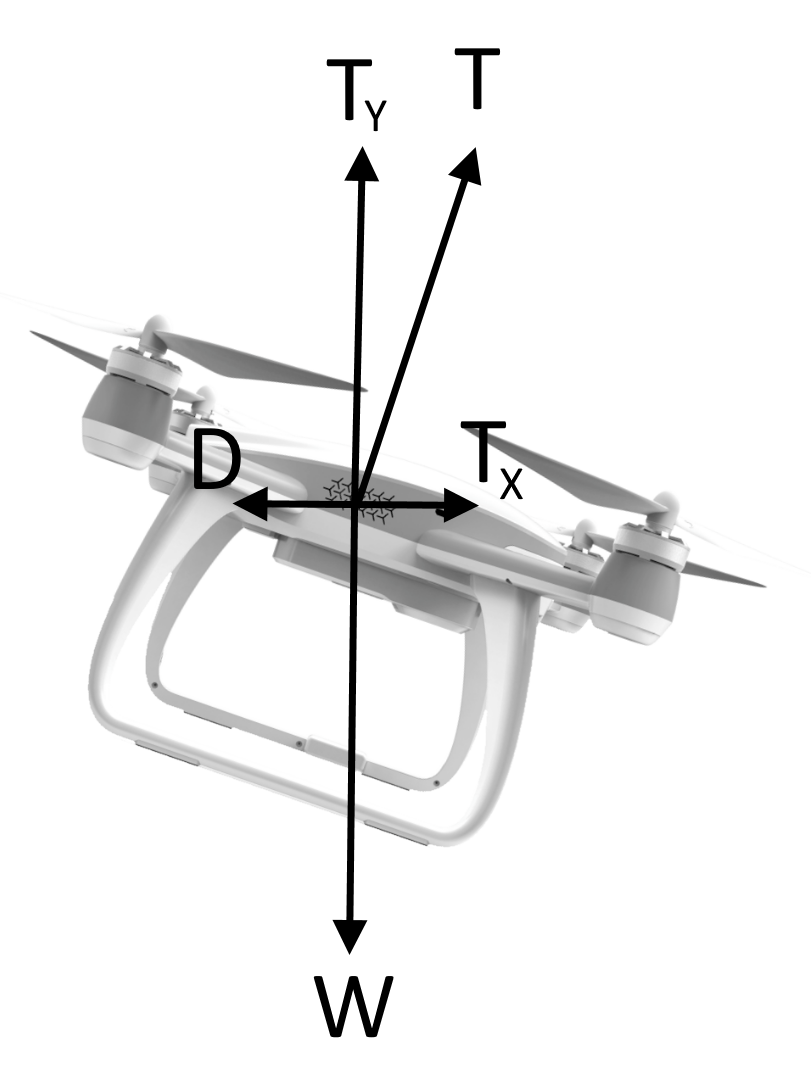}
\caption{2D free body diagram.}
\label{fig:quad_FBD_2D}
\end{subfigure}
\begin{subfigure}[t]{0.45\textwidth}
\centering
\includegraphics[width=0.7\columnwidth]{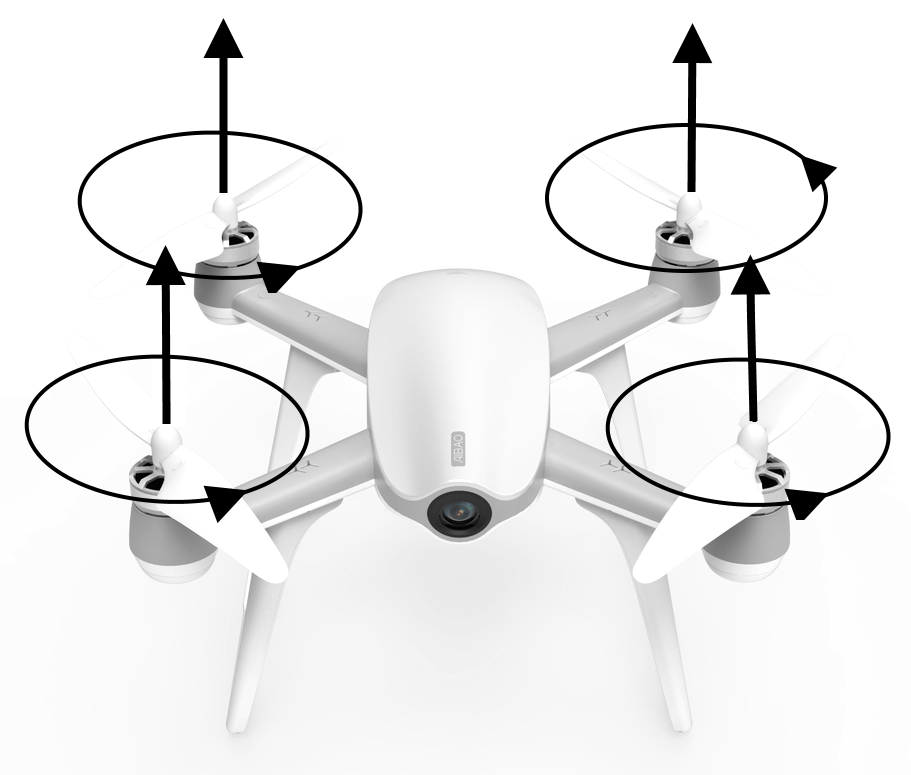}
\caption{3D free body diagram.}
\label{fig:quad_FBD_3D}
\end{subfigure}
\caption{Forces acting on a quadcoptor. Knowing these forces is necessary in understanding how cyber and physical quantities impact one another.}
\end{figure}

Compute impacts mission time through its physical mass (mass cluster shown in Figure~\ref{fig:CIG_mass_paths} with green-color/double-sided paths). Increasing onboard compute affects the total weight of the MAV, which impacts the MAV's maximum acceleration and velocity, and consequently, that affects mission metrics (i.e., flight time).
To understand this impact, first, we need to understand the forces acting on a quadcopter. The free-body diagrams shown in Figures~\ref{fig:quad_FBD_2D}
and~\ref{fig:quad_FBD_3D} illustrate these forces in steady flight.\footnote{In a steady flight, the vehicle's linear and angular velocities are constant.}
The force generated by the motor is called thrust. In steady flight, the $y$ component of this thrust vector ($T_y$) compensates gravity ($W$) to keep the drone in the air while the $x$ component ($T_x$) combats the air drag ($D$). When decelerating, both thrust and drag act in the direction opposite to flight and make the vehicle slow down. The resulting maximum acceleration can be derived from Equation~\ref{eq:mass_acceleration}, where $m$ is the total vehicle mass, $D$ is the total drag and $T_{x_{max}}$ the maximum applicable thrust in the horizontal direction (negative when slowing down). Note that since we would like to calculate the worst case deceleration, we remove drag from the equation.

\begin{equation}
\label{eq:mass_acceleration}
    a_{max} = \frac{T_{x_{max}}-D}{m}
\end{equation}
\setlength{\belowcaptionskip}{-1ex}

Adding more mass to the drone demands a higher portion of thrust to battle weight, i.e., the drone requires higher $T_y$. Given the limited total thrust ($T_{max}$) that the motors can generate, a higher $T_y$ leaves the drone with less $T_x$ to accelerate with (Equation~\ref{eq:t_max}).
\begin{equation}
\label{eq:t_max}
    T_{x} = ~\sqrt[]{T_{max}^2 - T_{y}^2}
\end{equation}

Putting this all together, Equation~\ref{eq:mass_all} captures the relationship between mass and acceleration.

\begin{equation}
\label{eq:mass_all}
\boxed{
    a_{max} = \frac{~\sqrt[]{T_{max}^2 - W^2}}{m}}
\end{equation}
\setlength{\belowcaptionskip}{-1ex}

Increasing onboard compute capability increases the weight of a MAV. As the amount of onboard compute increases, the thermal design power ($TDP$) escalates. Higher TDP requires more cooling effort and that automatically necessitates a more robust and heavier cooling system. 

To study the effect of mass, we consider four different system-on-chips (SoCs)
each with a different compute capability. Table~\ref{compute_platform_data} shows  
the mass associated with the different chipsets.\footnote{The weights are collected either through direct inquiry of the developing company or a thorough online search. We could not find a heat sink for the Xavier online, hence we exacted the heat sink weight through linear interpolation of the rest of data points.} 
We observe that the overall compute subsystem's weight vary from \SI{144}{\g} to \SI{1109}{\g}, i.e., a 7.9X increase, increasing the total MAV mass from \SI{2544}{\g} to \SI{3509}{\g}, i.e., a 1.4X increase.

\begin{table*}[!b]
\caption{Characteristics of the compute platforms we use for mass and holistic experiments. The platforms range from light yet high-end embedded platforms such as Jetson TX2 to heavy yet powerful high-end server such as an Intel Core i9.}
\label{compute_platform_data}
\resizebox{\columnwidth}{!}{    
\begin{tabular}{|l|c|c|c|c|c|c|c|c|c|c|}
\hline
\multicolumn{10}{|c|}{MAV Compute subsystem}                                                                                                                                                                                                                                                                                                                                                                                                                                                                                                                                                                                                                                          & \begin{tabular}[c]{@{}c@{}}MAV \\ Robot \\ Complex\end{tabular}     \\ \hline
\multicolumn{3}{|c|}{Platform}                                                                                                                    & \multicolumn{3}{c|}{Performance}                                                                                                                                                                                                                   & \multirow{2}{*}{\begin{tabular}[c]{@{}c@{}}Thermal\\  Power \\ (W)\end{tabular}} & \multicolumn{3}{c|}{Mass}                                                                                                                                                                 & \multirow{2}{*}{\begin{tabular}[c]{@{}c@{}}Mass\\ (g)\end{tabular}} \\ \cline{1-6} \cline{8-10}
Name          & \begin{tabular}[c]{@{}c@{}}Number\\  of \\ Cores\end{tabular} & \begin{tabular}[c]{@{}c@{}}CPU \\ Frequency\\  (GHz)\end{tabular} & \multicolumn{1}{l|}{\begin{tabular}[c]{@{}c@{}}Latency \\     (s)\end{tabular}} & \multicolumn{1}{l|}{\begin{tabular}[c]{@{}c@{}}Throughput\\      (Hz)\end{tabular}} & \multicolumn{1}{l|}{\begin{tabular}[c]{@{}c@{}}Total\\   (s)\end{tabular}} &                                                                                  & \begin{tabular}[c]{@{}c@{}}Board and\\  processor\\  (g)\end{tabular} & \begin{tabular}[c]{@{}c@{}}Heat \\ Sink\\ (g)\end{tabular} & \begin{tabular}[c]{@{}c@{}}Total \\ (g)\end{tabular} &                                                                     \\ \hline
i9-9940X      & 14                                                            & 1.2                                                               & .243                                                                            & 13.3                                                                                & .318                                                                       & 165                                                                              & 506                                                                   & 603                                                        & 1109                                                 & 3509                                                                \\ \hline
i7-4790K      & 7                                                             & 2                                                                 & .426                                                                            & 4.46                                                                                & .65                                                                        & 88                                                                               & 483                                                                   & 285                                                        & 768                                                  & 3168                                                                \\ \hline
Jetson Xavier & 8                                                             & 2.2                                                               & .586                                                                            & 3.25                                                                                & .894                                                                       & 30                                                                               & 280                                                                   & 100                                                        & 380                                                  & 2780                                                                \\ \hline
Jetson Tx2    & 6                                                             & 2                                                                 & .717                                                                            & 2.49                                                                                & 1.119                                                                      & 15                                                                               & 85                                                                    & 59                                                         & 144                                                  & 2544                                                                \\ \hline
\end{tabular}
}
\vspace{2pt}
\end{table*}

Using Equation~\ref{eq:mass_acceleration} and Table~\ref{compute_platform_data}, we study the impact of compute's added mass on the mission time (mass cluster denoted with green-color/double-sided impact paths in Figure~\ref{fig:CIG_mass_paths}) of a DJI Quadcopter, assuming it is equipped with the different compute platforms. We also study the effects of different environmental congestion levels (e.g., number of obstacles in the flight path). To study congestion levels, we introduce the notion of ``slow down ratio'' (SDR). This ratio is calculated as $v_{max}$, denoting maximum allowed velocity of a MAV, over $v_{avg}$, average velocity which the MAV maintains across its mission (Equation~\ref{eq:slow_down}), and it indicates environment's congestion. The higher the environment congestion, the higher the slow down ratio, resulting in a lower average velocity relative to maximum allowed velocity. This is because a congested space does not allow the drone to reach its top speed for long periods of time due to the frequent slowdowns caused by the numerous obstacles. 
\begin{equation}
\label{eq:slow_down}
    Slow\;Down\;Ratio\;(SDR) = \frac{V_{max}}{V_{avg}}
\end{equation}

\begin{figure}[t!]
\centering
     \begin{subfigure}{.4\linewidth}
   \centering
   \includegraphics[width=.9\linewidth]{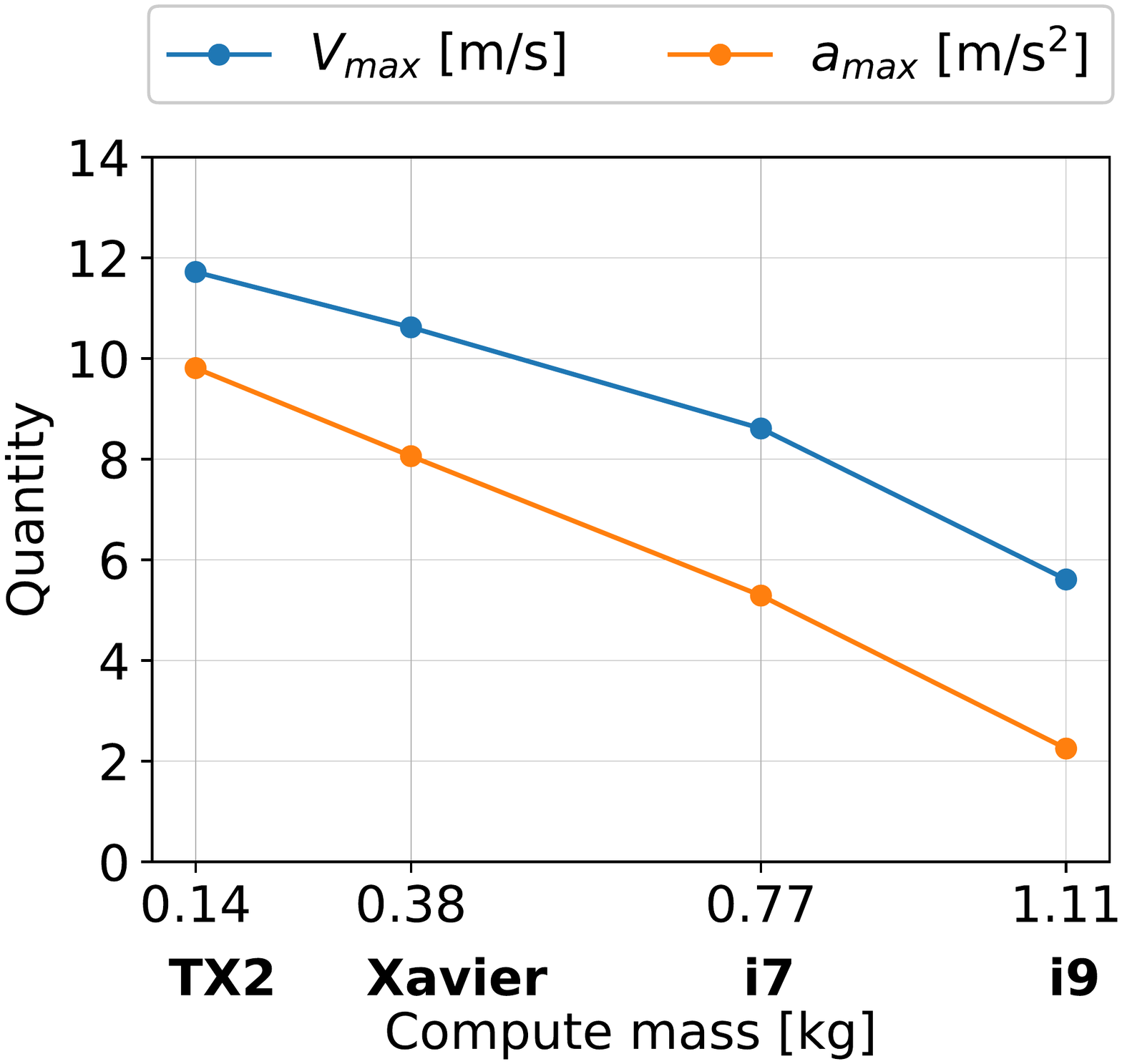}
   \vspace{-20pt}
   \caption{Impact on velocity and acceleration.}
   \label{fig:vel_acc_mass_impact}
    \end{subfigure}%
    \begin{subfigure}{.4\linewidth}
  \vspace{-7pt}
  \centering
   \includegraphics[width=.9\linewidth]{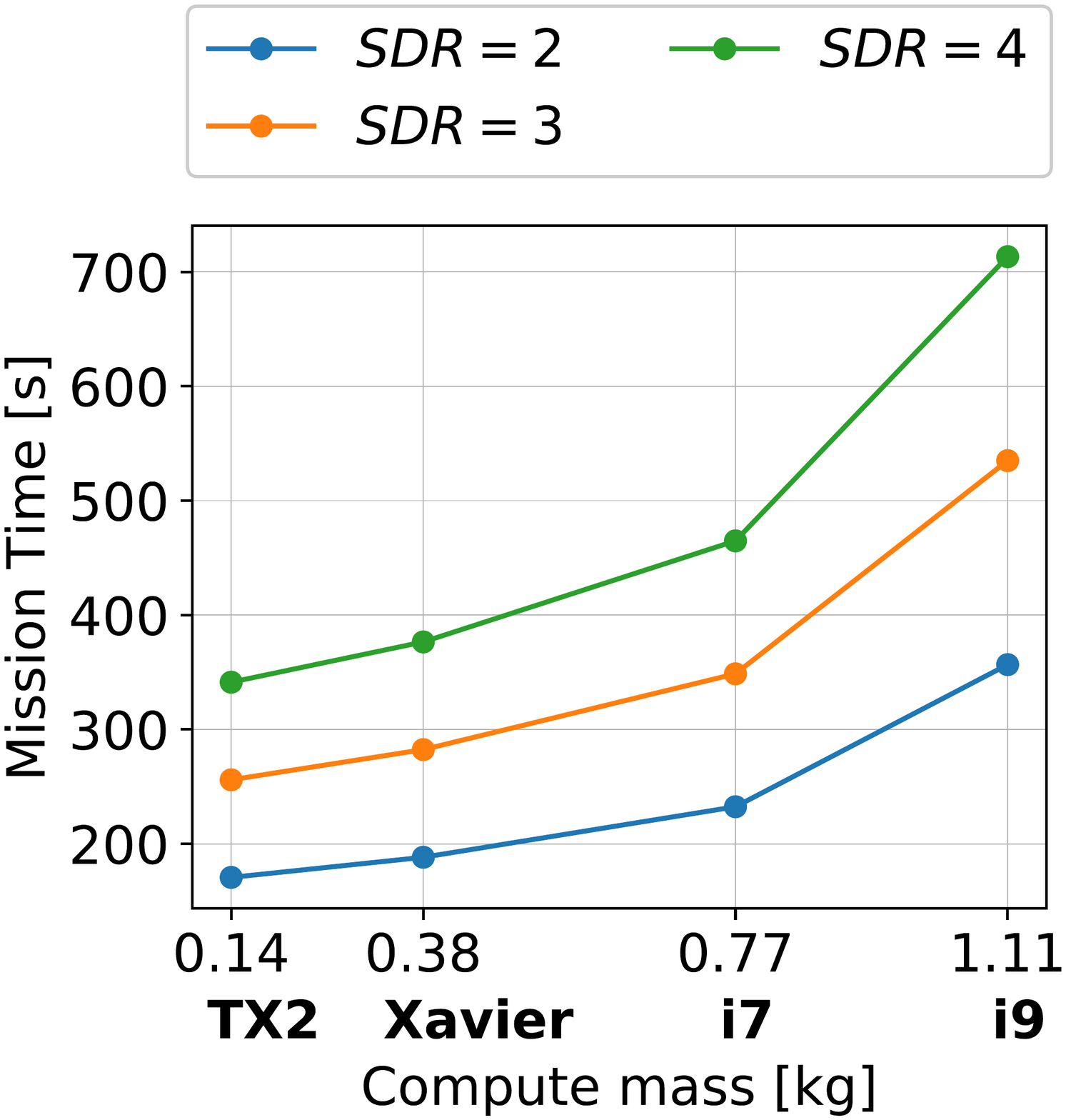}
    \vspace{-20pt}
    \caption{Impact on mission time.}
    \label{fig:comp_mt_mass_impact}
   \end{subfigure}
   \caption{Impact of compute mass, a physical quantity, on velocity, acceleration and mission time.}
      \label{fig:comp_stop_distance}
\end{figure}

Compute platform mass can considerably impact acceleration and velocity. Figure~\ref{fig:vel_acc_mass_impact} shows the impact of compute platform mass on $a_{max}$ and $v_{max}$. Different points on a line correspond to the different platforms in Table~\ref{compute_platform_data}. We see an acceleration of \SI{9.8}{\meter/\second\squared} vs. \SI{2.3}{\meter/\second\squared}, i.e., a 4.4X difference, for our lightest (TX2) and heaviest (i9) platforms, respectively. This difference leads to a velocity of \SI{11.7}{\meter/\second} vs. \SI{5.6}{\meter/\second}, respectively, i.e., a 2.1X difference, for our lightest and heaviest platform.


Since compute mass impacts acceleration and velocity, it impacts mission time. Figure~\ref{fig:comp_mt_mass_impact}
shows this impact. Different points on a line correspond to the different platforms, and different lines correspond to different slow down ratios (SDRs). The acceleration and velocity differences discussed previously result in mission time of \SI{341}{\second} and \SI{713}{\second}, i.e., a 2X difference for TX2 and i9, respectively (for the most congested environment with the SDR of 4). Note that higher environment congestion grows the difference between the two extreme designs. For example, the mission time of the best and worse designs for SDR of 2 are \SI{170}{\second} and \SI{356}{\second} resulting in a difference of \SI{186}{\s} whereas same mission times for SDR of 4 are \SI{341}{\second} and \SI{713}{\second} resulting in a difference of \SI{372}{\s}. 

In summary, given these numbers, we conclude that lighter compute systems are of high value. Since the compute induced mass is mainly the result of cooling solutions to meet $TDP$, system designers need to target power-efficient designs. Furthermore, the analysis shows that dealing with congested spaces (such as in an indoor search and rescue mission) requires more compute efficiency from a mass perspective, and thus warrants greater demand for attention from system designers.

\section{Compute Impact on Mission Energy}
\label{sec:comp_mission_energy}
Compute can impact mission energy through its performance, added power and mass.
We dive deep into such impacts and study each impact cluster separately to isolate their effect so that we gain better insights into their inner working. First, we explain the impact paths in the power cluster (Figure~\ref{fig:CIG_power_paths}, red-color/fine-grained-dashed paths). These are the paths originate with compute power. 
Then,
we explain performance cluster impacts paths (Figure~\ref{fig:CIG_performance_paths}, blue-color/coarse-grained-dashed paths). These  paths originate with compute performance quantities such as
sensing-to-actuation latency and throughput. Finally, we discuss the effect of mass cluster impact paths (Figure~\ref{fig:CIG_mass_paths}, green-color/double-sided paths). These paths originate with compute mass.

\subsection{Compute Power Impact on Mission Energy}
\label{sec:comp_power_impact_on_mission_energy}
Compute impacts MAV's overall energy consumption through its power consumption (power cluster shown in Figure~\ref{fig:CIG_power_paths} with red-color/fine-grained paths). The more the power consumption associated with the compute platform, the more the overall MAV's energy consumption.

\begin{figure}[!t]
\begin{subfigure}{.4\linewidth}
\centering
    \includegraphics[trim=0 2cm 0 0, clip, height=.8\columnwidth]{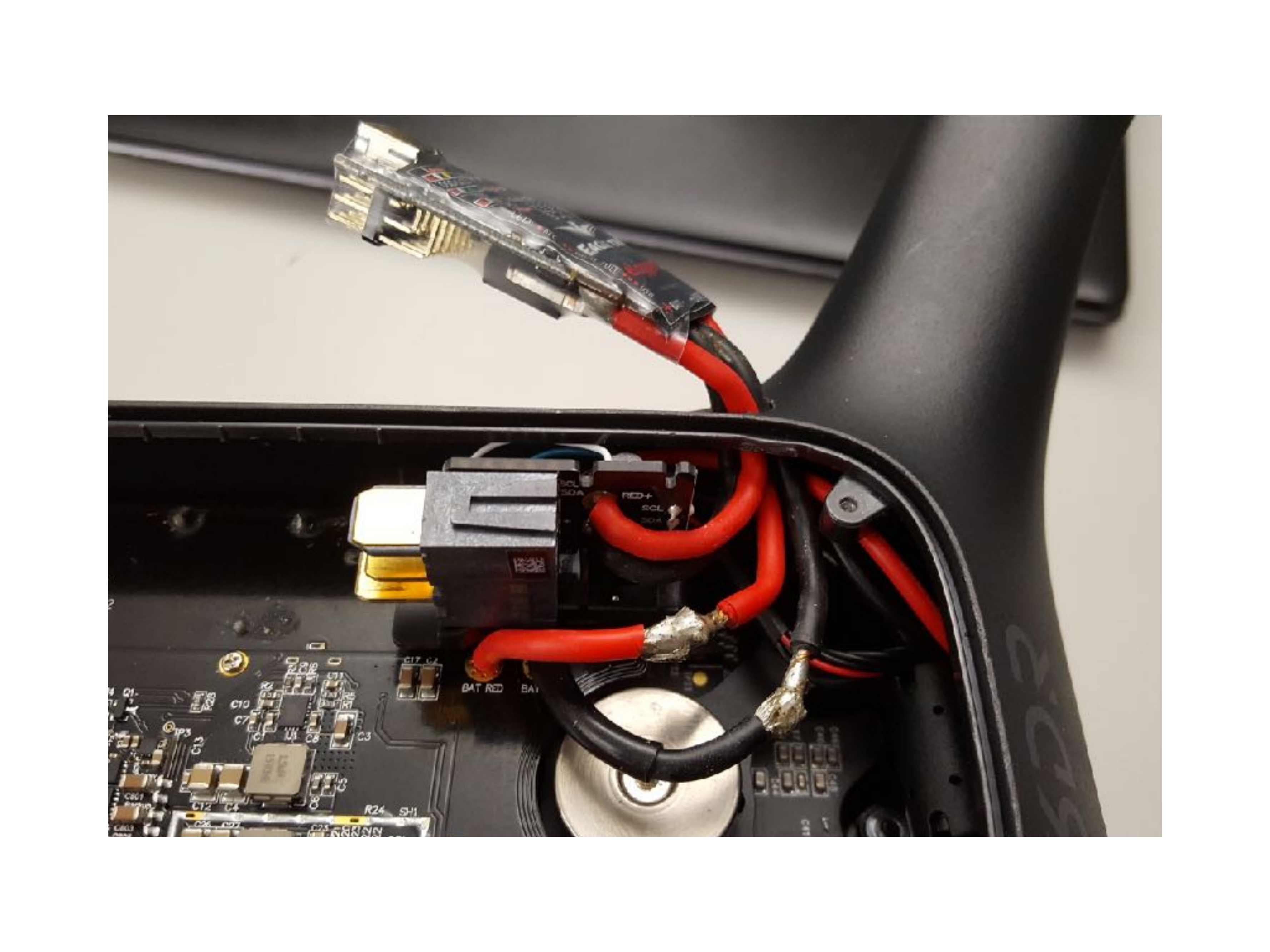}
    \caption{eLogger inside view.}
    \label{fig:elogger_placement}
\end{subfigure}%
\begin{subfigure}{.4\linewidth}
\centering
    \includegraphics[trim=0 2cm 0 0, clip, ,height=.8\columnwidth]{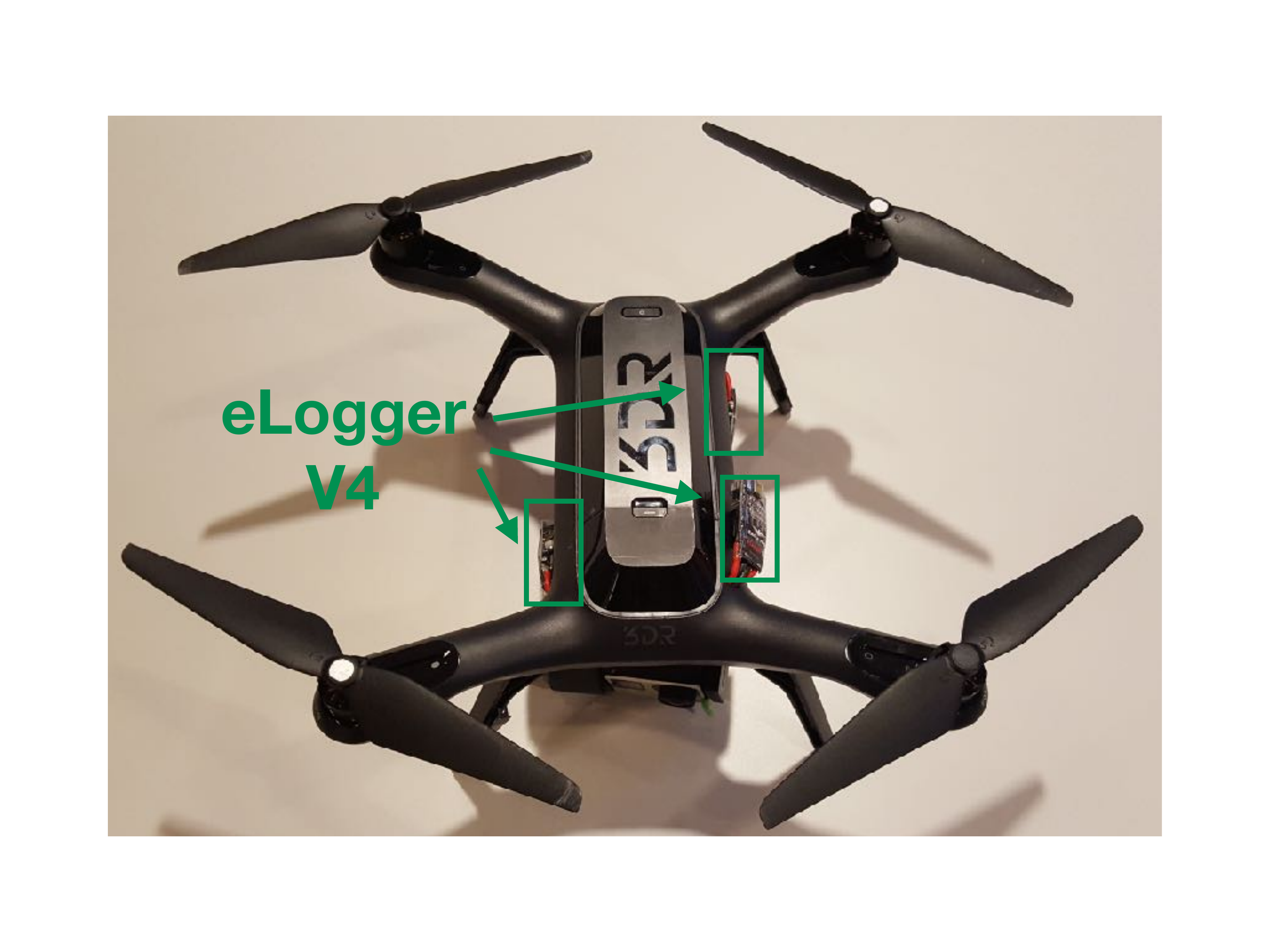}
    \caption{eLogger outside view.}
    \label{fig:elogger_outside}
    \end{subfigure}    
    \caption{Power collection. Drone is instrumented with eLogger and data is collected during flight.}
\label{fig:elogger}
\end{figure}

To study this impact path, we present the power breakdown of a commonly used off the shelf drone, the \solo~\cite{solo3DR}. We use an Eagle~Tree Systems
eLogger~V4~\cite{eLoggerV4} setup to measure power consumption (Figures ~\ref{fig:elogger_placement} and ~\ref{fig:elogger_outside}).
eLogger allows us to collect power consumption data at 50~Hz during flight. We command the drone to fly for \SI{50}{\second} and pull the data off of the power meter after the drone lands. Note that during this section, we isolate the impact of the compute power. Later on, we examine the power and performance impact together on mission metrics in Section~\ref{sec:impact_holistic}.


\begin{figure}
\begin{subfigure}[!t]{.4\linewidth}
\vspace{5pt}
\centering
    \includegraphics[trim=0 0 0 0, clip, width=\columnwidth]{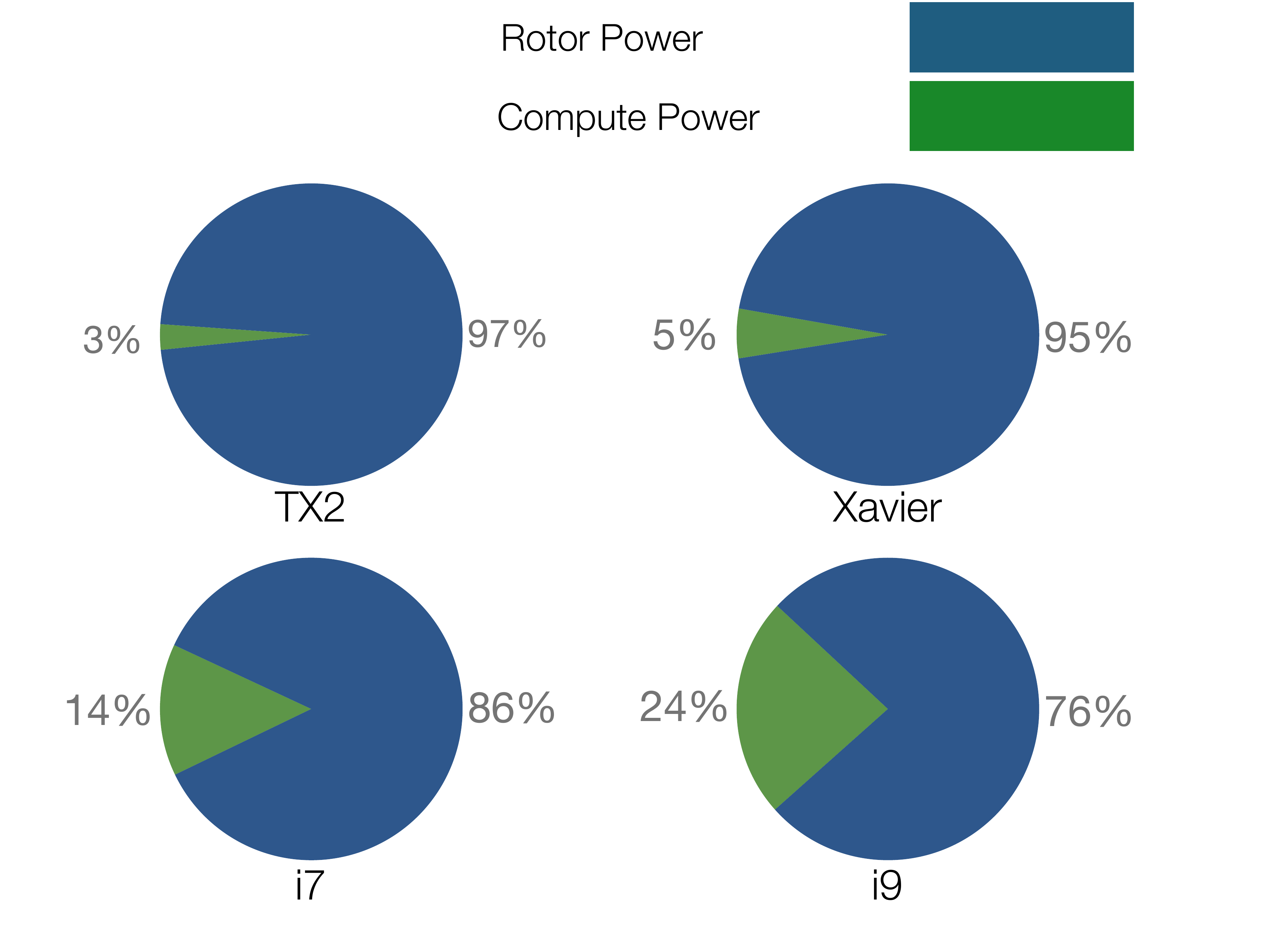}
     \vspace{18pt}
    \caption{Measured power breakdown.}
    \label{fig:power_break_down}
\end{subfigure}%
\begin{subfigure}[!t]{.52\linewidth}
   \vspace{2pt}
    \centering
    \includegraphics[trim=0 0 0 0, clip, width=\columnwidth]{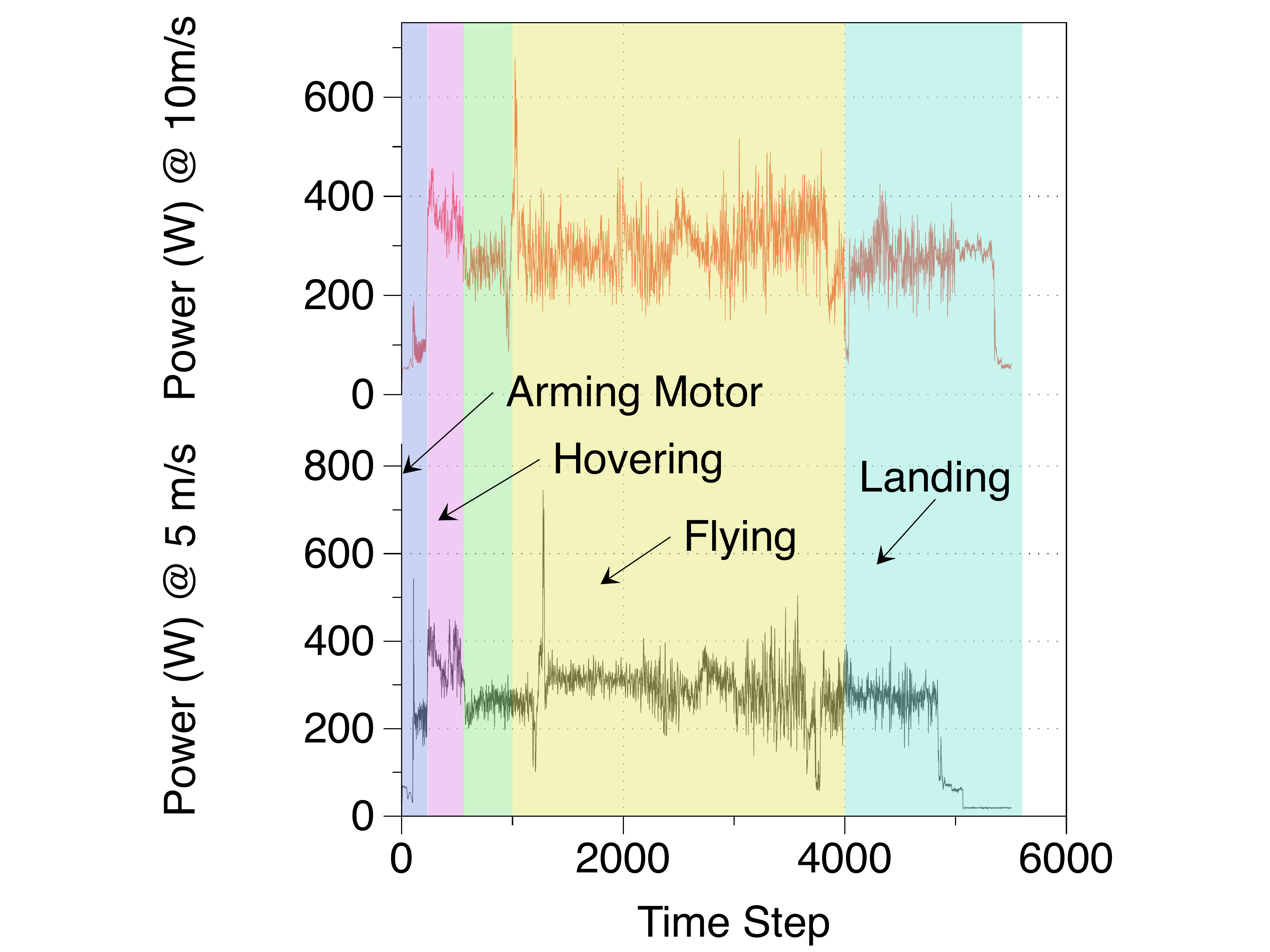}
    \caption{Power during flight at 5 and 10~m/s.}
    \label{fig:drone-power-time-series}
\end{subfigure}
\caption{Power profiling and breakdown. Compute generally makes up a small portion of the MAV's overall power pie. Velocity has a minor impact on power consumption for our MAV.}
\end{figure}

A drone with higher compute capability generally spends a higher budget of its total power on the compute subsystem in order to improve its performance. \Fig{fig:power_break_down} demonstrates this point by showing the power consumption of both the compute and the rotors for the
four platforms described earlier in Table~\ref{compute_platform_data}. Depending on the type of onboard compute system (i.e., TX2, Xavier, i7 or i9), the compute subsystem will directly consume 3\% to 24\% of the total system power. Hence, system designers must pay attention to such breakdowns since power-efficient designs will have a more significant impact on systems in which compute currently uses a more substantial proportion of total system power. 


\begin{figure}[t!]
    \centering
    \begin{subfigure}[t!]{.18\linewidth}
    \centering
    \includegraphics[width=\linewidth] {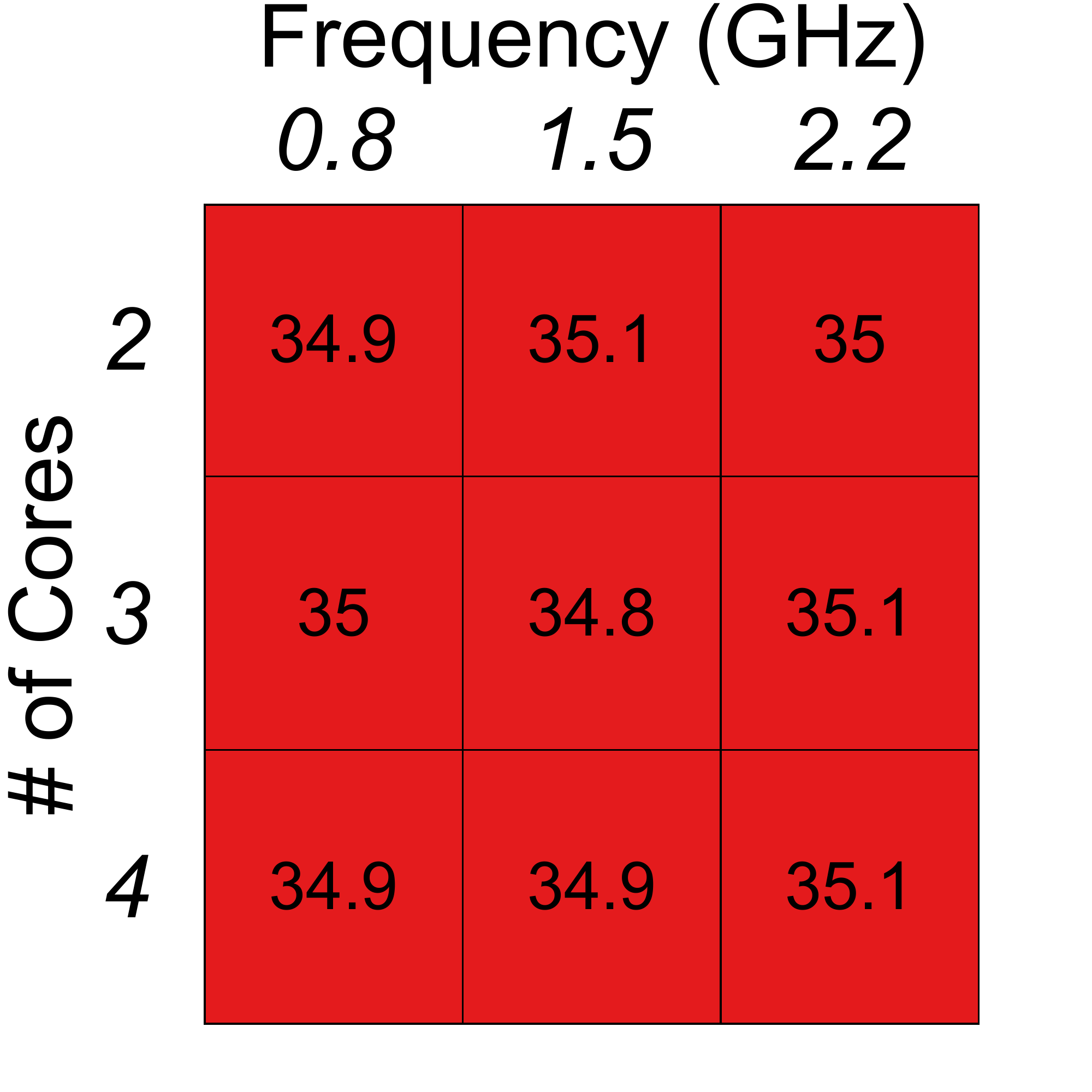}
    \label{fig:benchmarks:OPA:scanning:energy}
    \vspace*{-12pt}
    \caption{\scriptsize Scanning.}
    \end{subfigure}
    \begin{subfigure}[t!]{.18\linewidth}
    \centering
    \includegraphics[width=\columnwidth] {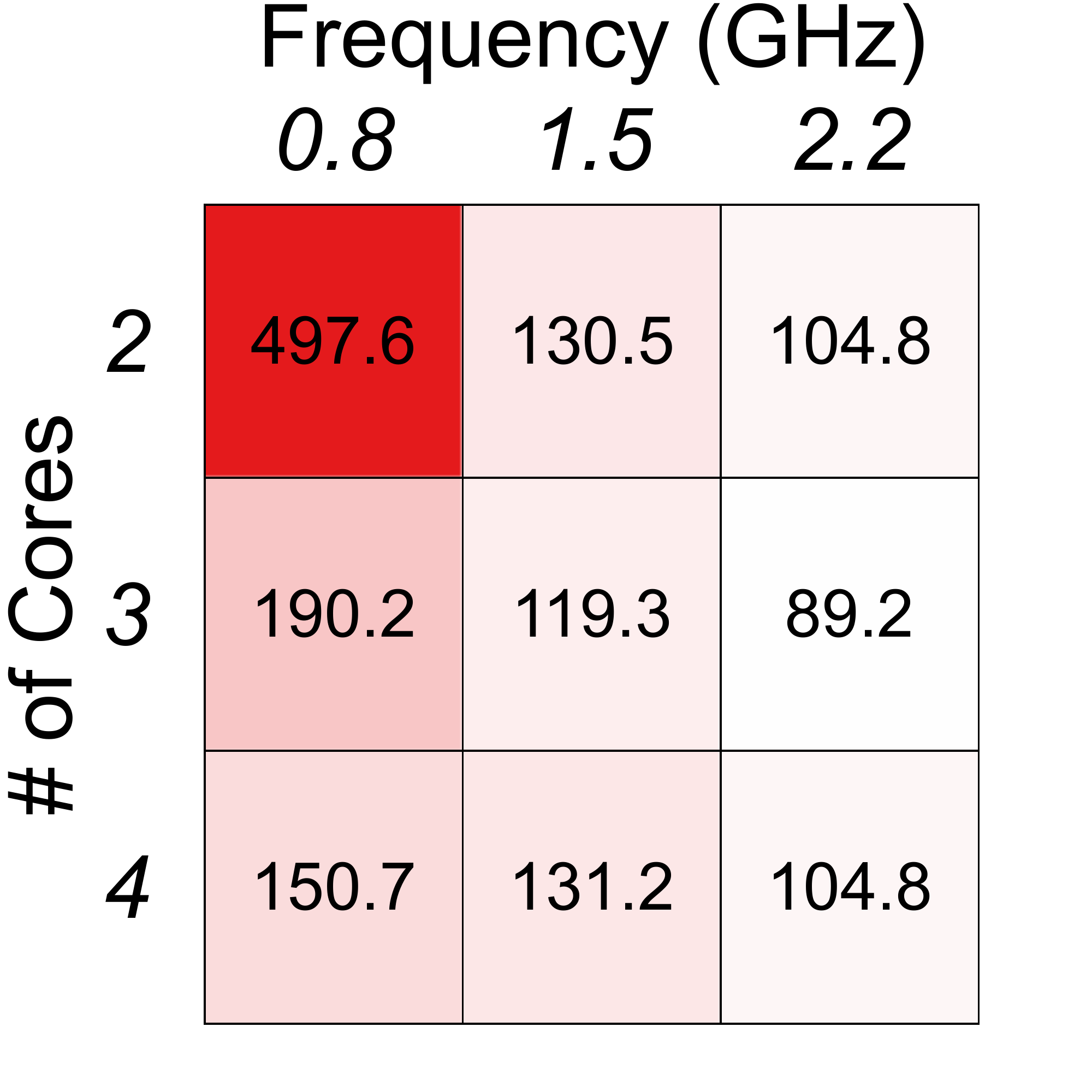}
     \label{fig:benchmarks:OPA:pd:energy}
     \vspace{-12pt}
       \caption{\scriptsize Package Delivery.}
    \end{subfigure}%
    \begin{subfigure}[t!]{.18\linewidth}
    \includegraphics[width=\linewidth] {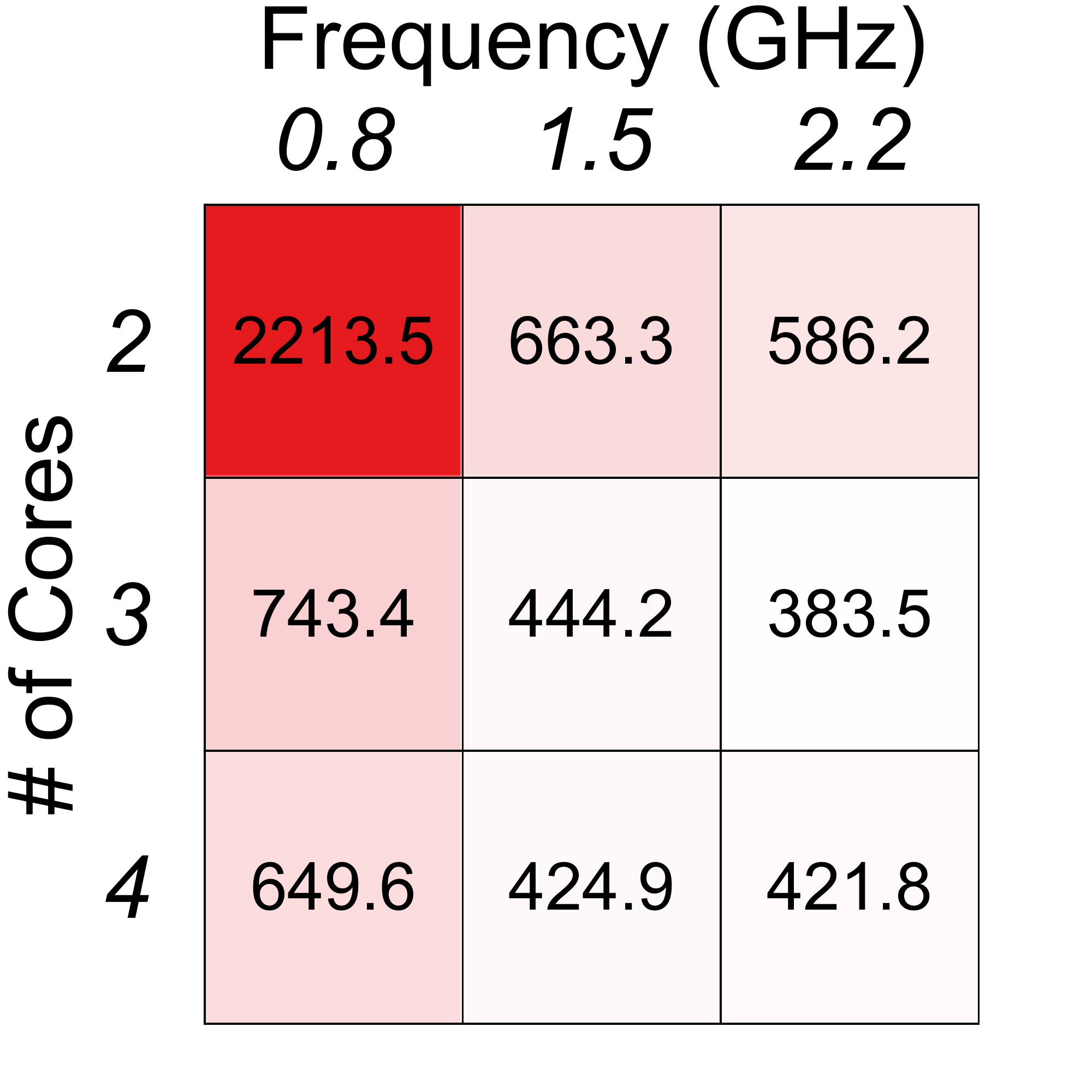}
     \label{fig:benchmarks:OPA:mapping:energy}
       \vspace{-12pt}
        \caption{\scriptsize 3D Mapping.}
    \end{subfigure}%
    \begin{subfigure}[t!]{.18\linewidth}
    \centering
    \includegraphics[width=\linewidth] {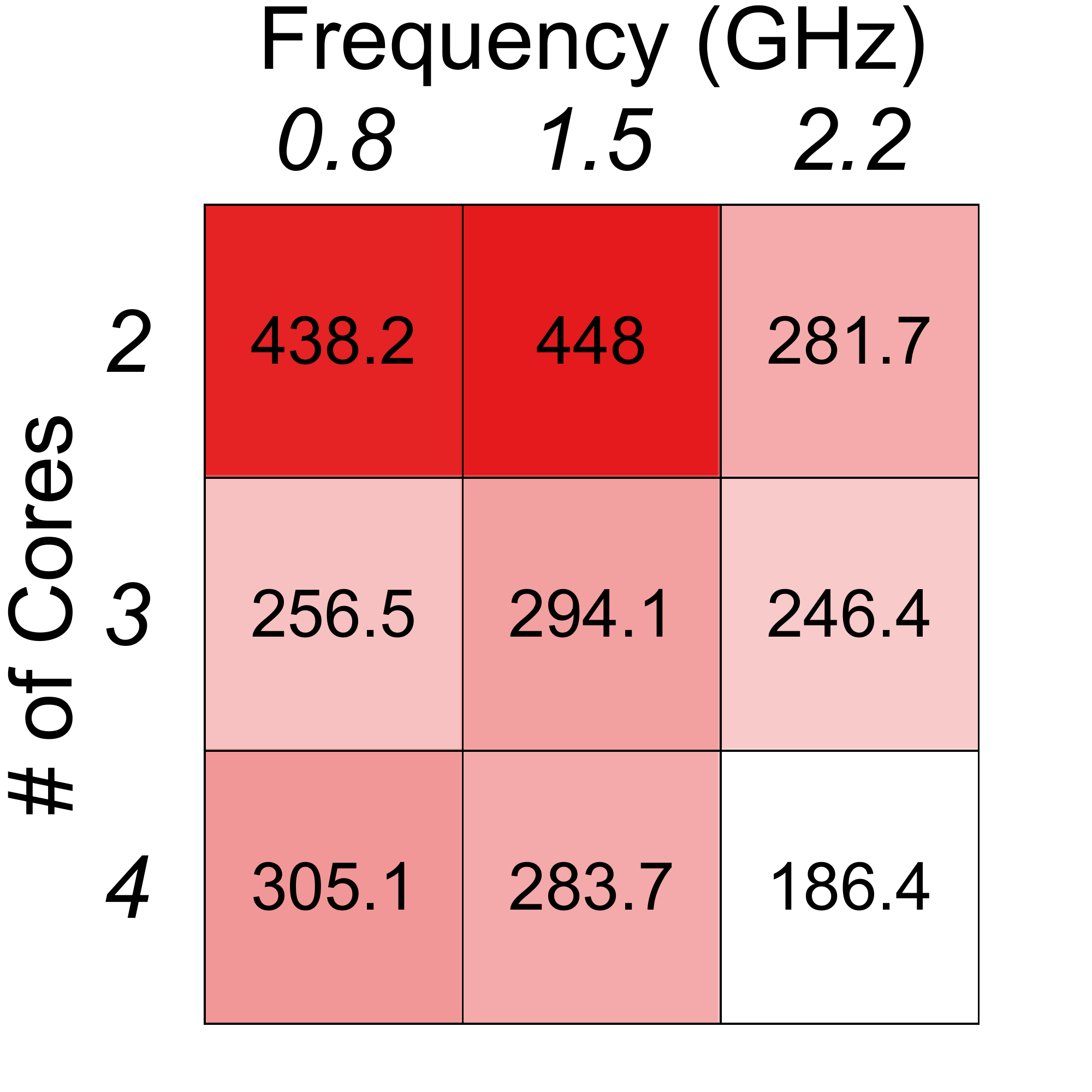}
     \label{fig:benchmarks:OPA:sar:energy}
      \vspace{-12pt}
      \caption{\scriptsize Search \& Rescue.}
    \end{subfigure}%
    \begin{subfigure}[t!]{.18\columnwidth}
    \centering
    \includegraphics[width=\columnwidth] {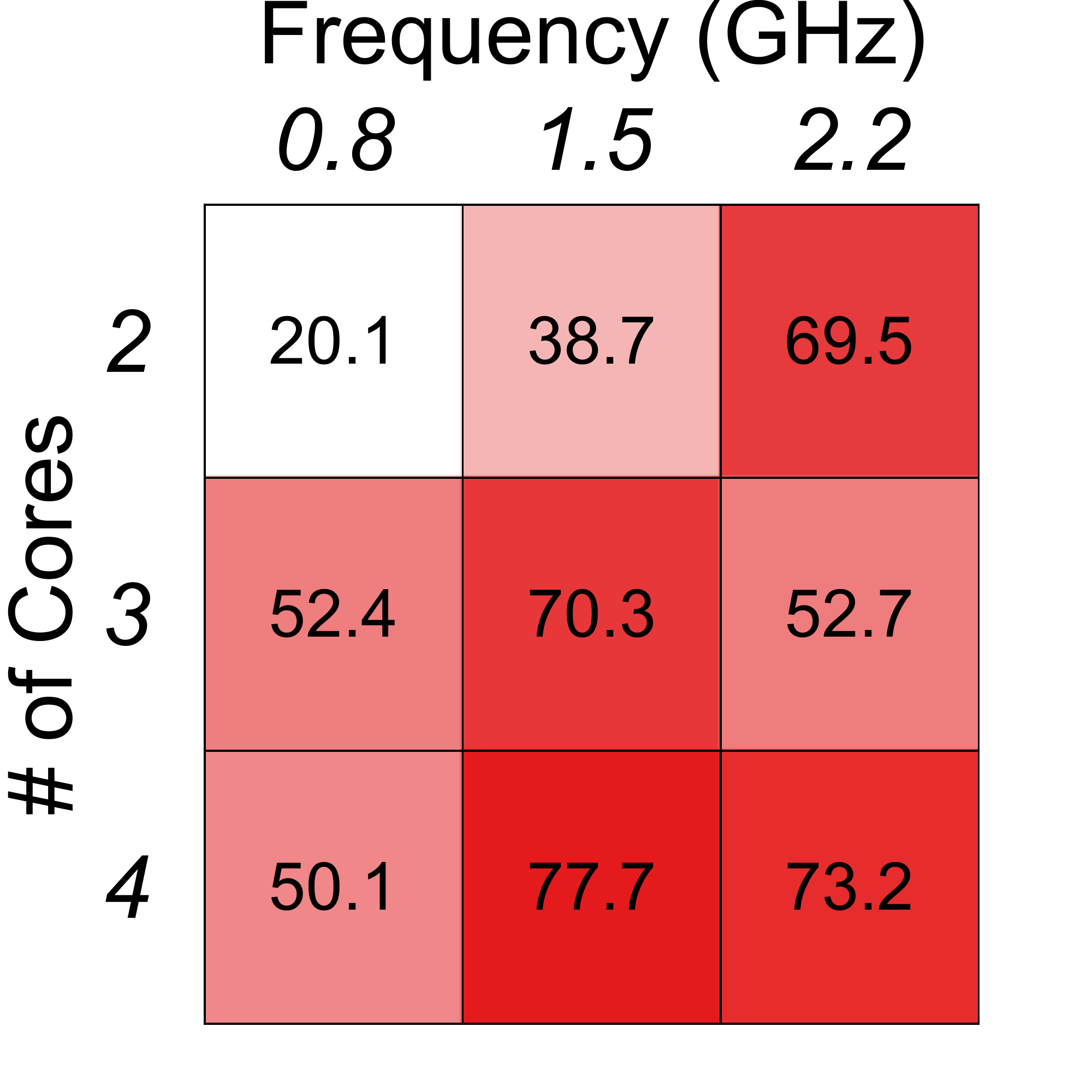}
     \label{fig:benchmarks:OPA:ap:energy}
     \vspace{-12pt}
     \caption{\scriptsize Aer Photography.}
    \end{subfigure}
    \caption{Core/frequency sensitivity analysis of mission energy for various benchmarks.}
    \label{fig:benchmarks_mission_energy}
\end{figure}

\subsection{Compute Performance Impact on Mission Energy}
\label{sec:comp_perf_impact_on_mission_energy}

Compute impacts mission energy through cyber quantities such as sensing-to-actuation latency, throughput, and etc, in other words, through the performance cluster (Figure~\ref{fig:CIG_performance_paths}, blue-color/coarse-grain-dashed paths). All such impacts start with cyber quantities
and then go through velocity, a physical quantity, to ultimately influence mission energy. 

Compute performance impacts a MAV's velocity (\Sec{sec:comp_perf_impact_on_mission_time}), which then impacts power consumption and ultimately impacts the MAVs total energy consumption.  To measure this impact, we used our eLogger~V4 setup.
As \Fig{fig:drone-power-time-series} shows, power variation as a result of velocity, be it 5~m/s (top) or 10~m/s (bottom), is rather minor for our MAV. This is because the majority of the rotor's power is spent keeping the drone airborne ($T_y$ from Figure~\ref{fig:quad_FBD_2D}), and a 
relatively small amount is used for moving forward during the flight ($T_x$ from Figure~\ref{fig:quad_FBD_2D}) for our allowed velocities.

In addition to the impact on energy through power, velocity can significantly reduce the total mission energy by reducing the mission time.
This is because as was shown in the previous section, rotors consume a bigger portion of the power consumption pie. Hence, by reducing the mission time, rotors spend less time in the air and hence consume less power and energy. 

Using the models provided in Section~\ref{sec:energy}, we profiled the mission time and the energy associated with the micro and end-to-end benchmarks discussed in the previous section, a SLAM microbenchmark and the MAVBench end-to-end benchmark suite. For the SLAM microbenchmark, as
shown in the previous section in Figure~\ref{fig:slam-velocity-energy}, a higher compute capability increases velocity (bottom graph) and lowers the total system energy (top graph). We see that by increasing processing speed from 1 FPS to 8 FPS, i.e., a 8X speed up, we are able to reduce the drone's energy consumption from \SI{76.9}{\kilo\joule} to \SI{21.1}{\kilo\joule}, i.e., close to 4X reduction.
For our end-to-end benchmarks, by conducting a core-frequency sensitivity analysis,
we show that more compute reduces the mission energy by as much as 5.8X (In 3D Mapping energy consumption is reduced from \SI{2213}{\kilo \joule} to \SI{283}{\kilo\joule}). These results are shown as a heatmap (Figure~\ref{fig:benchmarks_mission_energy}). Note that energy values closely follows the mission time trend discussed in Section~\ref{sec:comp_mission_time}, since a faster mission time lowers mission energy. 

\subsection{Compute Mass Impact on Mission Energy}
\label{sec:comp_mass_impact_on_mission_energy}
Compute mass both directly and indirectly impacts rotor power consumption (mass cluster shown in Figure~\ref{fig:CIG_mass_paths} with green-color/double-sided paths). This is because mass itself, as shown in Section~\ref{sec:comp_mass_impact_on_mission_time}, impacts acceleration and velocity, and all three impact power. This can be seen in the mechanical power model provided in Section~\ref{sec:energy}. Equation~\ref{eq:DJI_power} shows this model specifically for our DJI quad. 
 \begin{equation}
    \begin{aligned}
    P = \begin{bmatrix}
            -1.526 \\
            3.934 \\
            0.968
        \end{bmatrix}^{T}
        \begin{bmatrix}
            \norm{\vec{v}_{xy}} \\
            \norm{\vec{a}_{xy}} \\
            \norm{\vec{v}_{xy}}\norm{\vec{a}_{xy}}
        \end{bmatrix}
        +
        \begin{bmatrix}
            18.125 \\
            96.613 \\
            -1.085
        \end{bmatrix}^{T}
        \begin{bmatrix}
            \norm{\vec{v}_{z}} \\
            \norm{\vec{a}_{z}} \\
            \norm{\vec{v}_{z}}\norm{\vec{a}_{z}}
        \end{bmatrix}
        +
        \begin{bmatrix}
            .22\\
            1.332\\
            433.9
        \end{bmatrix}^{T}
        \begin{bmatrix}
            m \\
            \vec{v}_{xy} \cdot \vec{w}_{xy} \\
            1
        \end{bmatrix}
    \end{aligned}
    \label{eq:DJI_power}
    \end{equation}
 
\begin{figure}[t!]
\centering
     \begin{subfigure}{0.4\linewidth}
   \includegraphics[width=.9\linewidth]{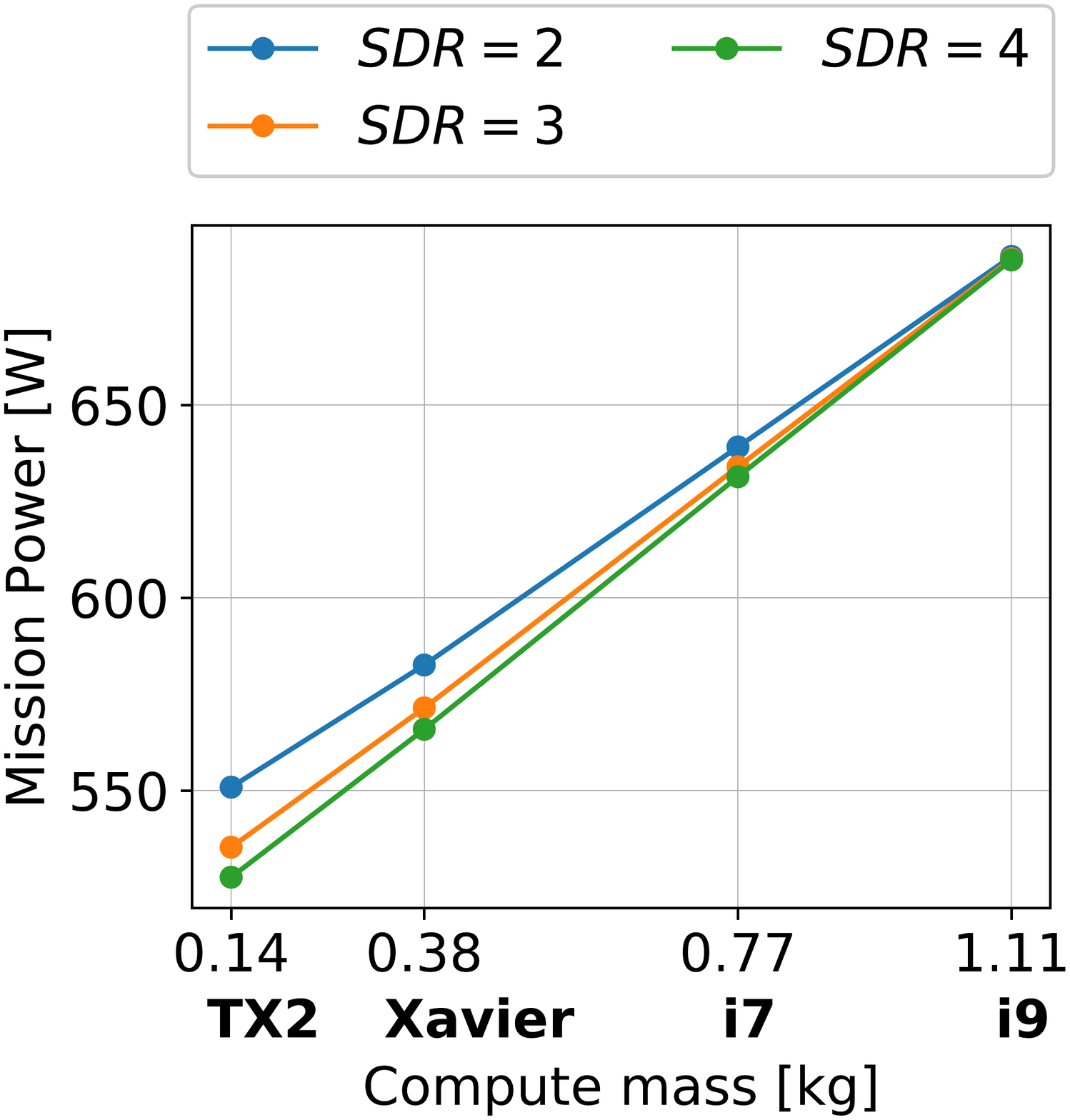}
   \vspace{-20pt}
   \caption{Impact on power.}
   \label{fig:mass_power}
    \end{subfigure} 
    \begin{subfigure}{0.4\linewidth}
   \includegraphics[width=.9\linewidth]{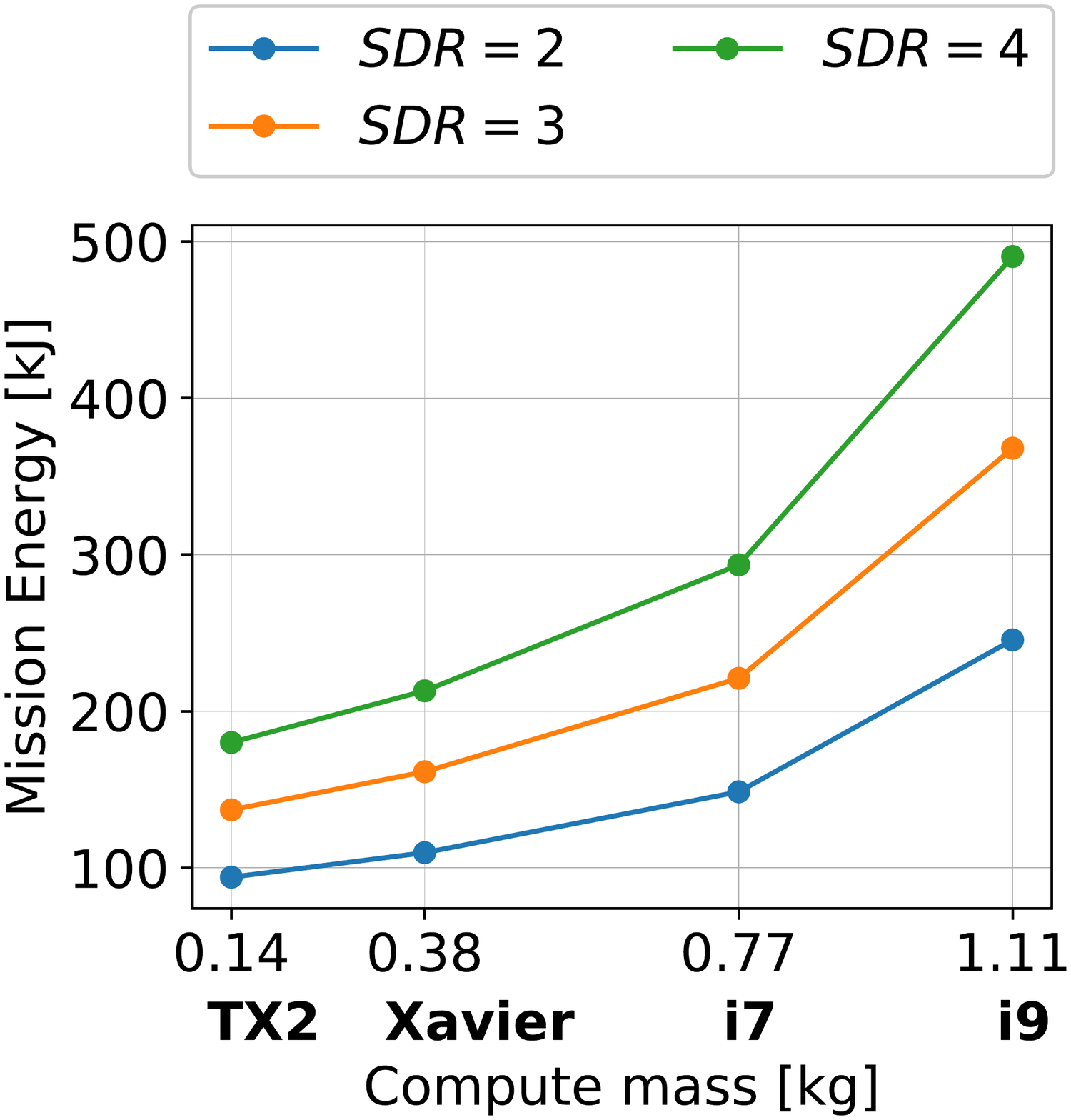}
    \vspace{-20pt}
    \caption{Impact on energy.}
    \label{fig:mass_energy}
    \end{subfigure}
   \caption{Impact of compute mass, a physical quantity, on power and energy.}
      \label{fig:comp_stop_distance}
\end{figure}

As mass increases, so does power. Figure \ref{fig:mass_power} shows this effect for the DJI quadrotor across different compute platforms. An increase in the compute mass from \SI{144}{\g}, associated with the lightest MAV with a TX2, to \SI{1109}{\g}, associated with the heaviest MAV with an i9, i.e., a 7.9X increase, causes the total MAV mass to exacerbate from \SI{2544}{\g} to \SI{3509}{\g}, i.e., a 1.4X increase. This then heightens the power consumption from \SI{527}{\watt} to \SI{687}{\watt}, i.e., 20\% increase in total power (for SDR of 4). Note that the slow down ratio has a minor impact on power as all the lines are very close to one another. This is because the velocity has a minor impact on power as discussed in Section~\ref{sec:comp_perf_impact_on_mission_energy}.  

Compute mass impacts energy consumption by impacting both power and mission time. An increase in compute mass exacerbates the energy consumption by increasing power as was detailed in the previous paragraph. In addition, an increase in mass exacerbates energy consumption by increasing the mission time. This is due the reduction in velocity and acceleration that was discussed in Section~\ref{sec:comp_mass_impact_on_mission_time}. \Fig{fig:mass_energy} shows an overall system energy consumption increase from \SI{180}{\kilo\joule} to \SI{490}{\kilo\joule}, i.e. a 2.7X increase, between the lightest (TX2) and the heaviest (i9) platforms (for SDR of 4). Note that like our observation for mission time, this impact grows with the environment's difficulty level, i.e., the environment's congestion. Therefore, as the environment becomes harder to navigate, the MAV requires more power-efficient designs.

\section{Role of Compute, a Holistic Outlook}
\label{sec:impact_holistic}
\blueDebug{
    section flow:
    - hollistic look at instead of in isolation (why)
    - explain the experiment very briefly
    - describe the data
    - motivate exhaustive DSE
    - explain data generation
        - explain how we arrive at this data 
        - explain bounderis
    - explain the results
}

The impact of compute clusters needs to be studied not only in isolation, as we have done so far to gain an in-depth understanding, but also simultaneously. The latter sheds light on the aggregate impact of all clusters together. Such a holistic outlook is especially valuable when the clusters have opposite impacts, i.e., one with a positive and the other with a negative impact on a mission metric. 

To this end, in this section, we study the simultaneous effect of all three clusters, performance, mass, and power, on mission time and energy. 
Our studies, similar to Section~\ref{sec:comp_mass_impact_on_mission_time}, deploy a combination of benchmarking and analysis for a package delivery application.
First, we uncover various impacts by performing holistic analysis using the four platforms in Table~\ref{compute_platform_data}. Subsequently, we expand the design space to its boundaries, 
allowing designers to determine which impact clusters, and hence which design techniques,
are the most beneficial.



\subsection{Compute Impact on Four Platforms}
\label{sec:eval_4_platforms}
Compute mass, power and performance, holistically, impact acceleration, velocity, mission time and energy. We use a combination of simulation and analysis to examine the various impacts collectively on a DJI Matrice 100 drone equipped with the four platforms mentioned previously in Table~\ref{compute_platform_data}. 
 
\begin{figure}[t!]
\centering
    \begin{subfigure}[b]{.52\textwidth}
   
    \centering 
   
   \includegraphics[height=1.5in]{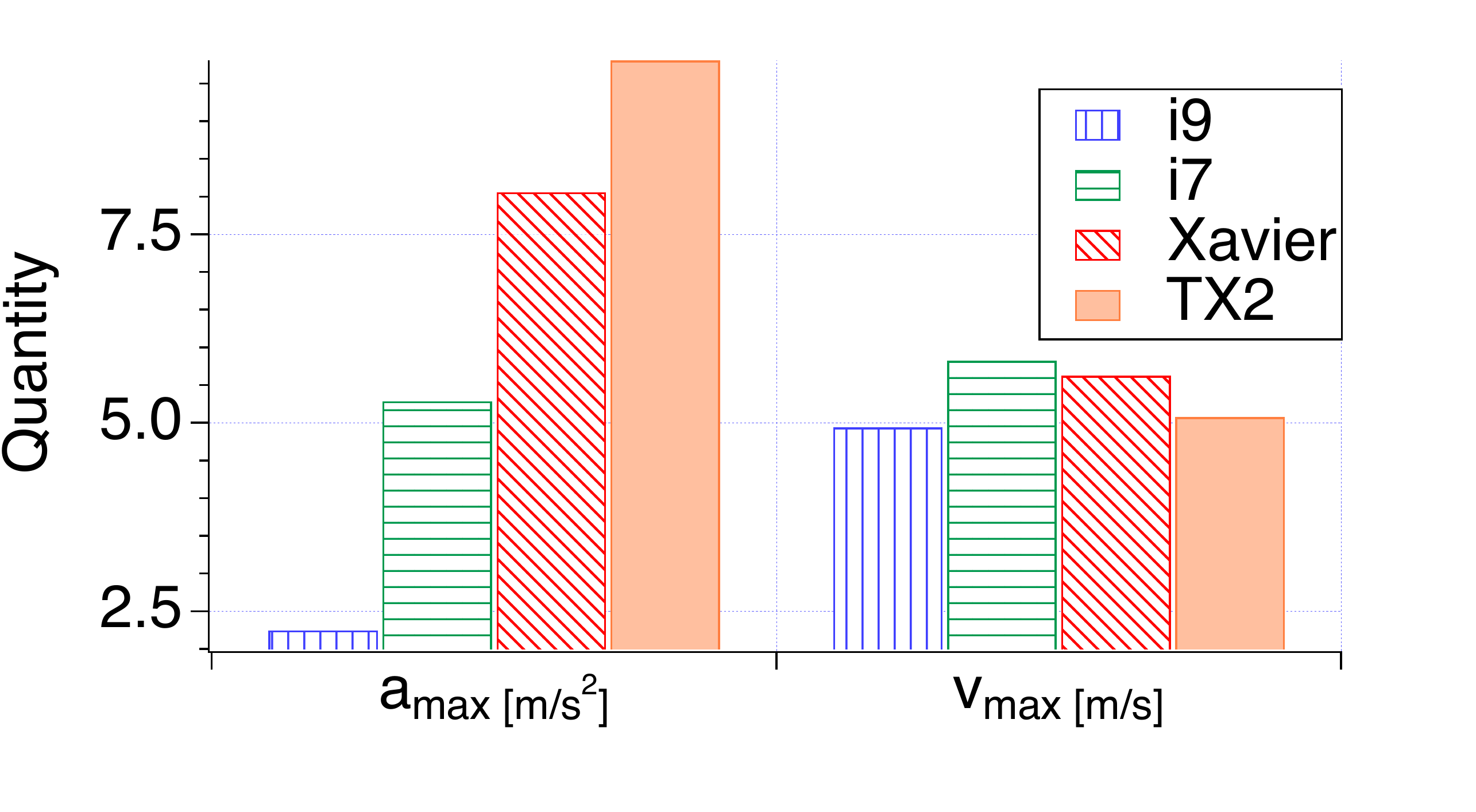}
   \caption{Velocity and acceleration.}
   \label{fig:vel_acc_full_impact}
   \end{subfigure} 
   \begin{subfigure}[b]{0.47\textwidth}
    \centering
    \includegraphics[height=1.5in]{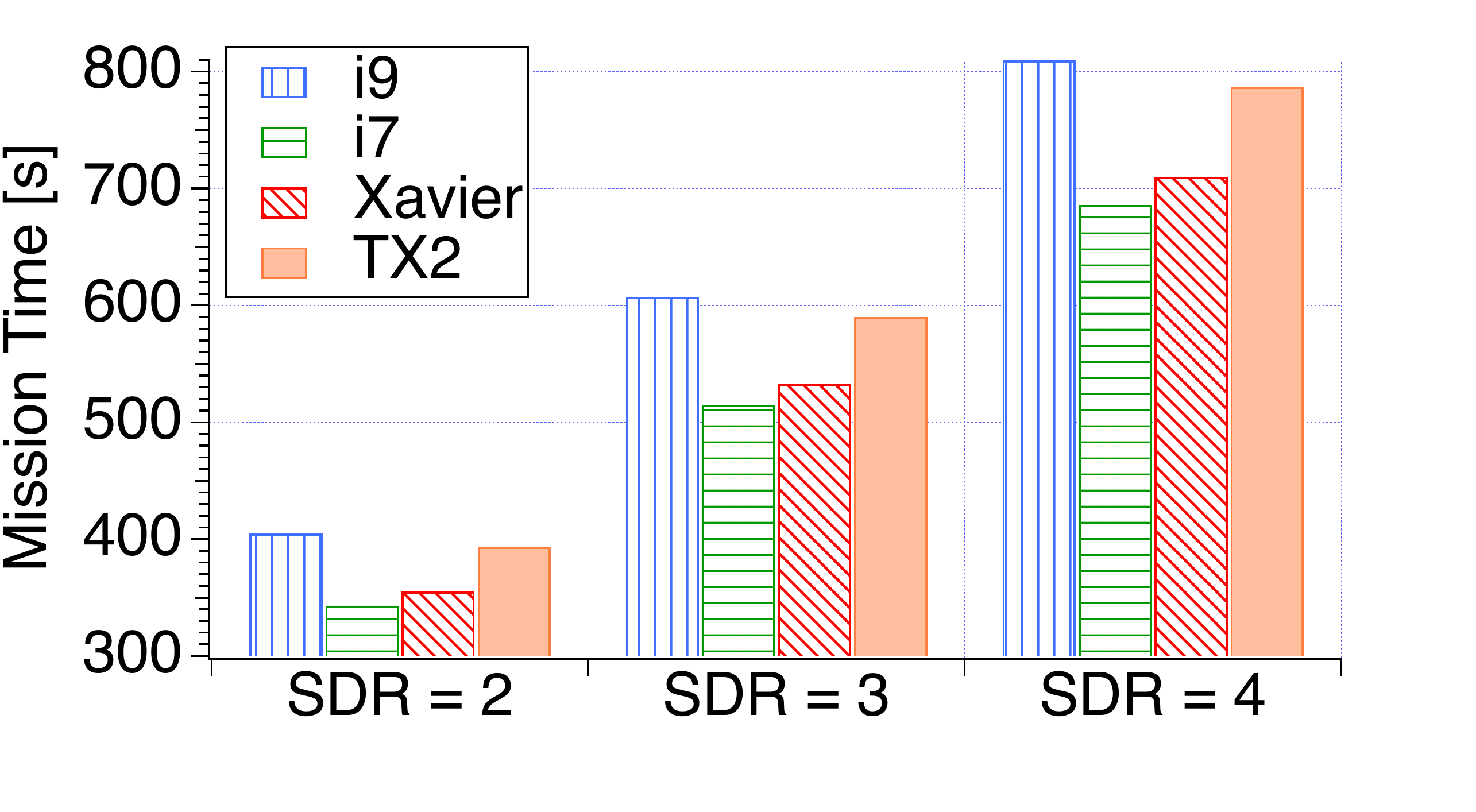}
    \caption{Mission time.}
    \label{fig:comp_mt_full_impact}
    \end{subfigure}
    
    \vskip\baselineskip
   \begin{subfigure}[b]{.49\textwidth}
    \centering
   \includegraphics[height=1.5in]{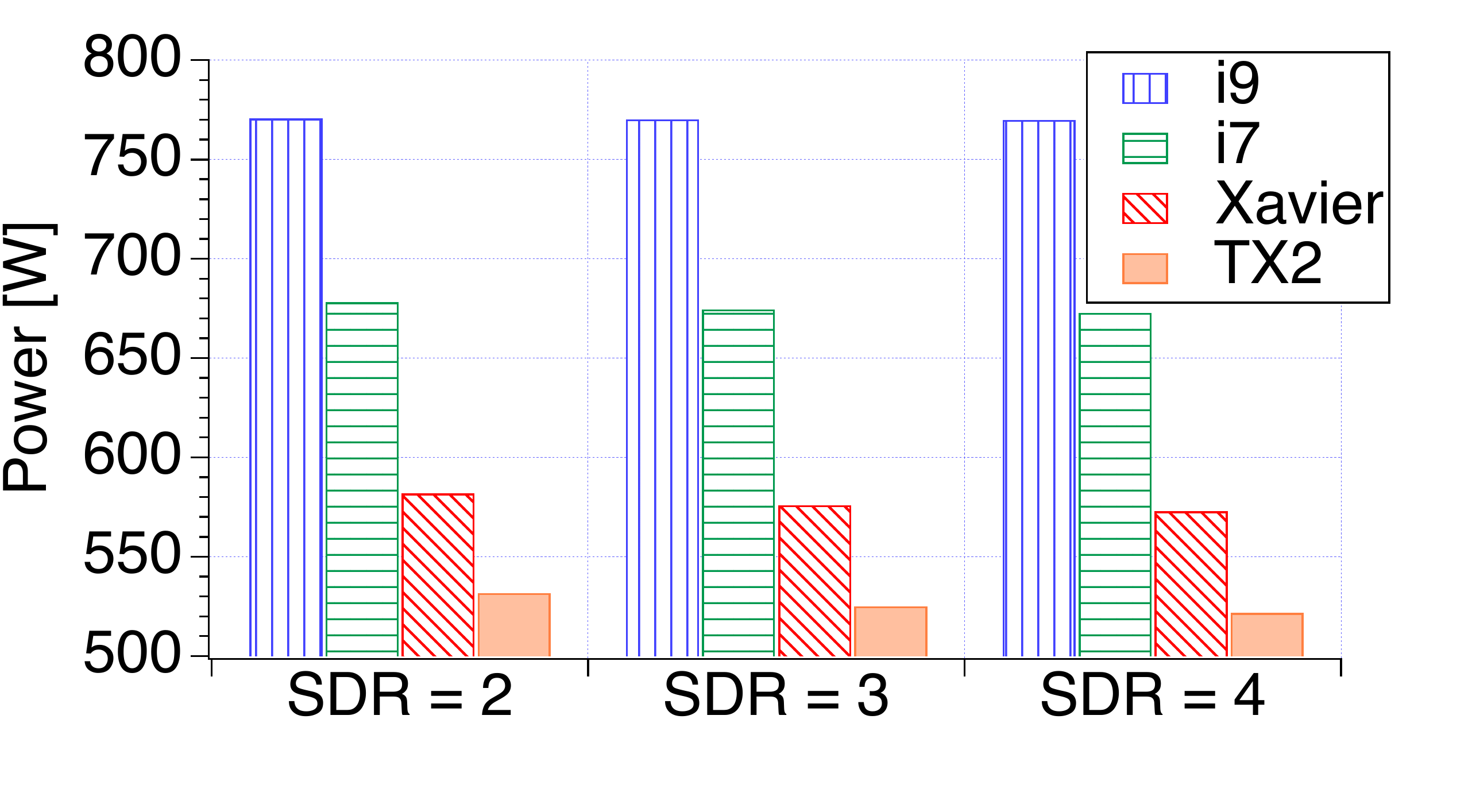}
    \caption{Mission power.}
    \label{fig:comp_power_energy_full_impact}
    \end{subfigure}
    \begin{subfigure}[b]{0.49\textwidth}
    \centering
    \includegraphics[trim=0 0 0 0, clip, height=1.5in]{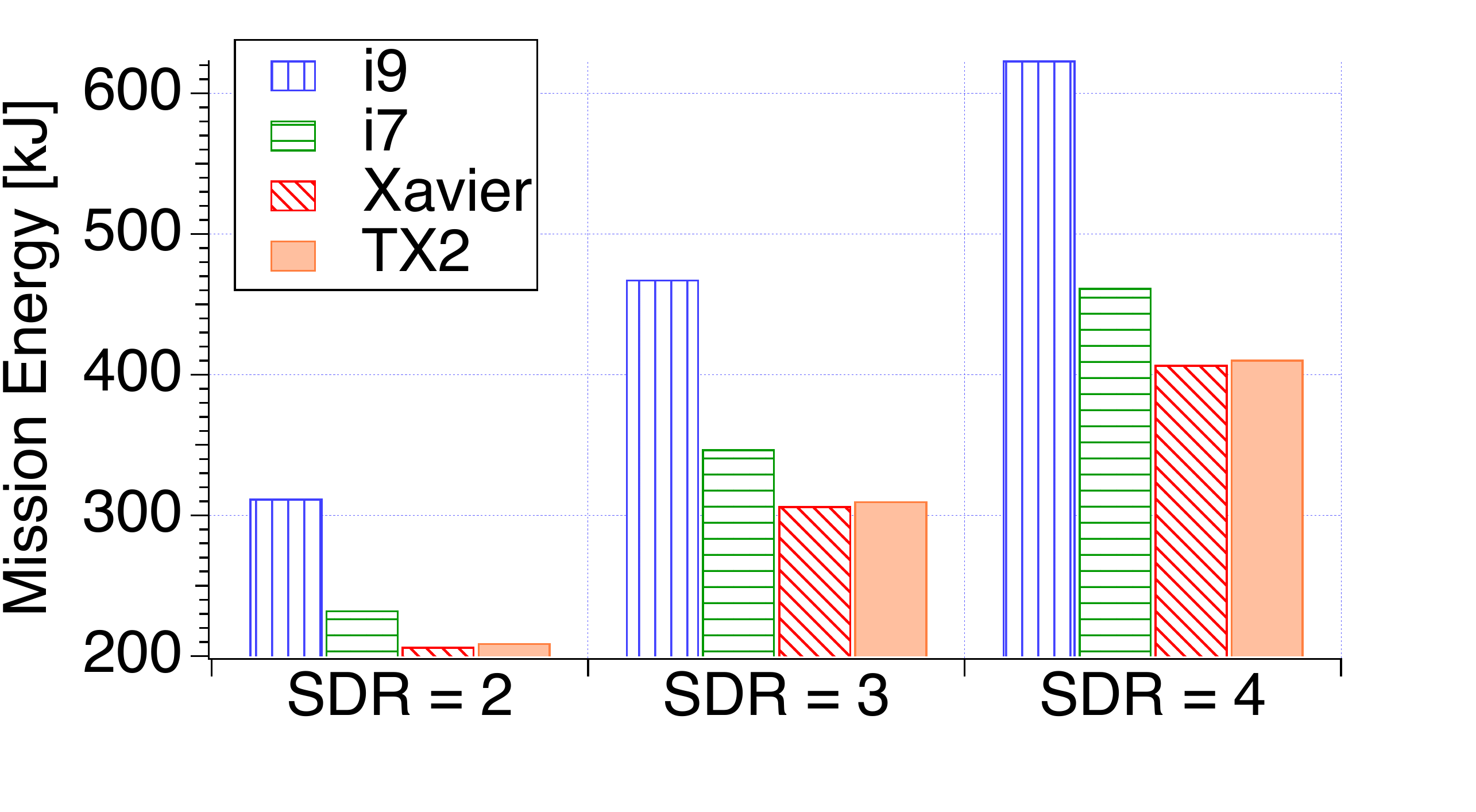}
    \caption{Mission energy.}
    \label{fig:comp_me_full_impact}
    \end{subfigure}
    \caption{Holistic impact of compute on mission metrics. The data enables cyber-physical co-design  where the designers consider the impact of compute's cyber quantities on the robot's physical quantities and further their ultimate impact on robot's end-to-end behavior and mission success. When all impacts are considered together, there is no obvious best design choice. For instance, the i7 has the best (shortest) mission time (highlighted in green) but the Xavier has the lowest mission energy (highlighted in red).}
      \label{fig:full impact}
\end{figure}

We find that increasing onboard compute capability impacts acceleration negatively. Figure ~\ref{fig:vel_acc_full_impact} shows this impact where we see an acceleration of \SI{9.8}{\meter/\second\squared} vs. \SI{2.3}{\meter/\second\squared}, i.e., a 4.4X decrease, when our MAV swaps the lightest (TX2) with the heaviest (i9) platform.
In contrast, an increase in compute has both a negative and positive impact on velocity. 
Through the positive path, i.e., the performance cluster as we showed in Section~\ref{sec:comp_perf_impact_on_mission_time}, more compute reduces response time, and hence increases velocity. Consequently, robot designers should deploy more compute. On the other hand, through the negative path, i.e., the mass cluster as was described in Section ~\ref{sec:comp_mass_impact_on_mission_time}, more compute increases mass, and hence reduces acceleration and velocity. Consequently, robot designers should deploy less compute. A holistic study however enables us to weigh these competing impacts simultaneously. 

For our four platforms, when paths are examined in combination, we see that not the least nor the most compute capable system, but a middle ground compute platform, i7, has the highest maximum velocity, i.e., \SI{5.8}{\meter/\second}. Our most compute capable platform, i9, has the least velocity, i.e., \SI{4.9}{\meter/\second}. Compute causes a 1.2X difference between our slowest and fastest MAVs. A DJI MAV with an i7 is faster than a DJI MAV with either the TX2 or Xavier due to the reduction in response time caused by more compute capability. However, increasing the compute capability to i9 does not improve the velocity due to the negative impact of the added mass, which outweighs the benefits of the response time reduction for this compute platform.




A similar trend to velocity, where the least nor the most compute capable system is ideal, is observed with mission time. Figure ~\ref{fig:comp_mt_full_impact} shows this impact where our fastest MAV with i7 has the shortest mission time, i.e., \SI{686}{\second}, comparing to our slowest MAV with i9 which has the longest mission time, i.e., \SI{809}{\second}. Between the two, i7 shows 15\% improvement. Slow down factor (SDR) expands the gap between the best (\SI{62}{\s}) and the worst (\SI{124}{\s}) mission time difference. This again reminds us that for higher congested spaces, more efficient design has a bigger impact.


More compute negatively impacts mission power consumption (Figure~\ref{fig:comp_power_energy_full_impact}). This is due to the high impact of mass on power as discussed in Section~\ref{sec:comp_perf_impact_on_mission_energy}. As such, our most compute capable and hence heaviest platform, i9, consumes the most power, i.e., \SI{770}{\watt}, comparing to the least compute capable and hence lightest platform, TX2, which consumes \SI{521}{\watt} power.
Note that the slow down ratio has a minor impact on power since the velocity (which gets impacted by SDR) has a minor effect on power, as discussed in Section~\ref{sec:comp_perf_impact_on_mission_energy}.


In the case of mission energy, there are two competing clusters with opposite impacts. Through the positive path, i.e., performance cluster as we showed in Section ~\ref{sec:comp_perf_impact_on_mission_energy}, more compute results in less energy consumption due to the reduction of mission time. Hence, more compute is desirable. On the other hand,
through the negative path, i.e., the mass cluster as we showed in Section ~\ref{sec:comp_mass_impact_on_mission_energy}, more compute results in more energy consumption. Hence, robot designers should deploy less compute. However, when combined, we see that not the most nor the least compute capable platforms, but another middle ground platform, Xavier performs the best. Xavier burns only \SI{407}{\kilo \joule}, comparing to the least energy efficient design, i.e., i9, which burns \SI{623}{\kilo \joule} (Figure~\ref{fig:comp_me_full_impact}).
Higher slow down factor (SDR) expands the gap between the best and worst design which
reminds us that for higher congested spaces, more efficient design has a bigger impact.

In summary, the MAV with the best mission time is not necessarily the most energy efficient design or vice versa. For instance, in our example, the MAV with the best mission time which carries an i7 is not the most energy efficient design which carries an Xavier. This is because, with i7, the negative impact of higher power consumption outweighs the benefit of lower mission time. The choice between the Xavier and the i7 depends on whether mission time or energy has more significance for the task at hand. 

\subsection{Expanding the Design Space}
\label{sec:eval_all_inclusive_DSE}
Understanding the compute subsystem's design space for a MAV, from both the cyber and physical perspective, can help guide the design process.
This section goes beyond the four platforms from Table~\ref{compute_platform_data} and provides an example of, and the insights gathered from, a full design space investigation. Concretely, we examine how the entire design space changes as a function of three compute subsystem quantities and clusters, i.e., compute mass, compute power and performance. 


Our evaluation employs the same analytical and simulation-based models discussed in the previous subsection.
We assume that the quantities can be modified independently. 
For example, we assume that we can increase or decrease mass without effecting response-time or power. MAV's constraints determine the bounds for each quantity.  The maximum payload that the drone can
carry and still lift off the ground determines the compute mass upper bound. The total available energy on board determines the response time and power upper bounds. If the compute
power causes the battery to drain fast enough, or the response time to be long enough (and hence the MAV's speed low enough) that the MAV becomes unable to complete its mission before it runs out of
battery, we discard this design point. For power, we also take into account the upper current limit the battery can provide.\footnote{Although we assume no relationship between the three quantities, in reality since compute impacts
all three simultaneously, there is an indirect relationship between them all. This narrows the design space beyond the constraints we have already put into place. 
We are aware that sharpening such constraints and replacing our high-level models with more accurate ones will improve the insight details, nevertheless, 
we believe that lower resolution models have an essential role in the initial stages of the iterative design and development process.
We welcome the community's assistance in both refining and replacing the models and constraints while considering the accuracy-performance tradeoffs of such changes.}

\begin{figure}[t!]
\centering
    \begin{subfigure}{.51\columnwidth}
   \includegraphics[width=\columnwidth]{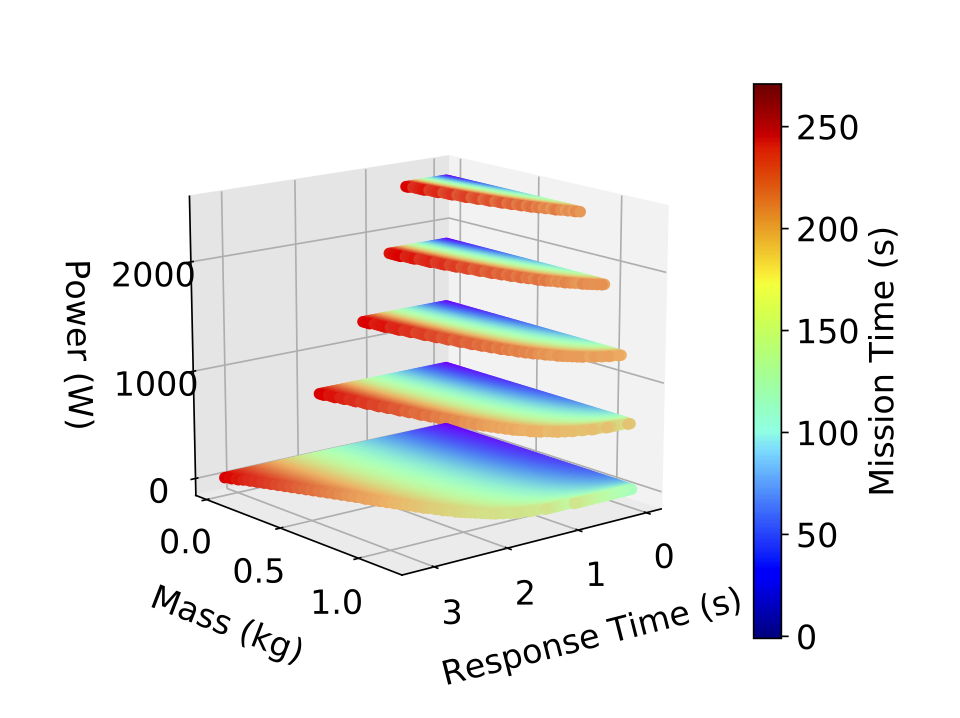}
   \caption{Mission time design space.}
   \label{fig:mt_ds}
    \end{subfigure} 
    \begin{subfigure}{.48\columnwidth}
   \includegraphics[width=\columnwidth]{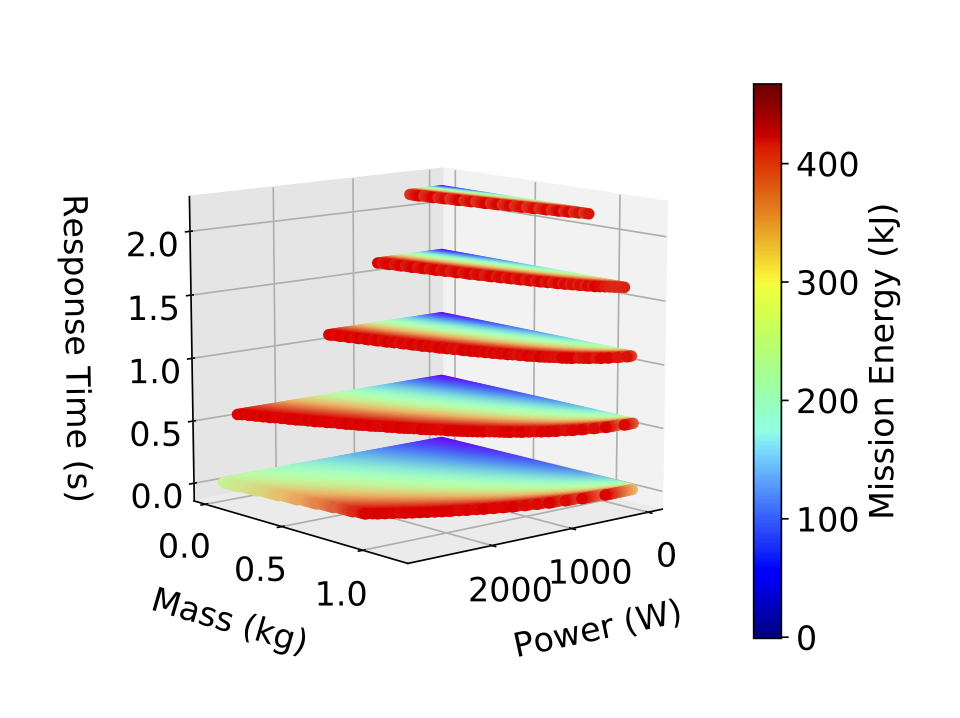}
    \caption{Mission energy design space.}
    \label{fig:me_ds}
    \end{subfigure}
    \caption{Mission metrics design space with respect to different clusters. Mission metrics are shown as heatmaps where hotter is less optimal. The design space is sliced and only a selected set of slices are shown for the ease of visualization.}
    \label{fig:DS}
\end{figure}

Our results are shown in Figure~\ref{fig:DS}, where we illustrate the mission time (Figure~\ref{fig:mt_ds}) and energy (Figure~\ref{fig:me_ds}) for a DJI Matrice 100 drone. Each point within the space corresponds to a DJI with a compute subsystem of a different mass, response time, and power. 
Mission metrics are shown as heat maps on the fourth dimension. Hotter colors are higher and hence, less optimal.
To aid the visualization and further understand the internals of the design space, we slice the space with horizontal cuts, concretely five slices for each mission metric. 
Furthermore, for selected slices, we show the gradient plots (i.e., plots of the rate of change in the most optimal direction) to demonstrate various trends. We spend the rest of this section detailing some essential takeaways from this analysis.

%


First, as one quantity becomes less optimal, it shrinks the design space of the other two. This can be seen by looking at the reduction in the area of the slices for both mission metrics shown in Figure~\ref{fig:DS}. For example, for mission time (Figure~\ref{fig:mt_ds}), as the power consumption increases (moving up the $z$-axis), the design space slices see a reduction in area for mass and response time. For example, for a 4X increase in power quantity, i.e., from \SI{615}{\watt} to \SI{2415}{\watt}, we see a 5X reduction in the area associated with the mission time's design space. Similar trends are observed for mission energy. Note that such a reduction in the area leaves designers with less space (area) to explore and design within. 

\begin{figure}[t!]
\centering
    \begin{subfigure}{.48\columnwidth}
     \includegraphics[width=\columnwidth]{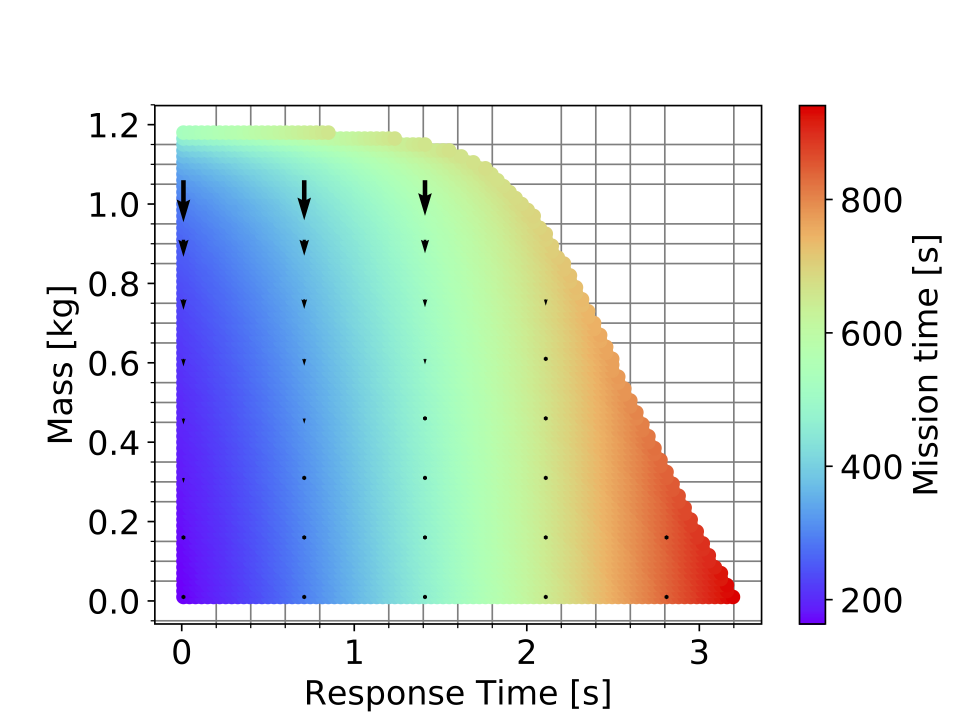}
     \caption{Mission time-mass gradient.}
     \label{fig:mt_gradient_mass}
    \end{subfigure}
   \begin{subfigure}{.48\columnwidth}
    \includegraphics[width=\columnwidth]{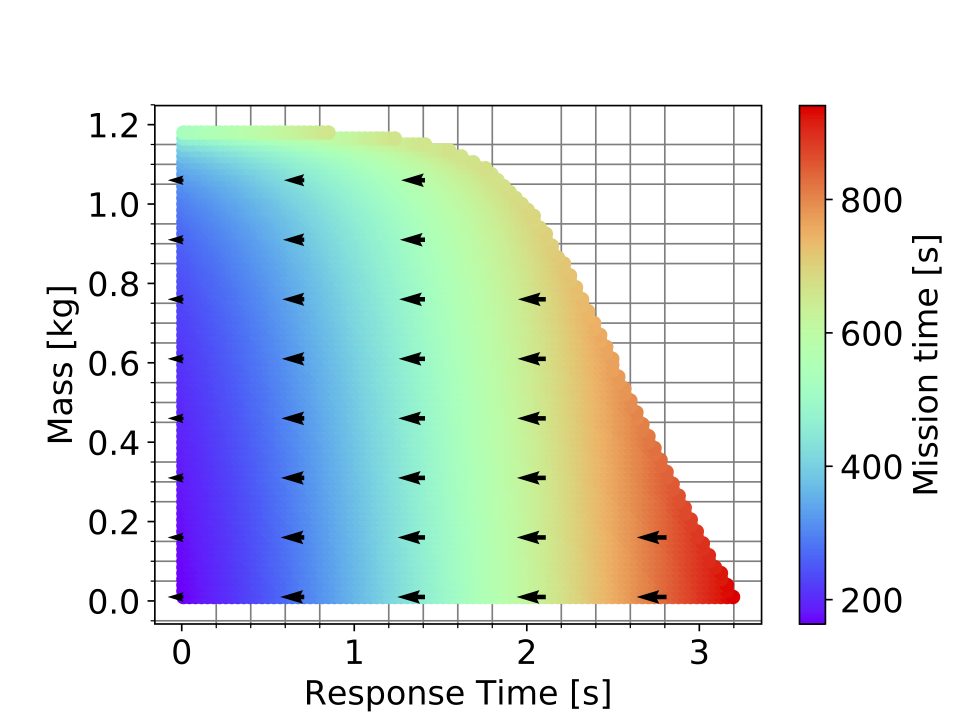}
    \caption{Mission time-response time gradient.}
    \label{fig:mt_gradient_response_time}
    \end{subfigure}
    
    \vspace{2pt}
    \begin{subfigure}{.48\columnwidth}
     \includegraphics[width=\columnwidth]{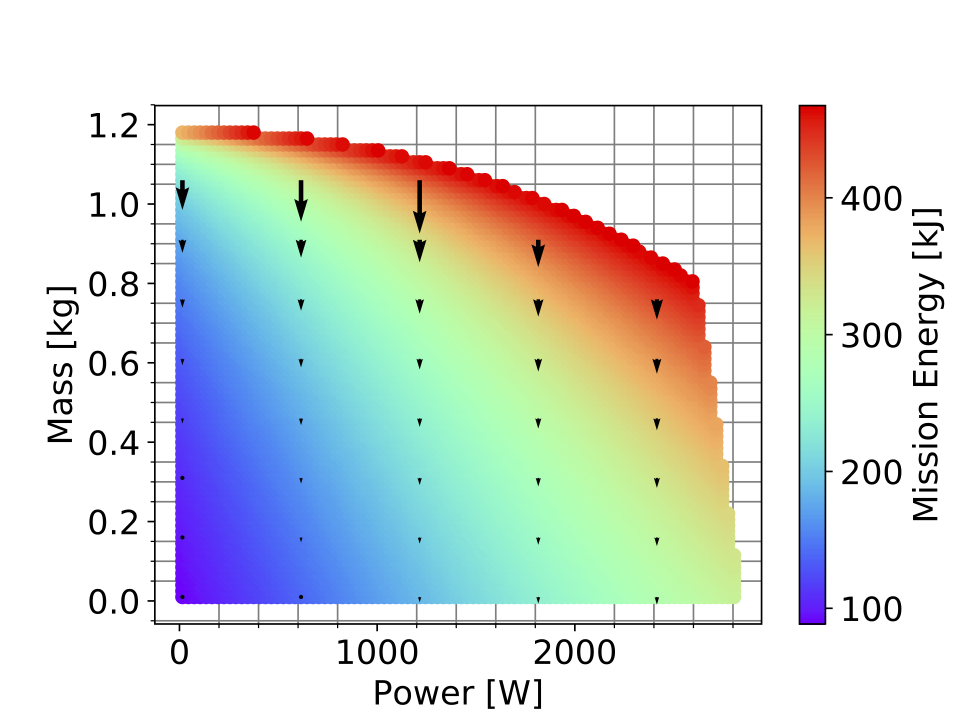}
     \caption{Mission energy-mass gradient.}
     \label{fig:me_gradient_mass}
    \end{subfigure}
   \begin{subfigure}{.48\columnwidth}
    \includegraphics[width=\columnwidth]{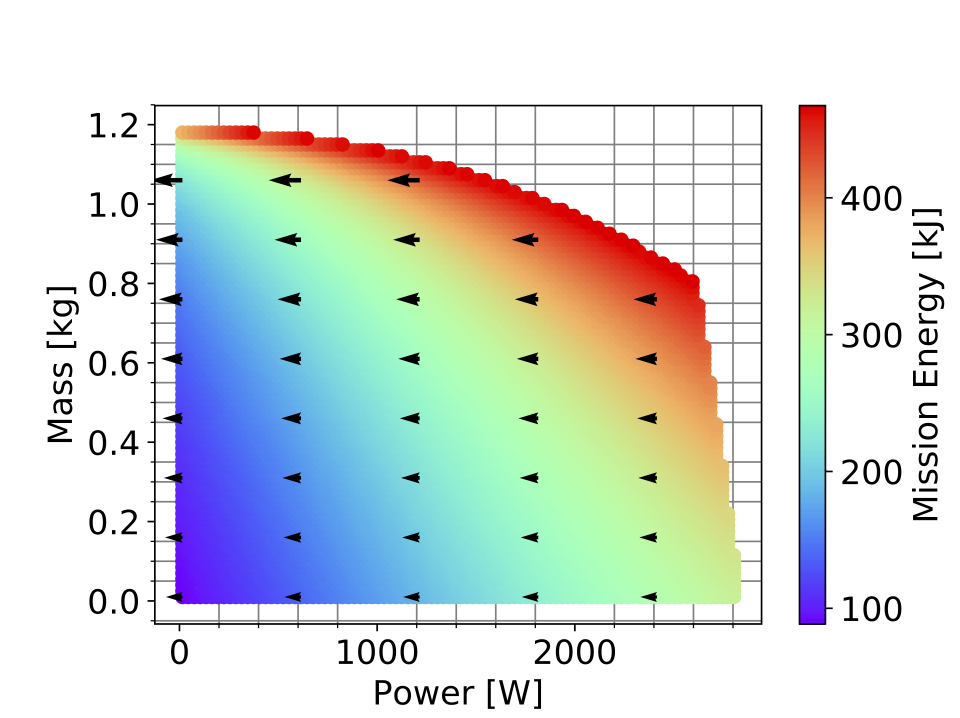}
    \caption{Mission energy-power gradient.}
    \label{fig:me_gradient_power}
    \end{subfigure}
     \caption{Mission metrics' gradient, the rate of change in the most optimal direction, and their components. Change in the gradient magnitude, shown in the arrow size, for a walk toward the center indicates diminishing return in (a), (b), and (c). However, in (d), gradient values stay the same since this gradient is equal to mission time which is not a function of power.}
         \label{fig:gradient}
\end{figure}

Second, some quantities exhibit diminishing returns, where moving towards more optimal (in this case smaller) values of that quantity results in lower gains. We demonstrate this by showing the gradient plots corresponding to one selected slice from Figure~\ref{fig:mt_ds} and Figure~\ref{fig:me_ds}. The other slices follow the same trend. The slice data are shown in Figures~\ref{fig:mt_gradient_mass}, and ~\ref{fig:mt_gradient_response_time} for Figure~\ref{fig:mt_ds} and in 
Figures~\ref{fig:me_gradient_mass} and ~\ref{fig:me_gradient_power} for Figure~\ref{fig:me_ds}. In these plots, gradient magnitude, i.e., mission metrics' rate of change, is shown by the size of the arrows overlaid on the figure. A change in the size of these arrows when moving along the $y$-axis or the $x$-axis indicates a change in the gradient magnitude. 

For mission time, lowering the mass reduces the gradient magnitude, indicating a smaller change in the mission time.
Figure~\ref{fig:mt_gradient_mass} shows that the size of the arrows shrinks as we lower the mass, i.e., as we move from top to bottom on the $y$-axis. 
Similarly, lowering the response time reduces the gradient magnitude of the mission time. Figure~\ref{fig:mt_gradient_response_time} shows that the size of the arrows shrinks as we lower the response time, i.e.,
as we move left along the $x$-axis.

For mission energy, a similar trend with respect to mass is observed (Figure~\ref{fig:me_gradient_mass}). However, such a trend is not universal across all quantities. For example, lowering power does not impact mission energy's gradient with respect to power. Figure~\ref{fig:me_gradient_power} shows that the size of the arrows stay the same as we lower the power, i.e., as we move left along the $x$-axis. This is because simply put, this gradient is equal to mission time which is not a function of power.
Hence, a change in power would not impact mission time, in other words, the gradient, and hence, the size of the arrows would not change.

Finally, different quantities have different impacts on mission time and energy. To demonstrate this, we use sensitivity analysis by comparing a mission metric's percentage change with respect to the quantities' percentage change. Table~\ref{tab:ds_sensitivity} shows the mean and variance for the sensitively values of different quantities throughout the entire design space. Mission time is most sensitive to the response time since the mean response time, i.e., 0.4, is the highest among the three quantities. In other words, improving the response time on average results in the mission time improvement the most. This means that if the drone's main objective is mission time optimality, system designers should focus their efforts on reducing the response time. On the other hand, mission energy shows equal sensitivity to all three quantities since the mean of all quantities are approximately the same, i.e., 0.4. This means that the designers who are focused on mission energy optimality should target the quantity whose improvement costs the least. It is worth noting power consumption does not have an impact on mission time, hence, the mean and variance are equal to 0. The lack of relationship is also shown in the cyber-physical interaction graph since there is no edge from power to mission time.
\renewcommand{\arraystretch}{1.5}
\begin{table}[t!]
\caption{Design space sensitivity to different compute related quantities. Higher mean means a higher sensitivity and hence a higher gain if the quantity is optimized.
The sensitivity analysis is done by comparing a mission metric's percentage change with respect to the quantities' percentage change.}
\footnotesize
\begin{tabular}{|l|c|c|c||c|c|c|}
\hline
     & \multicolumn{3}{c||}{\emph{Mission Time}} & \multicolumn{3}{c|}{\emph{Mission Energy}} \\ \hline
     & Response Time   & Mass   & Power  & Response time    & Mass   & Power   \\ \hline
Mean & .40             & .31    & 0      & .40               & .40     & .37     \\ \hline
Std  & .25             & .70    & 0      & .24              & .72    & .19     \\ \hline
\end{tabular}
\label{tab:ds_sensitivity}
\end{table}
\renewcommand{\arraystretch}{1}

\section{Compute Optimization Impact on Mission Time and Energy}
\label{sec:optimizations}
We use our benchmark suite, simulation environment, and the knowledge we acquired through examining compute's impact paths toward conducting two system optimization case-studies. We focus on the first and the third cluster (\Fig{fig:CIG_all}), i.e., optimizations exploiting the impacts through performance and power while leaving the second cluster for future work. The first case study examines the performance/power impact through a hybrid cloud-edge MAV complex, and the second one explores the performance/energy improvements through a runtime optimization.

\subsection{A ``Cloud-Edge Offloading Optimization'' Case Study}
We examine a cloud/edge drone where the computation is distributed across the edge and cloud endpoints. We compare a fully-onboard compute drone equipped with a TX2 versus a fully-in-cloud drone with a powerful cloud support. The ``cloud'' computational horsepower is composed of an Intel i7 4740 @ 4GHz with 32 GB of RAM and a GeForce GTX 1080. For network connectivity, we utilize a 1Gbp/s LAN, which mimics a future 5G network~\cite{agyapong2014design, gupta2015survey}.

\begin{figure}[t!]
\centering
    \begin{subfigure}{.5\columnwidth}
    \centering
    \includegraphics[trim=5 0 20 0, clip, height=1.3in]{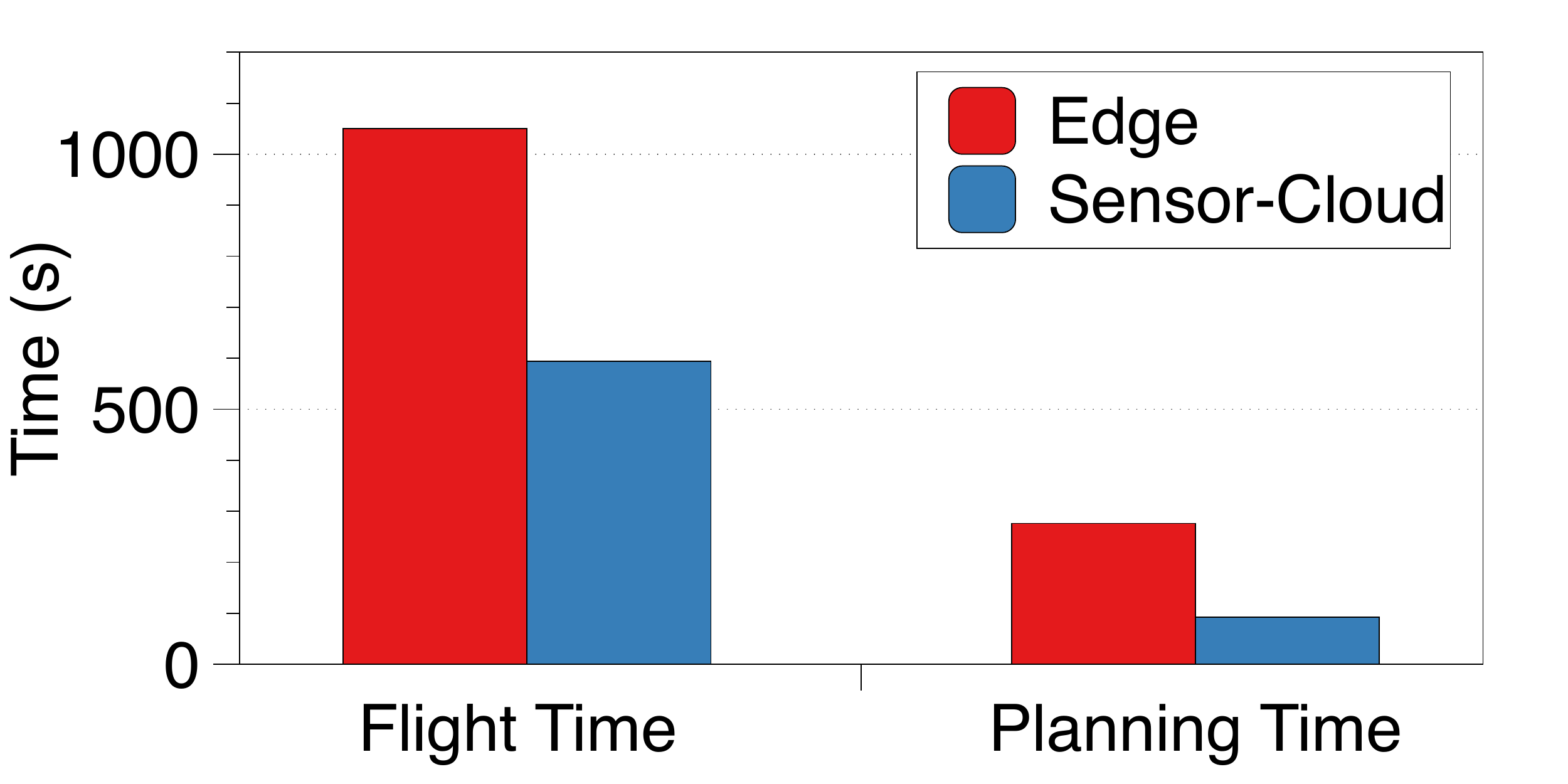}
    \caption{Performance.}
    \label{Perforamnce}
    \end{subfigure}
    \begin{subfigure}{.4\columnwidth}
    \centering
    \includegraphics[trim=5 10 0 0, clip, height=1.3in]{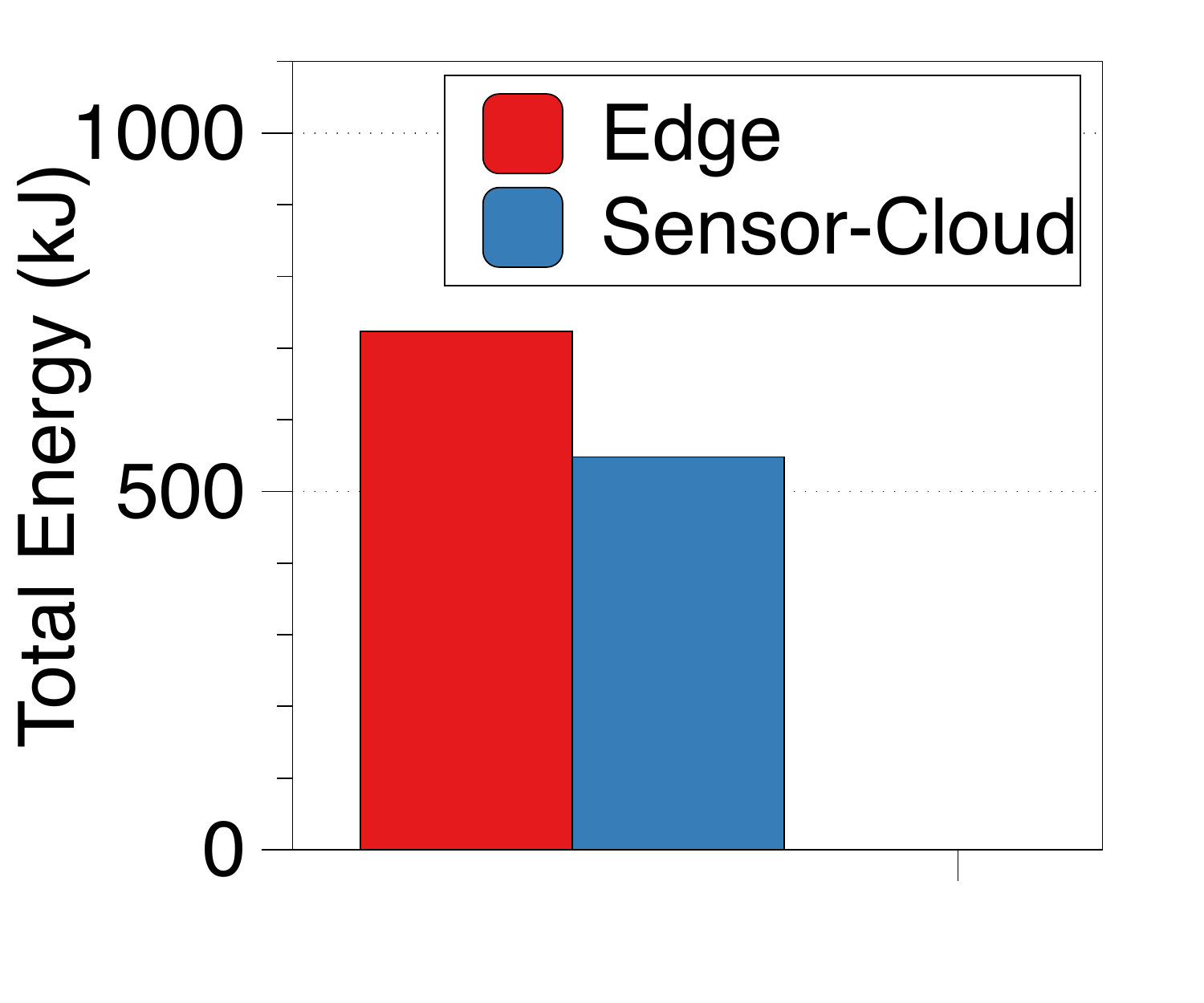}
    \caption{Energy.}
    \label{fig:cloud-edge-energy}
    \end{subfigure}
\caption{Comparing a fully-onboard compute drone versus a fully-on-cloud drone. Our system allows part or portion of the MAVBench workloads to be offloaded to the cloud.}
\label{fig:cloud_edge}
\end{figure}

We target the planning stage of the PPC pipeline and focus on the 3D Mapping as the application of choice to offload. As we show in \Fig{fig:cloud_edge}, a drone that can enjoy the cloud's extra compute power sees a 3X speed up in planning time. This improves the drone's average velocity due to hover time reduction, and hence reduces the drone's overall mission time by as much as 50\% (impact through compute performance). Reduction in mission time decreases the total system's energy consumption (Impact through performance). In addition, by offloading the computation to the cloud, and hence avoid embedding a powerful i7 machine on edge, drone's power and therefore total energy consumption is reduced (impact through power). The overall reduction in energy is 1.3X for a cloud-edge hybrid system vs. a full-on edge drone.  

\subsection{A ``Context-Aware Runtime System'' Case Study} 
Focusing on energy efficiency, we conduct a kernel/environment sensitivity analysis using the OctoMap node~\cite{octomap}, which is a major bottleneck in three of our end to end applications, namely package delivery, 3D mapping and search and rescue. OctoMap is used for the modeling of various environments without prior assumptions. The map of the environment is maintained in an efficient tree-like data structure while keeping track of the free, occupied and unknown areas. Both planning and collision avoidance kernels use OctoMap to make safe flight possible, via costly compute cycles, by only allowing navigation through free space. Due to its implementation efficiency, OctoMap is widely adopted in the robotics community. Its broad adoption and impact in two out of three stages (Perception and Planning) make this kernel highly general and important for optimization.
\begin{figure*}[!t]
    \centering
    \begin{subfigure}[t]{.43\linewidth}
        \centering
        \includegraphics[width=2.2in]{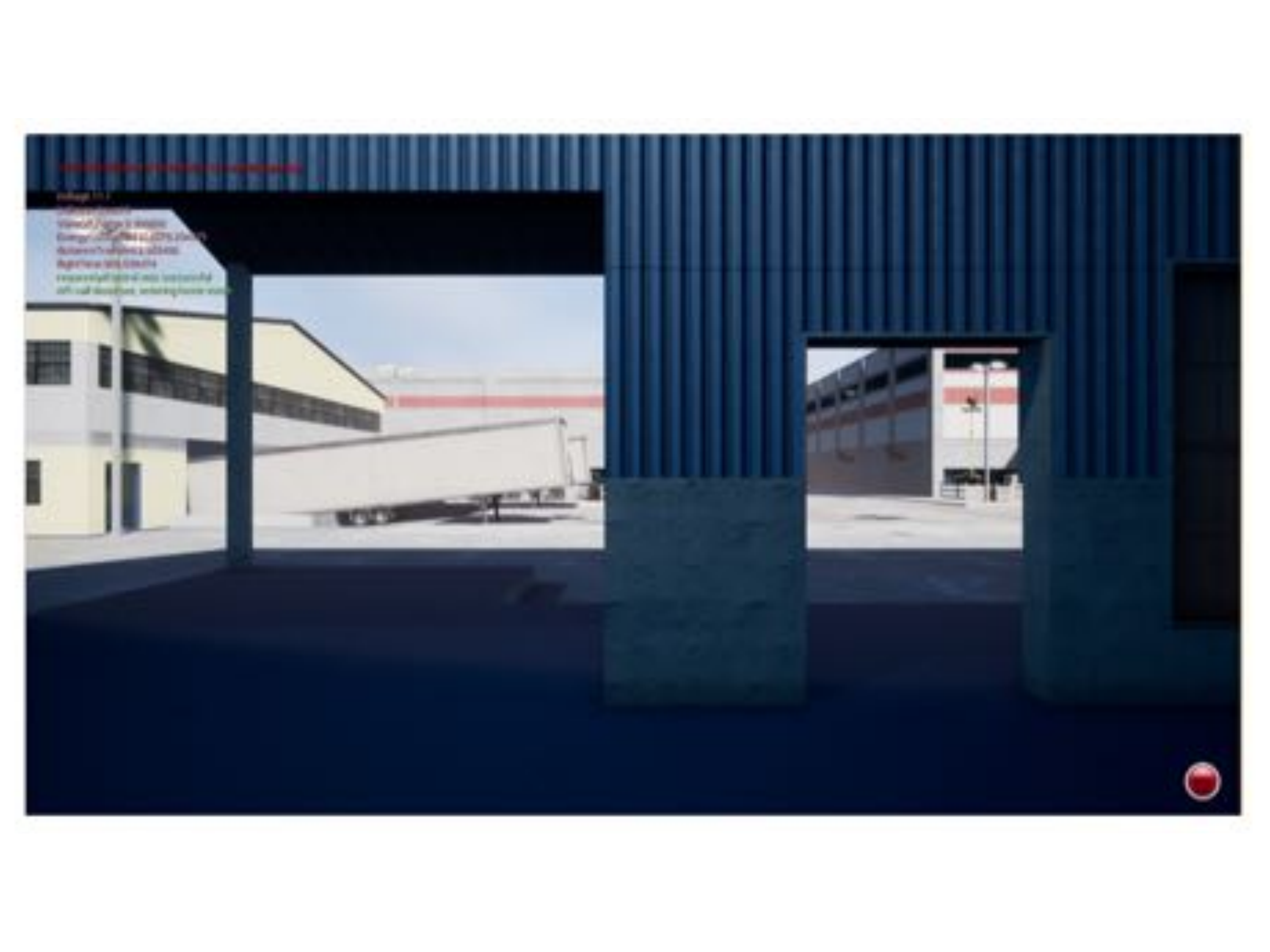}
        \caption{Environment's map.}%
        \label{fig:garage_sim}
    \end{subfigure}
   \hspace{20pt}
    \begin{subfigure}[t]{.43\linewidth}
        \centering
        \includegraphics[width=2.2in]{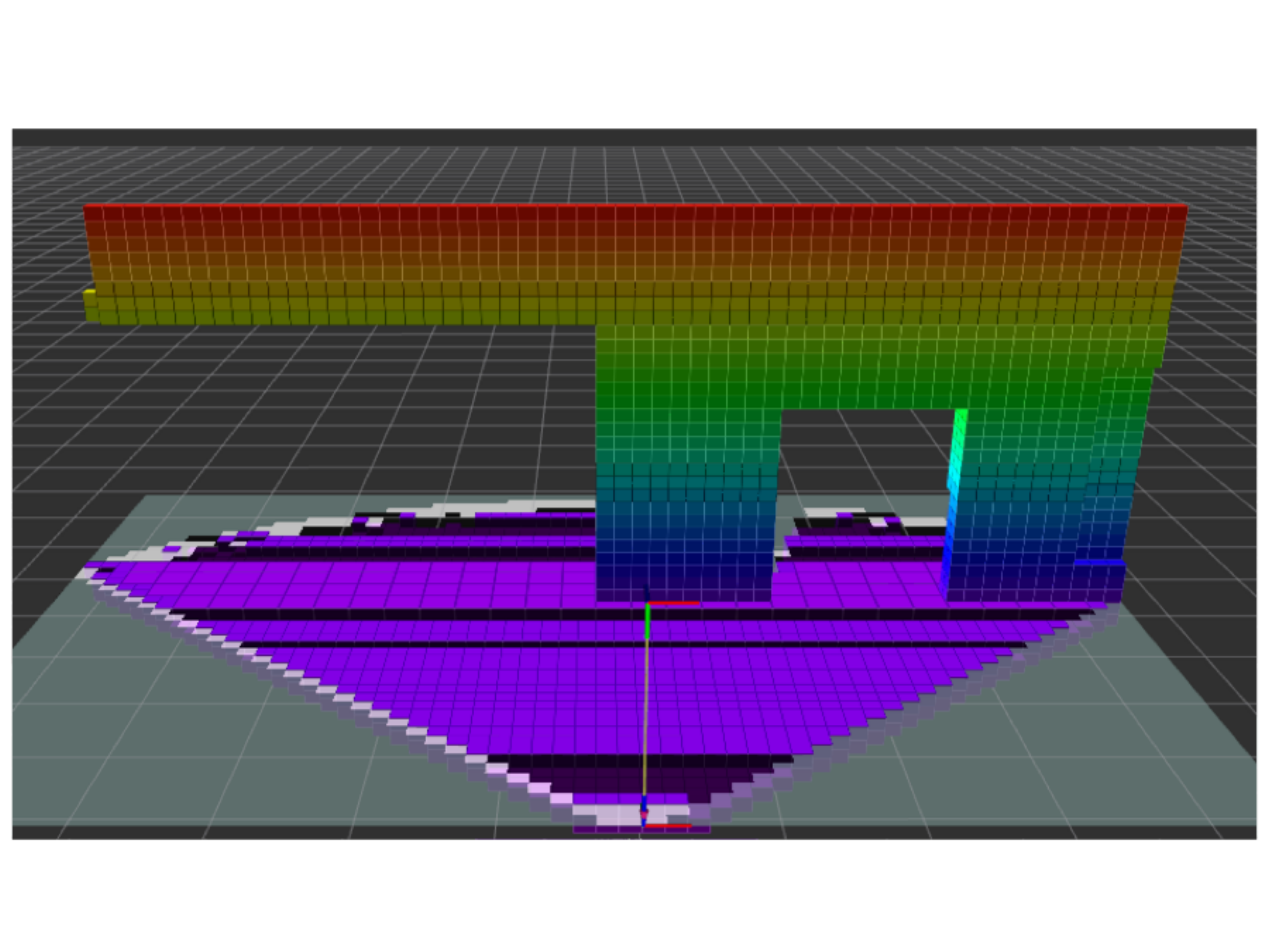}
        \caption{0.15~\emph{m} resolution.}
        \label{fig:rviz__15}
    \end{subfigure}
       
     \begin{subfigure}[t]{.43\linewidth}
        \vspace{5pt}
        \centering
        \includegraphics[width=2.2in]{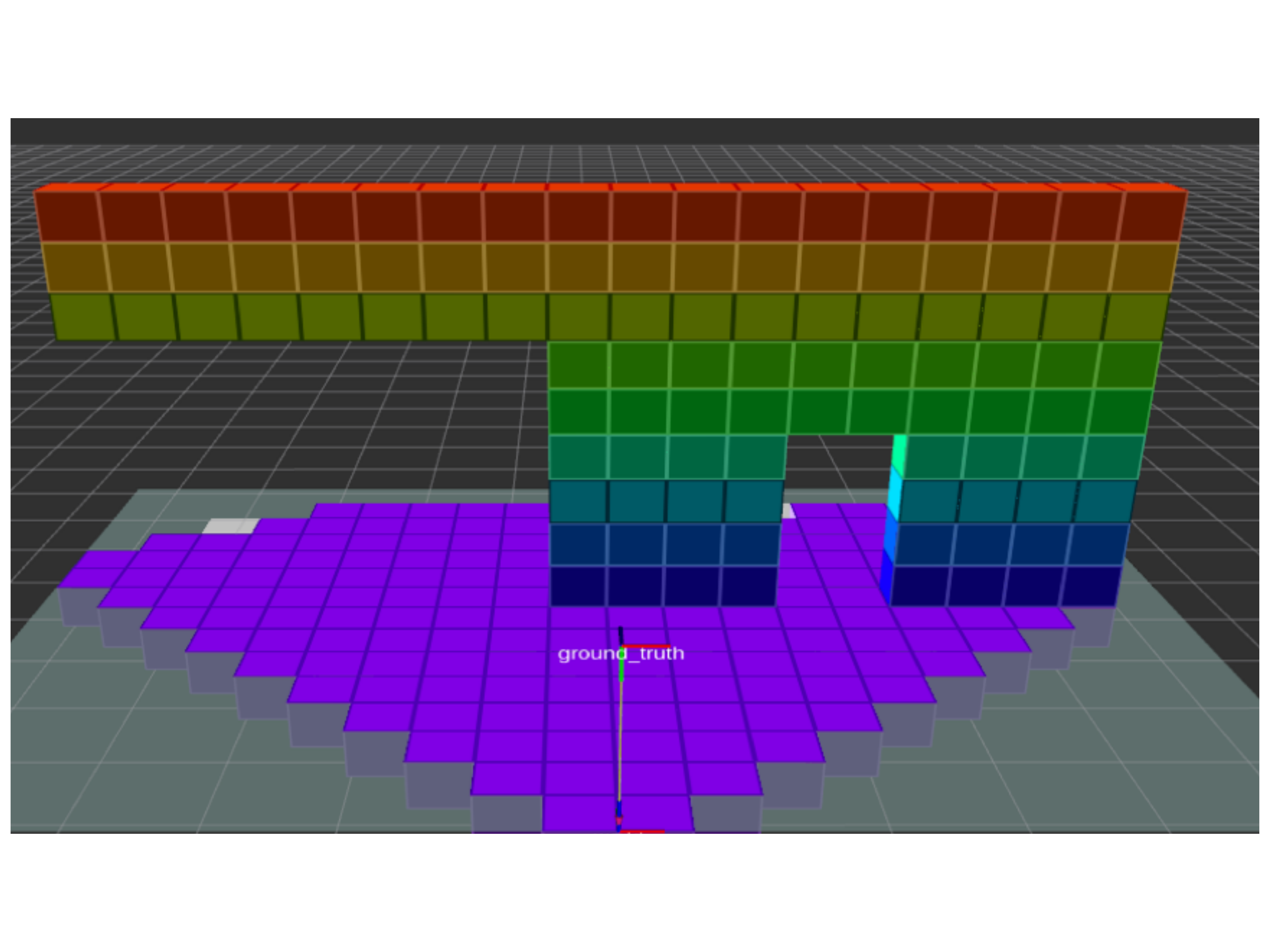}
     \caption{0.5~\emph{m} resolution.}
        \label{fig:game:rviz__5}
    \end{subfigure}
   \hspace{20pt}
   \begin{subfigure}[t]{.43\linewidth}
          \vspace{5pt}
        \centering
        \includegraphics[width=2.2in]{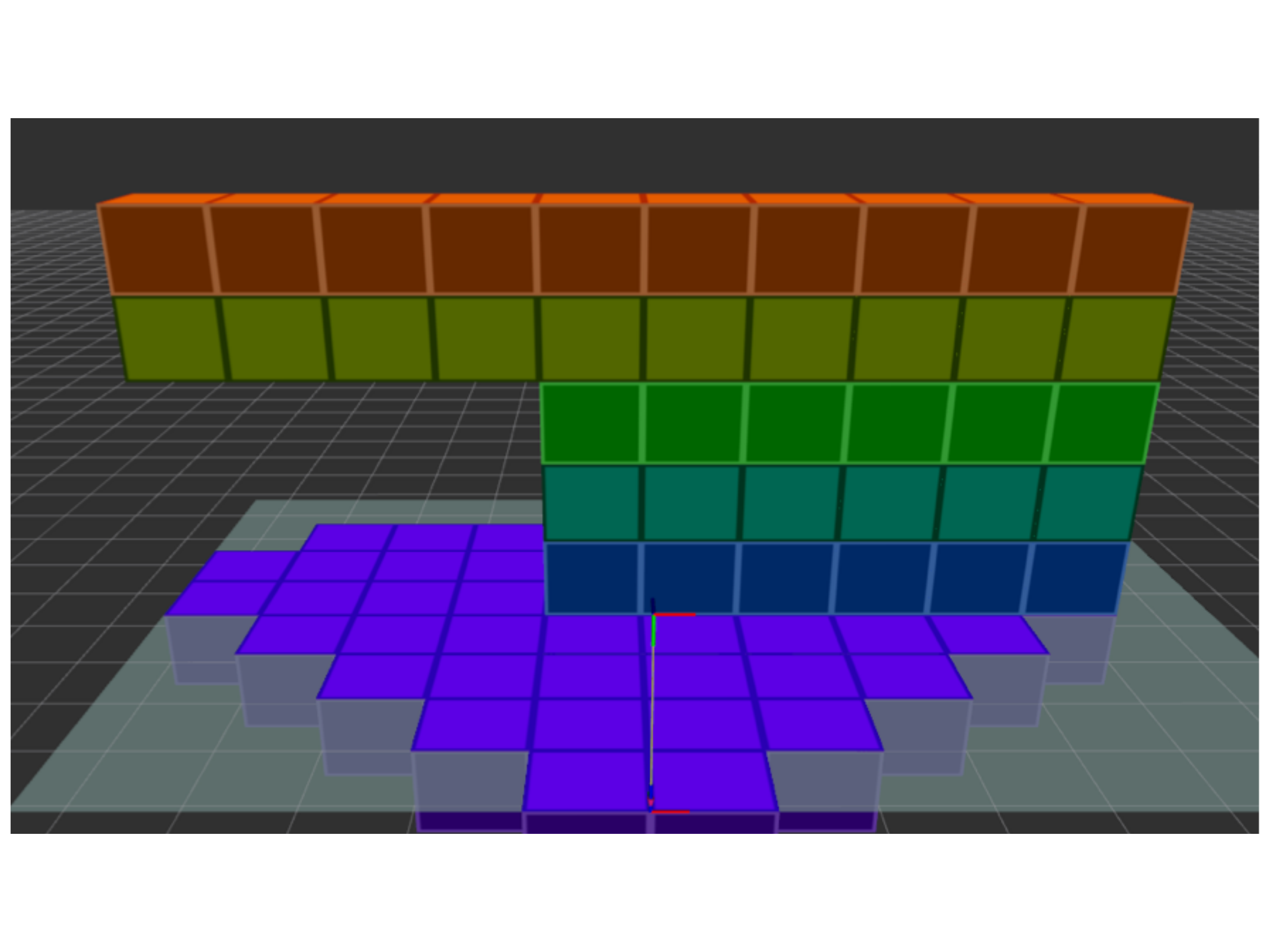}
        \caption{0.80~\emph{m} resolution.}
        \label{fig:game:rviz_1}
    \end{subfigure}
   \caption{For the environment in \emph{(a)}, OctoMap's resolution impact on the drone's perception of its environment is shown in \emph{(b)}, \emph{(c)}, \emph{(d)}. Large resolution means larger voxel size (lower is better).}
    \label{fig:octomap_perception}
\end{figure*}

\begin{figure*}[!t]
    \centering
    \begin{subfigure}[t]{.49\linewidth}
        \centering
        \includegraphics[trim=0 0 30 0, clip, width=1\columnwidth]{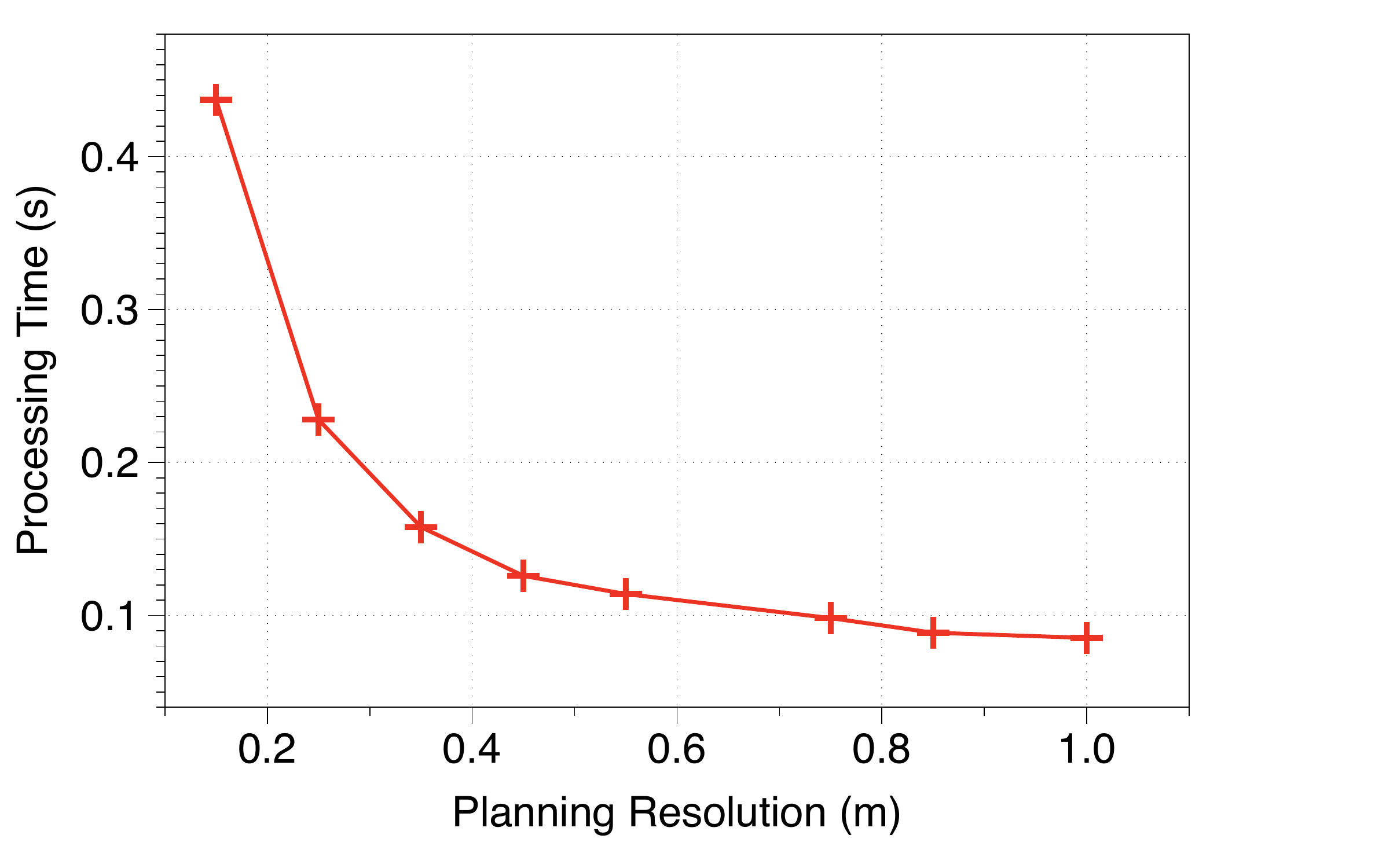}
        \caption{Octomap resolution Vs. performance time.}%
        \label{fig:octres}
    \end{subfigure}
    \begin{subfigure}[t]{.49\linewidth}
        \centering
        \includegraphics[trim=0 0 0 0, clip, width=1\columnwidth]{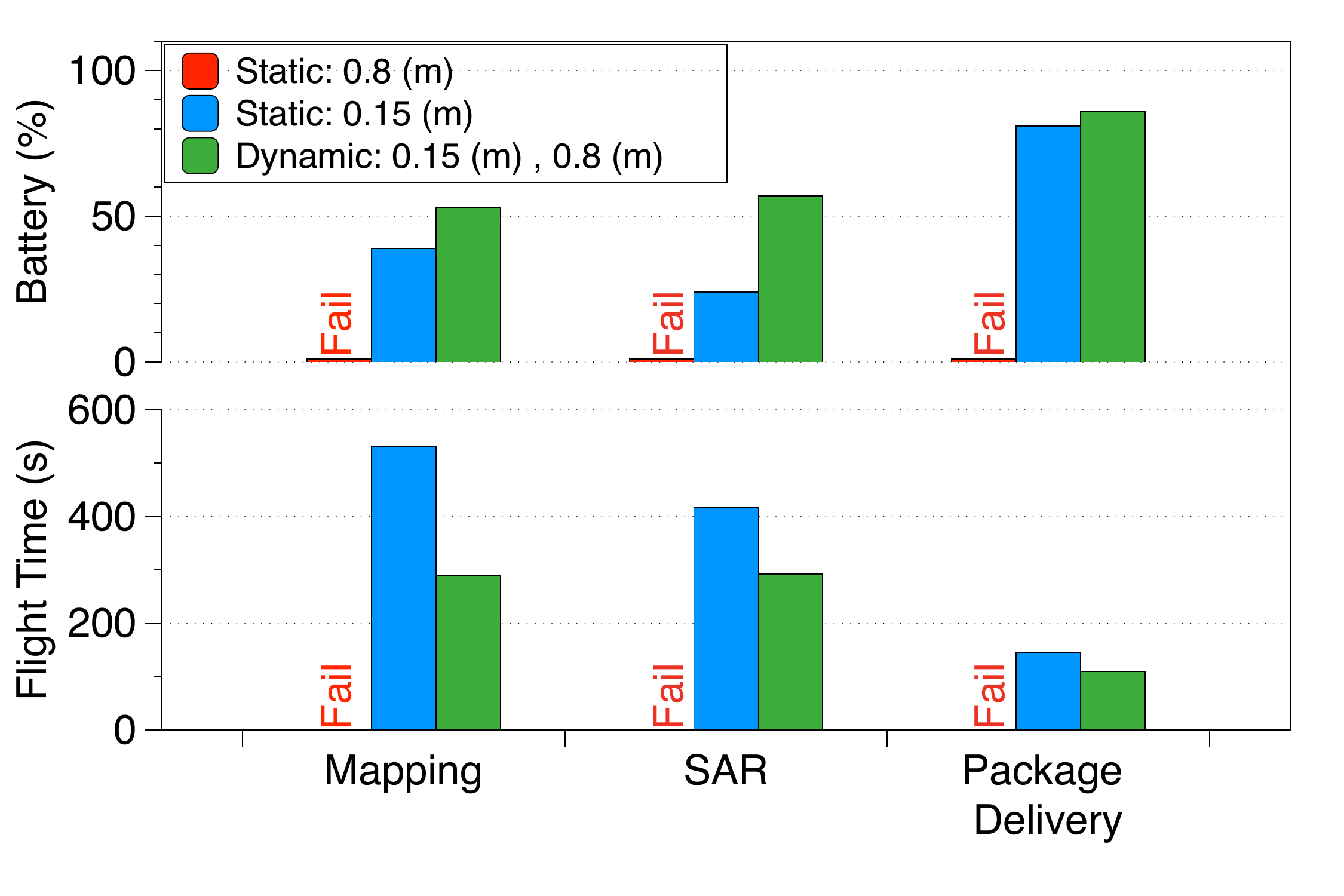}
        \caption{Context-aware Octomap resolution tuning.}
        \label{fig:dual-res}
    \end{subfigure}
\caption{\emph{(a)} Reduction in OctoMap resolution (accuracy) can be traded off with processing time. Increasing the $x$-axis means larger voxels to represent the space more coarsely (less accurately). A 6.5X reduction in resolution results in a 4.5X improvement in processing time. \emph{(b)} Switching between OctoMap resolutions dynamically leads to successfully finishing the mission compared to 0.80~m. It also leads to battery life improvement compared to 0.15~m. The $y$-axis in the top graph is the battery left on the drone upon mission completion.}
\end{figure*}

The voxel size in OctoMap, i.e., the map's resolution, introduces accuracy versus flight-time/energy tradeoff. By lowering the resolution, i.e., increasing \emph{voxel} sizes, obstacle boundaries get inflated; hence, the drone's perception of the environment and the objects within it becomes inaccurate. We illustrate the impact of OctoMap resolution on the drone's perception using \Fig{fig:octomap_perception}. \Fig{fig:garage_sim} {shows the environment and  Figures~\ref{fig:rviz__15}, \ref{fig:game:rviz__5}, \ref{fig:game:rviz_1} show the drone's perception of the environment as a function of OctoMap resolution. When the resolution is lowered, the voxels size increases to the point that the drone fails to recognize the openings as possible passageways to plan through (Figure~\ref{fig:game:rviz_1}). This results in mission time inefficiency and failures depending on the environment. In some cases, if the resolution is too large, the drone can't find a path to complete its mission.

To examine the accuracy versus performance tradeoff, we measured OctoMap kernel's processing time (running in isolation) while varying its resolution knob. \Fig{fig:octres} shows that as planning resolution increases (i.e., voxels are larger so space is represented more coarsely and hence less accurately), performance improves dramatically because less compute is needed. Going from one extreme to another, when the planning resolution goes from less than 0.2~m to 1.0~m ($x$-axis), OctoMap's processing time (or update rate) goes from more than 0.4~seconds to less than 0.1~seconds ($y$-axis). A 6.5X reduction in accuracy results in a 4.5X improvement in processing time.

Certain aspects like obstacle density in the environment determine the ``ideal'' OctoMap resolution. In low-density environments, where the drone has many obstacle-free paths to take, a low resolution can suffice. In dense environments, low resolutions can deprive the drone of viable obstacle-free paths because the drone perceives the obstacles to be larger than they are in the real world, and so plans to avoid them. Since the drone's environment constantly changes, a dynamic approach where a runtime sets the resolution is ideally desirable.

We study two environments during the mission, namely outdoors (low obstacle density) and indoors (high obstacle density). \Fig{fig:dual-res} shows the result of two static (predetermined) resolutions, 0.15~m and 0.80~m, and our dynamic approach that multiplexes between the two appropriately.\footnote{Resolutions are based on the environment like the door width size. A 0.15~m resolution is chosen to ensure that the drone (diagonal width of 0.65~m) considers an average door (width of 0.82~m) as an opening for planning.} The dynamic approach allows improvement of battery consumption by up to 1.8X. 
Intuitively, as compute reduces, OctoMap bottleneck eases, and therefore, the drone completes its mission faster (impact through performance). The figure also highlights another interesting relationship that statically choosing the 0.80~m resolution to optimize for compute (only) causes the drone to fail its mission since it is unable to plan paths through narrow openings in the indoor environments. Instead, by switching between the two resolutions according to the environment's obstacle density, the dynamic approach is able to balance OctoMap computation with mission feasibility and energy, holistically. In all cases, the dynamic approach uses less energy and retains more battery at mission end time.

\section{Related Work}
\label{sec:related}
There is prior work that focuses on building analytical models, simulators and benchmark suites to aid the development of autonomous MAVs. We address some shortcomings of previous approaches by
providing a more detailed, integrated, end-to-end solution.

\paragraph{Analytical Models} There are numerous work such as ~\cite{quad_dynamics, quad_dynm2, quad_survey, energyaware} that model the MAV's physical quantities' impact on one another and ultimately MAV's
behavior. These works do not consider cyber quantities and their influence. There are a few works such as ~\cite{high-speed-nav, embedded_img_proc, how_fast_too_fast} that brush the surface of cyber quantities' impact on physical quantities and mission metrics. However, none,
take a detailed and holistic compute subsystem design perspective. For example, ~\cite{high-speed-nav} and ~\cite{how_fast_too_fast} only consider the relationship between response time and velocity whereas
~\cite{embedded_img_proc} does not consider acceleration, velocity or system throughput. 
In addition ~\cite{embedded_img_proc} and ~\cite{how_fast_too_fast} only examine one stage of the pipeline, i.e., the perception stage, whereas our applications enjoy the end-to-end characteristics. 



\paragraph{Simulators} Simulators are essential to the study of aerial and robotic agents. Our simulation platform is built upon Microsoft's AirSim \cite{Airsim_paper}, a UAV simulator that uses the Unreal Game Engine to provide accurate physics models and photo-realistic environments. MAVBench uses the AirSim core and extends it with performance, power and battery models that are suited for architectural research, as well as with a gimbal, and dynamic and static obstacle creation capabilities that are not inherently part of AirSim. Another very popular simulator used in the robotics community for MAVs is Gazebo~\cite{gazebo}. However, Gazebo simulations lack photo-realism, while our work, with the help of AirSim and the Unreal Game Engine, enables more accurate visual modeling.

There are also numerous simulators widely used in industry and academia for studying autonomous agents such as ~\cite{boss-cmu,talos-mit,odin-vtech,uber-simulator, uav_benchmark_simulator, mujoco}. However, they either do not provide MAV models or does not consider the architectural insights.


A recent work FlightGoggles~\cite{flight-goggles}, creates virtual reality environments for drones using the images streamed from the Unity3D game engine. However, for maximum realism, FlightGoggles requires a fully functioning drone that must fly during tests, with its sensory data being streamed in from the game engine. MAVBench, on the other hand, does not have this constraint. Our users may provide real processors for hardware-in-the-loop simulation, but they are not required to fly the MAVs physically in the real world.

\paragraph{Benchmarks} Most robot benchmark suites target individual computational kernels, such as odometry or motion-planning, rather than characterizing end-to-end applications composed of many
different kernels. For example SLAMBench~\cite{slambench} and CommonRoad~\cite{common-road} solely focus on the perception and the planning stage respectively. However, our benchmarks allow for
holistic studies by providing end-to-end applications. 

\paragraph{Simulator + Benchmarks:} RoboBench~\cite{robobench} provides a common platform around simulators and benchmarks using software containers to combat software compatibility
problems across groups. This work is complementary to ours.


\section{Conclusion}
\label{sec:conclusion}

We show that a tight interaction between
the cyber and physical processes dictates autonomous mobile machines' behavior. Hence, a robot design methodology needs to consider such interactions for optimal results. Furthermore, this consideration
can uncover co-design (cyber-physical co-design)
opportunities, similar to hardware-software co-design, to be taken advantage of. This paper sets its goal as to probe and investigates the cyber and physical interactions of Micro Aerial Vehicles (MAVs) as examples of complex
robots. We frame this goal as a simple question: \textit{what is role of compute in the operation of autonomous MAVs?}
To answer this question, we combine analytical models, benchmarks and simulations showing how \textit{fundamentals of compute and motion are interconnected}.
For our analytical models, we use detailed physics, capturing compute's impact on various mission metrics. For our simulator and benchmarks, we address the lack of systematic benchmarks and
infrastructure for research by developing MAVBench, a first of its kind platform for the holistic evaluation of aerial agents, involving a closed-loop simulation framework and a benchmark suite. Using
our tool sets, we provide two optimization case studies through which we improve mission time and energy. 
As a concluding remark, while we focus on autonomous drones, the lessons learned about the role of computing, and the simulation methodology (i.e., closed/hardware in the loop) readily extend to other autonomous machines and robots.
By open sourcing our tool-sets, we hope to \textbf{(1)} help raising the
understanding of compute subsystem (system) designers of cyber-physical machines, \textbf{(2)} enable the cyber-physical co-design paradigm, and \textbf{(3)} start a closer discussion/collaboration between the robotics and
system design communities. 



\section*{Acknowledgements}

This material is based upon work supported by the NSF under Grant No. 1528045 and funding from Intel and Google.

\bibliographystyle{ACM-Reference-Format}
\bibliography{references}

\end{document}